UNIVERSITY OF THE DISTRICT OF COLUMBIA 1851

# Predictive Models for Wind Speed

By

Md Amimul Ehsan

A thesis submitted to the

University of the District of Columbia

in Partial Fulfillment of the

Requirements for the Degree of

Master of Science in

Electrical Engineering

Washington, DC

December 2019

# University of the District of Columbia
# Department of Electrical and Computer Engineering

## Certificate of Approval of Thesis

**Student Name**: Md Amimul Ehsan Student ID: N00293795

**Major Field of Study:** Digital Systems Engineering

a. **Thesis Title:**
Predictive Models for Wind Speed Using

We have met, reviewed, recommended, and approved the thesis with a grade

## I. Approvals

| | |
|---|---|
| Name – Major Advisor (typed and signed)<br><br>Dr. Amir Shahirinia | Name – Committee Member (typed and signed)<br><br>Dr. Sasan Haghani |
| Name – Committee Member (typed and signed)<br><br>Dr. Jeff Gill | Name – Committee Member (typed and signed)<br><br>Dr. Nian Zhang |

| | | |
|---|---|---|
| **Office Use Only** | Recommended by: MSEE Program Director<br><br>Dr. Wagdy H. Mahmoud | Date |
| | Recommended by: ECE Department Chair<br><br>Dr. Esther Ososanya | Date |
| | Recommended by: SEAS Dean<br><br>Dr. Devdas Shetty | Date |

**Approved for the Faculty**

______________________________
Dr. Devdas Shetty
Date: ____________________

**CERTIFICATE OF APPROVAL OF THESIS**

Predictive Models for Wind Speed Using Artificial Intelligence and Copula

By

**Md Amimul Ehsan**

**Graduate Advisory Committee**

______________________ ______________

Dr. Amir Shahirinia, Major Advisor Date

______________________ ______________

Dr. Sasan Haghani Date

______________________ ______________

Dr. Jeff Gill Date

______________________ ______________

Dr. Nian Zhang Date

**Approved for the Faculty**

________________________

Dr. Devdas Shetty,

Dean of SEAS

Date: ___________________

UNIVERSITY OF THE DISTRICT OF COLUMBIA
SCHOOL OF ENGINEERING AND APPLIED SCIENCES

## University of the District of Columbia
## Department of Electrical and Computer Engineering
### MSEE Program of Study

***Important: The advisory/thesis committee must be approved before reviewing the MSEE student program of study (POS). It is recommended to submit the POS by the end of the second semester of registration. Approved committee members should review and sign the form before submitting it to the Graduate school for approval.***

Student Name: Md Amimul Ehsan          Student ID: N00293795

### I. Planned Graduate Program

*Important: if transferring graduate credits from another university, a transcript must be attached. All graduate course taken for graduate credits must be included in this form. Courses taken for zero graduate credits may be shown in the POS for information purposes.*

| Line | Course Number | Course title (abbreviate to fit on one line) | Credits | Semester/Year |
|---|---|---|---|---|
| 1 | ELEC 571 | Linear systems | 3.00 | Spring/2018 |
| 2 | ELEC 507 | Probability and Random Processes | 3.00 | Spring/2018 |
| 3 | ELEC 584 | Digital System-level Design | 3.00 | Spring/2018 |
| 4 | STAT 618 | Bayesian Statistics | 3.00 | Fall/2018 |
| 5 | ELEC 559 | Computer Architecture | 3.00 | Spring/2019 |
| 6 | ELEC 678 | Adv Digital Integrated Cir Design | 3.00 | Spring/2019 |
| 7 | CSCI 578 | Machine Learning | 3.00 | Spring/2019 |
| 8 | CSCI 578 | Deep Learning | 3.00 | Spring/2019 |
| 9 | ELEC 586 | Advanced Embedded Sys Design | 3.00 | Fall/2019 |
| 10 | ELEC 699 | Thesis | 6.00 | Fall/2018/Fall/2019 |

### II. Approvals

| | |
|---|---|
| Dr. Amir Shahirinia – Major Professor<br>A. H. Sh | Dr. Sasan Haghani – Committee Member<br>Sasan Haghani |
| Dr. Jeff Gill – Committee Member | Dr. Nian Zhang – Committee Member<br>Nian Zhang |

| | | |
|---|---|---|
| **Office Use Only** | Student's Signature | Date: 10/07/2019 |
| | Recommended by: MSEE Program Director | Date: |
| | Recommended by: ECG Department Chair | Date: |
| | Approved by: Dean of Graduate School | Date: |

***This thesis is dedicated to my parents.***

*For their endless love, support and encouragement*

# Acknowledgment

I am sincerely thankful to my advisor, Dr. Amir Shahirinia. He was always helpful and available whenever I faced difficulties and guided me in the right direction throughout my studies. I will always appreciate his support and cooperation.

My earnest appreciation to the committee members: Dr. Jeff Gill, Dr. Sasan Haghani, and Dr. Nian Zhang, for their invaluable input to my thesis.

I also appreciate the professors, who guided me through, taught classes, and supported me all along. Special thanks to Dr. Esther Ososanya and Dr. Devdas Shetty, who's support was precious.

I would like to express my deep appreciation to the Department of Electrical and Computer Engineering, and the School of Engineering and Applied Sciences (SEAS), for providing the opportunity and support me during my program at the University of the District of Columbia.

I am also thankful to my Family members and Friends for their continuous support and presence whenever I needed.

Finally, I am thankful to God for bringing all these amazing people in my life and guiding me through the right direction towards my dreams.

# Table of Contents

## List of Figures

# List of Tables

# Abstract

Electricity generation from burning fossil fuel is one of the major contributors to global warming. Renewable energy sources are a viable alternative to produce electrical energy and to reduce the emission from power industry. These energy sources are the building blocks of green energy, which all have different characteristics. Their availabilities are also diverse, depending on geographical locations and other parameters. Low implementation cost and distributed availability all over the world uplifts their popularity exponentially. Therefore, it has unlocked opportunities for consumers to produce electricity locally and use it on-site, which reduces dependency on centralized utility companies.

Moreover, self-sustaining distributed generator entities are now able to sell their surplus electricity to the national grid that brings economic benefits. On the other hand, the conventional utility grid does not have provisions for the distributed energy sources integration. Therefore, a robust, and secured grid architecture, also known as a smart grid, is a need at present.

This research is focused on wind energy that is highly dependent on the wind speed profile of a certain location. Understanding the nature of wind speed is a challenge for optimum wind energy harness. The research considers two main objectives: the prediction of wind speed that simplifies wind farm planning and feasibility study. Secondly, the need to understand the dependency structure of the wind speeds of multiple distant locations.

To address the first objective, twelve artificial intelligence algorithms were used for wind speed prediction from collected meteorological parameters. The model performances were compared to determine the wind speed prediction accuracy. The results show a deep learning approach, long short-term memory (LSTM) outperforms other models with the highest accuracy of 97.8%. For dependency, a multivariate cumulative distribution function, Copula, was used to find the joint distribution of two or more distant location wind speeds, followed by a case study. We found that the appropriate copula family and the parameters vary based on the distance in between. For the case study, Joe-Frank (BB8) copula shows an efficient joint distribution fit for a wind speed pair with a standard error of 0.0094. Finally, some insights about the uncertainty aspects of wind speed dependency were addressed.

# Chapter One: Introduction

Renewable energy sources integration to the power grid not only promises to a distributed and robust smart grid of tomorrow but also it is the corridor to deal with greenhouse gas emission from the electric power generation sector [1, 2]. However, the intermittent nature of renewable energy sources and their diverse accessibilities based on geographical locations require a deep understanding of their inherent characteristics. This knowledge is crucial to design the grid control, operation, and management schemes [3]. Wind energy is getting great attention in some parts of the world, while its appeal is growing everywhere [4]. Wind speed is one of the key factors to understand before and after installing a wind farm [5]. However, understanding the nature of the wind speed has been a research topic, and the dimensions of insights are very diverse as well, yet still explored for more intelligent extrapolation. One of the important questions to answer during the feasibility study phase of any farm site is the wind speed profile at the proposed turbine height [6, 7]. The usual approach is to install a wind a speed monitoring setup at the proposed height, collect data for 1-3 years that later used for forecasting of both wind power and economic prospect [8]. Installation at the desired height is complex, and maintenance over the years require consistent monitoring. Therefore, we tackled this problem of finding a practical solution for wind speed profiling using the easy to collect meteorological information. Later, we attempted another research problem, how the dependency structure looks like for the wind speeds of distant locations. We propose to leverage the copula dependency modeling to find the joint distribution that can produce wind speed pattern reflecting real scenario. In the following sections, the motivation and objectives of this thesis are articulated.

## 1.1 Motivation and Overview

The key motivation behind this research is how to reduce the currently experiencing climate-changing effects all around the world, especially the emissions from the power industry. The renewable energy grid integration is scaling up every day that mandates the power grid to evolve to a renewable-penetrated system that encapsulates higher degrees of robustness and interoperability. Wind energy is one of such renewable energy sources that is studied in this research to understand some extent of its characteristics.

### 1.1.1 Climate Change and Electricity

At present, climate change is one of the most challenging questions in the research and non-research community. Everyone from policymakers to researchers is trying to identify the causes and ways to mitigate the effects [9]. It puts a question mark on the survival of humankind in tomorrow's picture. Global warming is a starting cause of climate change, as the existence of an increasing trend in the distribution of global temperature characteristics [10]. Unplanned industrialization and deforestation are the main reasons for temperature rise. Again, the intensity of these factors is concentrated in some parts of the world, while the whole planet also suffers from its effects. A few of the evident consequences are- natural hazards, politics [11, 12], human health [13, 14] to more technical ones, such as- transportation industry [15, 16], power systems [17] and renewable energy trends [18]. While there are many factors behind the wheel, there is one obvious reason that drives others, electricity. It is the fuel behind the industrialization, and most of the human aspire from computing and biomedical applications to space missions. Therefore, careful and effective considerations are necessary to work out this issue bottom up. Until now, the major share of electric power generation worldwide is from conventional and nuclear sources [19]. However, recent discoveries in the last two decades drive the trend to renewable energy sources that are eco-friendly and sustainable. Most of these resources are well distributed worldwide and so is a viable option to fight the greenhouse gas emission which is the root cause of global warming.

### 1.1.2 Power Systems Evolution and Future Trends

There are currently two major sources of electricity- conventional fossil fuel [20] and nuclear energy [21]. The conventional power generation causes an extensive amount of carbon emission (33% of total U.S. carbon emission in 2018 [22]), and nuclear is a constant radioactive risk [21].

#### 1.1.2.1 Conventional Fossil Fuel

Conventional fossil fuel comes from nature and never replenishes once used. Few of such sources are- coal, natural gas, and petroleum [23]. All these sources come from the natural process of absorption. However, it takes millions of years to replenish what once used, and all these sources are hydrocarbons. Burning hydrocarbons produces carbon-based oxides (greenhouse gases) as a by-product, which is the major cause of global warming. Figure 1 [24] shows the share of fossil fuel in comparison with other sources in the global energy mix. The worldwide scenario of fossil fuel usage (64% of total energy consumption) is conspicuous from the pie chart.

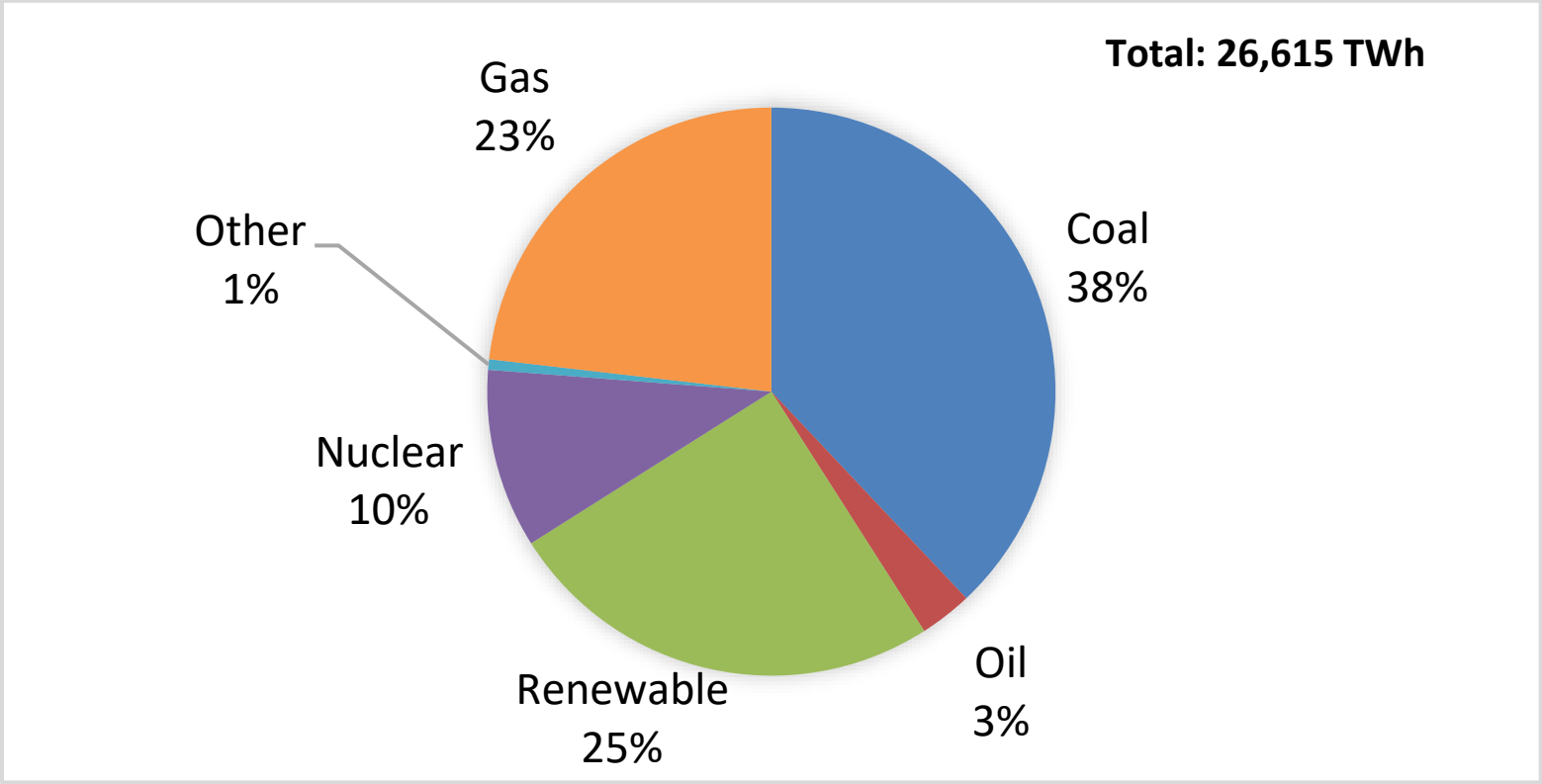

Figure 1. Global Energy Mix in 2018

### 1.1.2.2 Nuclear Energy

Nuclear energy is the utilization of energy yields from radioactive decay of chemically heavy molecules or isotopes to generate primarily heat and then electricity. 10.2% of global electricity use is coming from nuclear power plants; however, it does not show an increasing trend over the last ten years [25]. Figure 2 [26] shows how the whole process works. Currently, available technology can utilize only nuclear fission for commercial power production. Nuclear fission is the splitting of a heavy, unstable nucleus into two lighter nuclei, and the process dissipates a huge amount of energy in the form of heat that later converted into electricity. At the end of the nuclear fuel life cycle, we end up with highly dangerous radioactive nuclear waste [27]. Securing these hazardous materials is a risky task and may cause a catastrophe at any time [28]. In history, the world has suffered from radiation in many times in the form of both intentional and unintentional hazards. The recent one was the Fukushima nuclear disaster in Japan [29]. Therefore, many countries are reducing or reversing the nuclear power generation schemes from the energy mix.

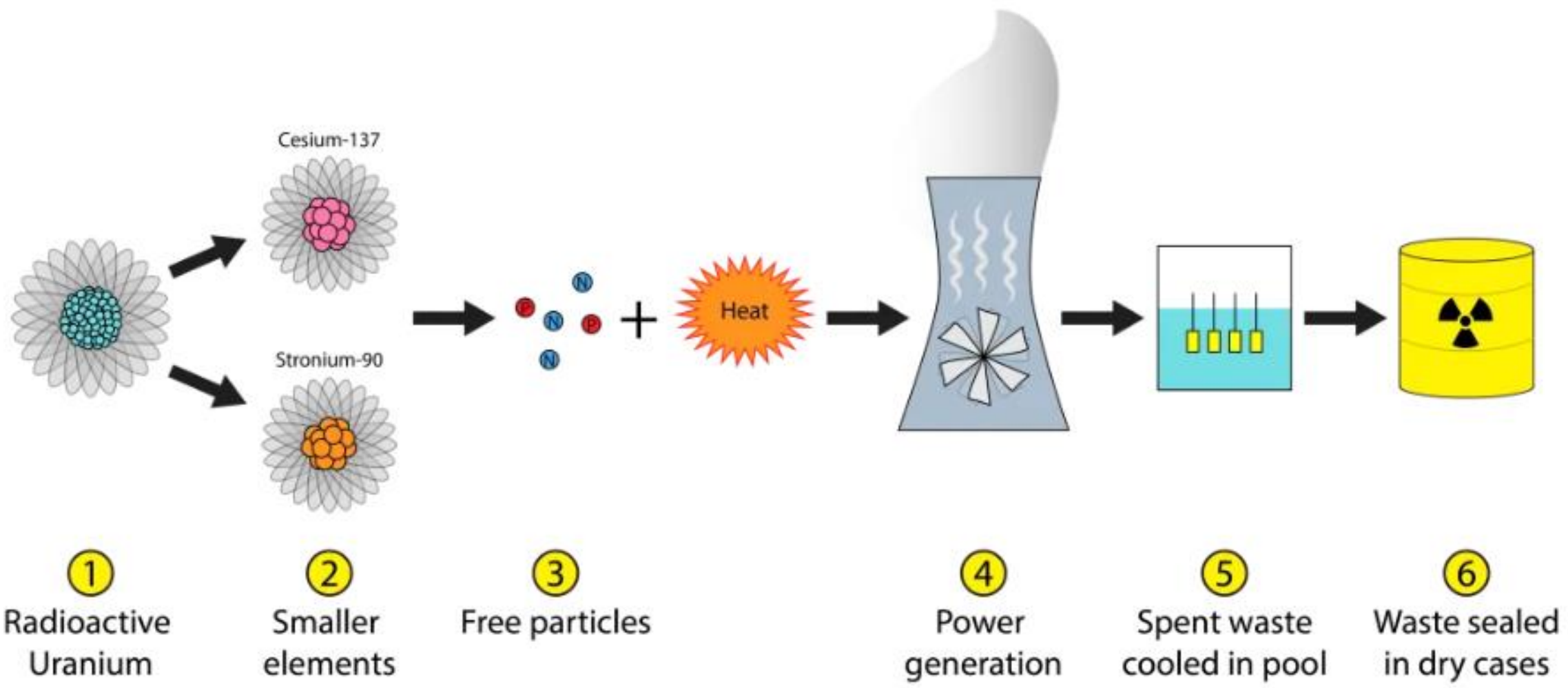

Figure 2. Life Cycle Illustration of Nuclear Fuel

### 1.1.2.3 Renewable Energy Sources

The United States Energy Information Administration (EIA) defines renewable energy as- "Renewable energy is energy from sources that are naturally replenishing but flow-limited, renewable resources are virtually inexhaustible in duration but limited in the amount of energy that is available per unit of time" [22]. A few of the commonly utilized forms of renewable energy sources are hydro, biomass, solar, biogas, and wind. The reduction of greenhouse gasses is a common benefit from all of them, while every source has its pros and cons [30].

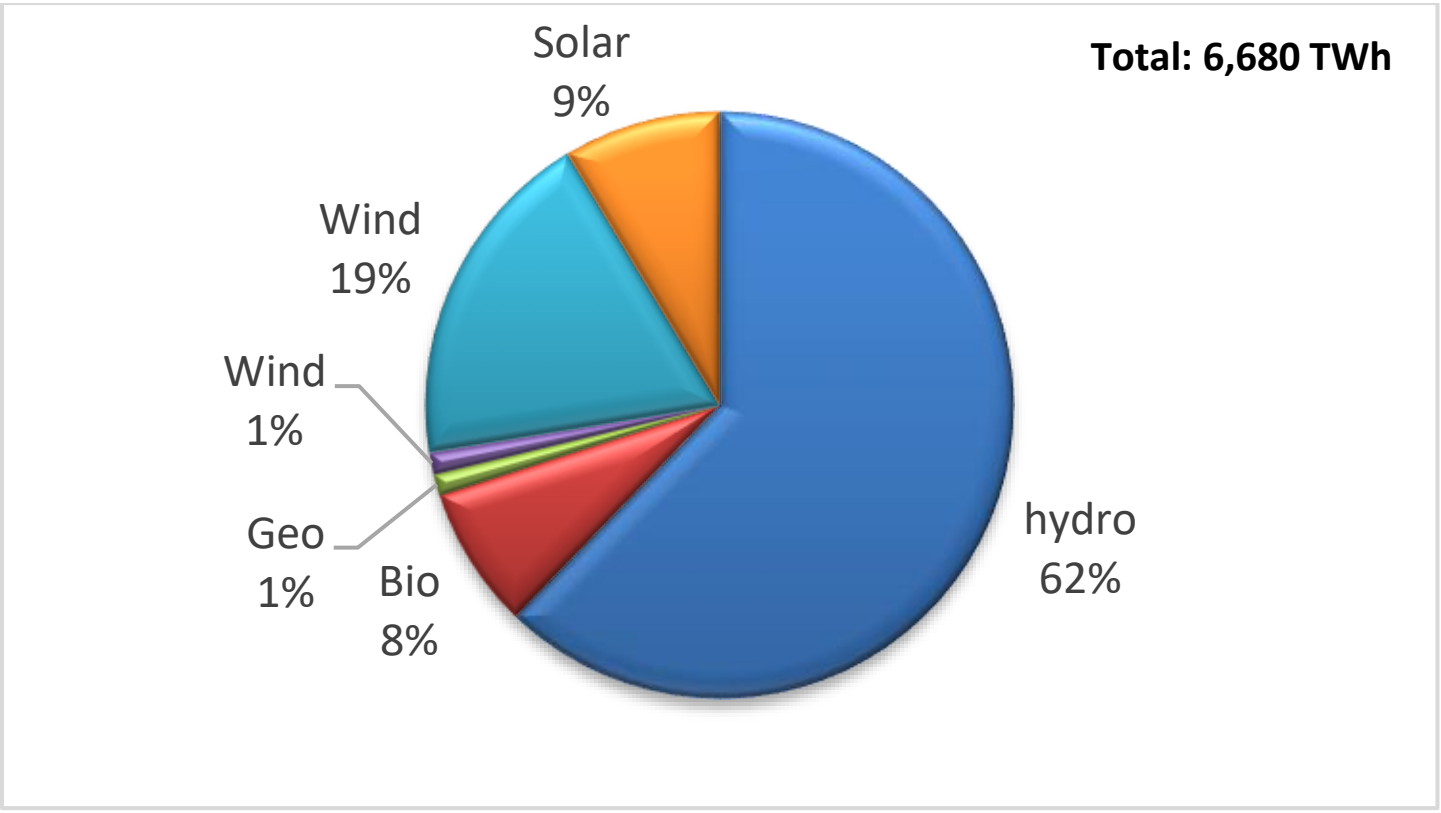

Figure 3. Renewable Energy Mix in 2018

Figure 3 depicts the renewable energy mix worldwide [24]. Hydroelectric generation contributes most in the mix due to the established technology and abundance of water in most parts of the world added with the geographic diversity. However, other sources such as- geothermal, wind, and solar are also contributing significantly. Both wind and solar technologies are still maturing as they are extremely intermittent and not uniformly distributed worldwide. The most important aspect of renewable energy is zero/almost zero-emission. Therefore, the research community is working to increase solar panel and wind turbine efficiency while making sure the environmental aspect is the priority. Again, aggregating these widely distributed and variable sources is another technological challenge, and a possible solution is smart grid technology.

#### 1.1.2.4 Future Trend

There is a global shift to renewable energy sources to limit the carbon emission increment to 9.7 gigatons by 2050 (projected 35 gigatons upsurge) [31]. This attempt is aligned with the < 2 °C target cap temperature rises of the Paris Agreement [32, 33]. The projected electricity generation in 2050 is 42,000 TWh (double of 2015 global demand) [34]. The goal is to meet 85% of this forecast from renewable energy sources, while the total primary energy supply will reach 63% from renewables. Another goal is to reduce the overall energy industry emission by 94% using renewable energy [4].

### 1.1.3 Renewable Energy Sources Grid Integration and Implications

As mentioned in the earlier sections, the intermittent nature of renewable energy sources is the core challenge when it comes to grid integration [35-37]. Therefore, the grid must be robust enough to interoperate among all connected renewable energy sources, regular loads, peak shaving generators, and large-scale power plants [38, 39]. Redesigning the grid control strategy for efficient demand-side management, energy management control, and intelligent resource optimization is thus very important [40-43].

### 1.1.4 Wind Speed Modeling: Importance and Research Scope

In this section, we review the aspects of wind speed modeling. The uncertain nature of wind speed is a barrier in optimal power generation and economic planning [44, 45]. Therefore, researchers are working on how to accommodate this uncertainty using predictive wind speed modeling to make the power system more resilient to this variability [46-49]. Again, another blockade is the quality of the available wind speed data we already have. Sometimes we have missing data in a sample. So, we need a technique to predict this missing data first for the optimal wind penetrated resilient power system. For wind speed prediction, missing data occur due to measurement error, malfunctioning sensors, and incorrect operation. A group of researchers used Gaussian process regression and multiple imputations approach in combination with the expectation-maximization algorithm to deal with this issue [50].

The Bayesian statistical approach is getting popularity in wind power modeling. Bayesian modeling represents a prospective solution to improve the accuracy and reliability of wind resource estimation and short-term forecasts [51]. Effective forecasting is the key to reduce economic and technical risk in wind speed applications. Again, the accuracy of the wind speed distribution is crucial. The accuracy is determined based on established statistical approaches known as goodness-of-fit criteria.

Rescaling and standardization are applied through standard normal CDF to ensure that all scores are comparable with each other. Then, each standardized score is subjected to multiplicative aggregation to determine the final scores. The distribution model with the minimal final score is suggested to be an accurate model for wind-speed data [52]. For accuracy, an important factor is the selection of distribution. Some of the distributions useful for wind speed modeling are Weibull, Rayleigh, Gamma, Inverse Gaussian, Log-Normal, Gumbel, Generalized extreme value, Nakagami, and Generalized Logistic distribution. However, it is important to determine the suitable one that gives the best distribution fit for a location of interest and describes the wind speed characteristics well [53]. The sample size is another important factor for fitting the above distributions. The Bayesian approach is useful to model while wind speed, while there is a lack of complete or partial data, is available [54].

Another research question regarding accurate wind speed prediction has been bouncing around in the research community, as listed in [55-64]. There are two different directions of the wind speed prediction: 1) time series forecasting of wind speed from a larger historical dataset (predicts possible future data points), and 2) wind speed prediction for a given time when few other meteorological parameters are known except the wind speed. The first one has been addressed for a long time using various approaches [65, 66]. The researchers have been using various approaches for later [60]; however, there is still a lack of comparative analysis between popular techniques such as statistical, machine learning, and deep learning models.

The wind speed distribution of a single location has been explored in various approaches. However, the joint distribution modeling of multiple wind speed locations has not been explored extensively and is getting attention. The copula is a noble mathematical approach to find the dependency between two (or more) random variables [67]. It considers the marginal distribution of each variable and combines them to a joint distribution by formalizing the dependency structure [68]. Therefore, the copula is highly applicable in wind speed modeling and investment decisions [69]. Another research shows a Vine-copula technique to reject outliers application to wind turbine power curve [70]. A reliability model is presented in [71] that shows the system reliability is inversely related to wind speed dependence. Similar research [72] used Copula-ARMA and concluded with a supporting result to [71]. One of the two available work [73] reports, the mathematical complexity is impractical for multivariate dependent joint wind speed distribution modeling for more than two-speed variables. This research examines a bivariate wind speed modeling analytically that requires further experimental verification. Another work [74] investigated Gaussian and Gumbel Copulas for a set of wind speed pair locating in different distances ranging from 22 to 2633 km. However, this unique work lacks in taking a methodical approach for copula family selection.

## 1.2 Aims and Objectives

After an extensive literature survey on wind speed prediction researches, we became interested in two major areas to explore in this research. Firstly, we would like to see how different algorithms perform for wind speed prediction from mereological data. Later, we would like to address a standard procedure for copula family selection to find the wind speed dependency (of two locations). The following summarizes the key objectives of this thesis-

- Wind speed prediction from meteorological data for a single location.
- Copula selection for the dependency of a set of distant wind speed pairs.
- Case study for copula modeling and analyze the joint distribution accuracy.
- Discuss the aspects of copula model uncertainty for wind speed dependency.

# Chapter Two: Wind Speed Prediction of a Single Location

## 2.1 Background

The placement of a wind turbine for wind power generation is often a challenging step due to the varying nature of wind speed from a location/height to another location/height [75]. Measuring wind speed at the level of turbine hub height is both expensive and requires continuous maintenance. Again, meteorological parameters play a vital role in the wind characteristics. Therefore, wind speed profiling with the variation of meteorological parameters has been a research problem that leads to the prediction of wind speed of a certain location based on those parameters [76, 77]. Therefore, utilization of easy to access parameters in a low elevation to predict corresponding wind speed at a higher height is a practical approach.

Literature show there have been statistical approaches for wind speed application, while artificial intelligence- deep learning and machine learning are being considered recently [78, 79]. In addition to machine learning regression algorithms, neural network-based deep learning techniques are also getting attention for similar problems due to higher accuracy. Long short-term memory (LSTM) is a form of Recurrent Neural Network (RNN) that is capable of learning long-term dependencies to make a prediction differentiates itself from other neural network methods such as Deep Neural Network (DNN) and Convolutional Neural Network (CNN) [80].

## 2.2 Site Description and Data Preprocessing

### 2.2.1 Dataset

We collected data from the National Renewable Energy Laboratory (NREL) database available online [81]. The dataset considered for this research contains samples of the three-months-long period starting from May 1, 2018, to July 31, 2018. The raw data entails samples of each minute. It was converted to average hourly instances. Primarily, the dataset had eighteen features, among which wind speed in 80m height is our response variable, and other 17 are predictors- solar radiation [listed as global PSP (Precision Spectral Pyranometer)], temperature (2m), estimated sea-level pressure, average wind speed (2m), average wind direction (2m), average wind shear, turbulence intensity, friction velocity, wind chill temperature, dew point temperature, relative humidity, specific humidity, station pressure, average wind speed (5m), accumulated precipitation, atmospheric electric field, and estimated surface roughness. Instances inside “( )” represents the height where the parameter was measured, ‘m’ stands for meters.

### 2.2.2 Preprocessing

This section describes the theoretical approach taken for this research, while Figure 4 illustrates a flow chart detailing every step. The following sections contain a detailed discussion of each of these elements.

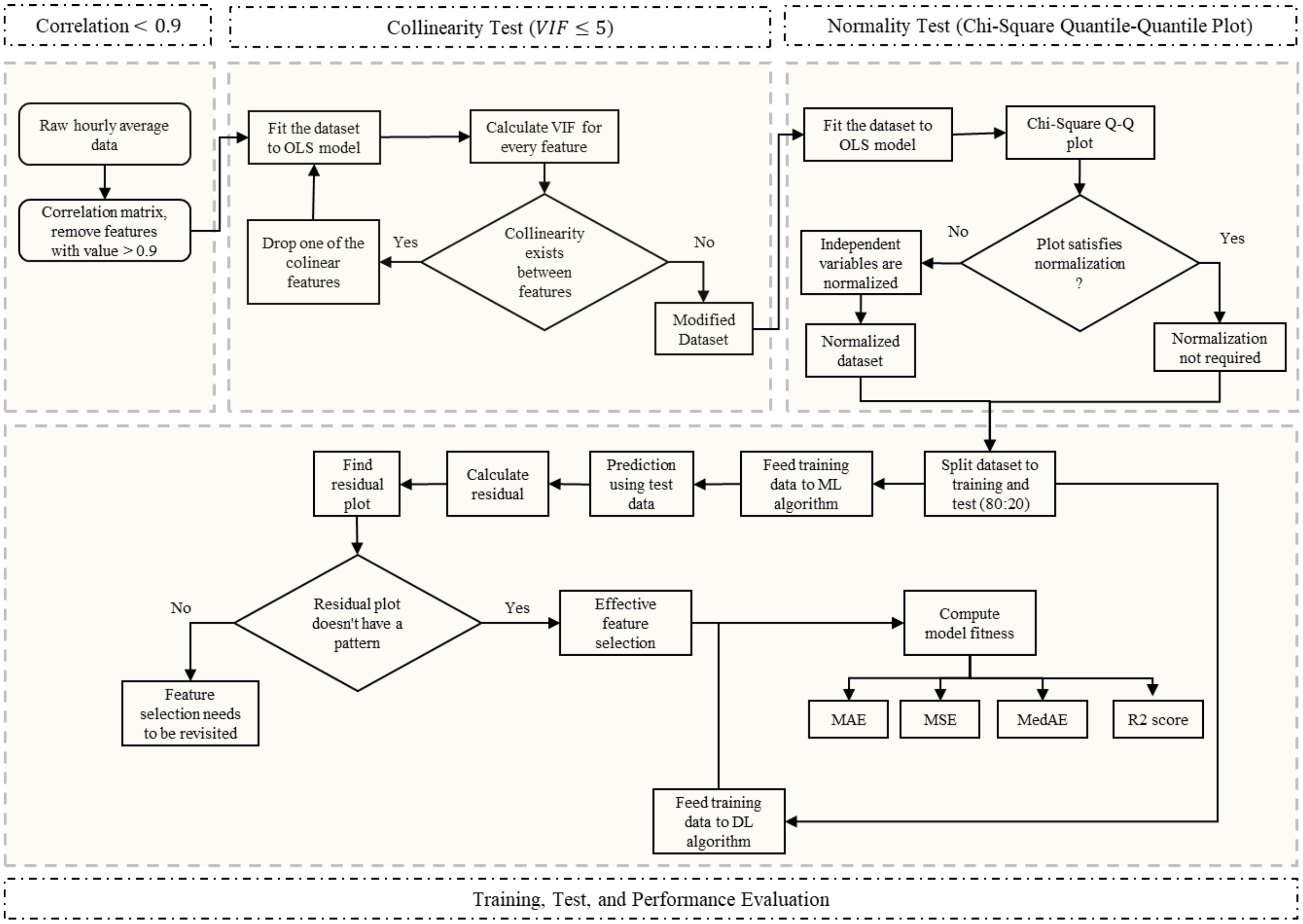

Figure 4. Flow Architecture of Wind Speed Prediction Steps

#### 2.2.2.1 Correlation Matrix

Correlation is a measure to identify how strongly a pair of variables ($X$, $Y$) are related. This score ranges between +1 and -1 based on their degree of relation. Pearson, Kendall, and Spearman are common correlation techniques. We have used Spearman correlation (1) as it does not carry any assumption about the distribution of the data (while Pearson's correlation assumes data is normally distributed, which is not always the case) [82]. We assume

$$\rho = \frac{cov(rg_X, rg_Y)}{\sigma_{rg_X}\sigma_{rg_Y}} \tag{1}$$

Where,

ρ: correlation coefficient

$rg_X$, $rg_Y$: ranks of X and Y

$cov$: covariance

$\sigma_{rg_X}\sigma_{rg_Y}$: the standard deviation of rank variables

The cross-correlation between the predictors (independent or explanatory variable, X) and the response (dependent variable, Y), as listed in Section 2.2.1, is the initial step toward feature selection. If the absolute value of a correlation measure is greater than 0 between two parameters, then they are considered correlated. Perfect correlation is one that usually indicates self-correlation or a problem called multicollinearity that leads to false predictions (more about multicollinearity in Section 2.2.2.2). Therefore, features with very high correlation (>0.9) are removed from the dataset as they might mislead the model performance [83, 84].

#### 2.2.2.2 Collinearity Test

Collinearity is a condition when some independent variables are highly correlated. In a regression model, if there are highly correlated predictors, they cannot independently predict the value of the dependent variable. In other words, they explain some of the same variances in the dependent variable, which causes inflated values of coefficients or even reverses the sign. If more than two predictors are associated in that manner, it is called multicollinearity [85].

Variance Inflation Factor (VIF) is applied to investigate if collinearity exists among the features [86]. For a multiple linear regression model with n predictors, VIFs are the diagonal elements of the inverse of the correlation matrix of the predictors. VIF for the *n*th predictor variable is expressed as (2).

$$VIF_n = \frac{1}{1-{R_n}^2} \qquad n = 1,2,3,\ldots \tag{2}$$

The smallest possible VIF is one representing not collinear at all and higher the value greater the collinearity. However, higher values of VIF (>5) represent the presence of multicollinearity, causing redundancy (more than one predictor describing the same effect on the response variable) [86]. Therefore, one from each collinear subset of predictors is kept, and others are removed from data [87, 88]. When applied to the wind speed dataset, two colinear pairs were found and treated: temperature (2m) & wind chill temperature, and average wind speeds (2m & 5m). Then wind chill temperature and average wind speed (5m) are removed from the dataset, and VIF applied again to make sure the absence of collinearity. The steps, as described here, are visualized in the top-right block of Figure 4. After considering correlation, and collinearity, the selected best features are- temperature (2m), estimated sea-level pressure, average wind speed (2m), average wind direction (2m), average wind shear, turbulence intensity, and friction velocity. The cross-correlations between each model parameters are listed in Table 1, where parameter numbered as '8' is our response, average wind speed at 80m height. After analyzing the table, it is clear that average wind speed at 2m height and average wind shear both are highly correlated (+) with the response variable. On the other hand, sea-level pressure and turbulence intensity are negatively correlated with the wind speed at 80m.

### 2.2.2.3 Normality Test

Normality tests are applied to investigate if the dataset is well modeled (likelihood of data to be normally distributed). Some approaches to test for normalization are- graphical method, back-of-the-envelope test, frequentist test, and Bayesian test [89]. In this research, we use the graphical test. In this method, the Chi-Square Quantile-Quantile (Q-Q) plot of multivariate distribution is analyzed to see if the features are normally distributed [90]. If normal, the plot should follow the 45-degree baseline. If not, then normalization is required before fitting the data to any model. Figure 5 shows the Q-Q plot that tells the dataset is not normalized. Thus, prior normalization is required before feeding it to the models.

Table 1. Cross-correlation of model parameters

| | Parameters [unit] | 1 | 2 | 3 | 4 | 5 | 6 | 7 | 8 |
|---|---|---|---|---|---|---|---|---|---|
| 1 | Temperature (2m) [deg C] | 1.0 | -0.02 | 0.24 | -0.22 | 0.12 | 0.42 | 0.05 | 0.16 |
| 2 | Sea-Level Pressure (Est) [mBar] | -0.02 | 1.0 | -0.13 | -0.05 | -0.09 | 0.03 | -0.07 | -0.13 |
| 3 | Avg Wind Speed (2m) [m/s] | 0.24 | -0.13 | 1.0 | 0.08 | 0.64 | 0.07 | 0.23 | 0.87 |
| 4 | Avg Wind Direction (2m) [deg] | -0.22 | -0.05 | 0.08 | 1.0 | 0.26 | -0.52 | 0.19 | 0.25 |
| 5 | Avg Wind Shear [1/s] | 0.12 | -0.09 | 0.64 | 0.26 | 1.0 | -0.07 | 0.52 | 0.88 |
| 6 | Turbulence Intensity (2m) | 0.42 | 0.03 | 0.07 | -0.52 | -0.07 | 1.0 | -0.01 | -0.03 |
| 7 | Friction Velocity (u*) [m/s] | 0.05 | -0.07 | 0.23 | 0.19 | 0.52 | -0.01 | 1.0 | 0.41 |
| 8 | Avg Wind Speed (80m) [m/s] | 0.16 | -0.13 | 0.87 | 0.25 | 0.88 | -0.03 | 0.41 | 1.0 |

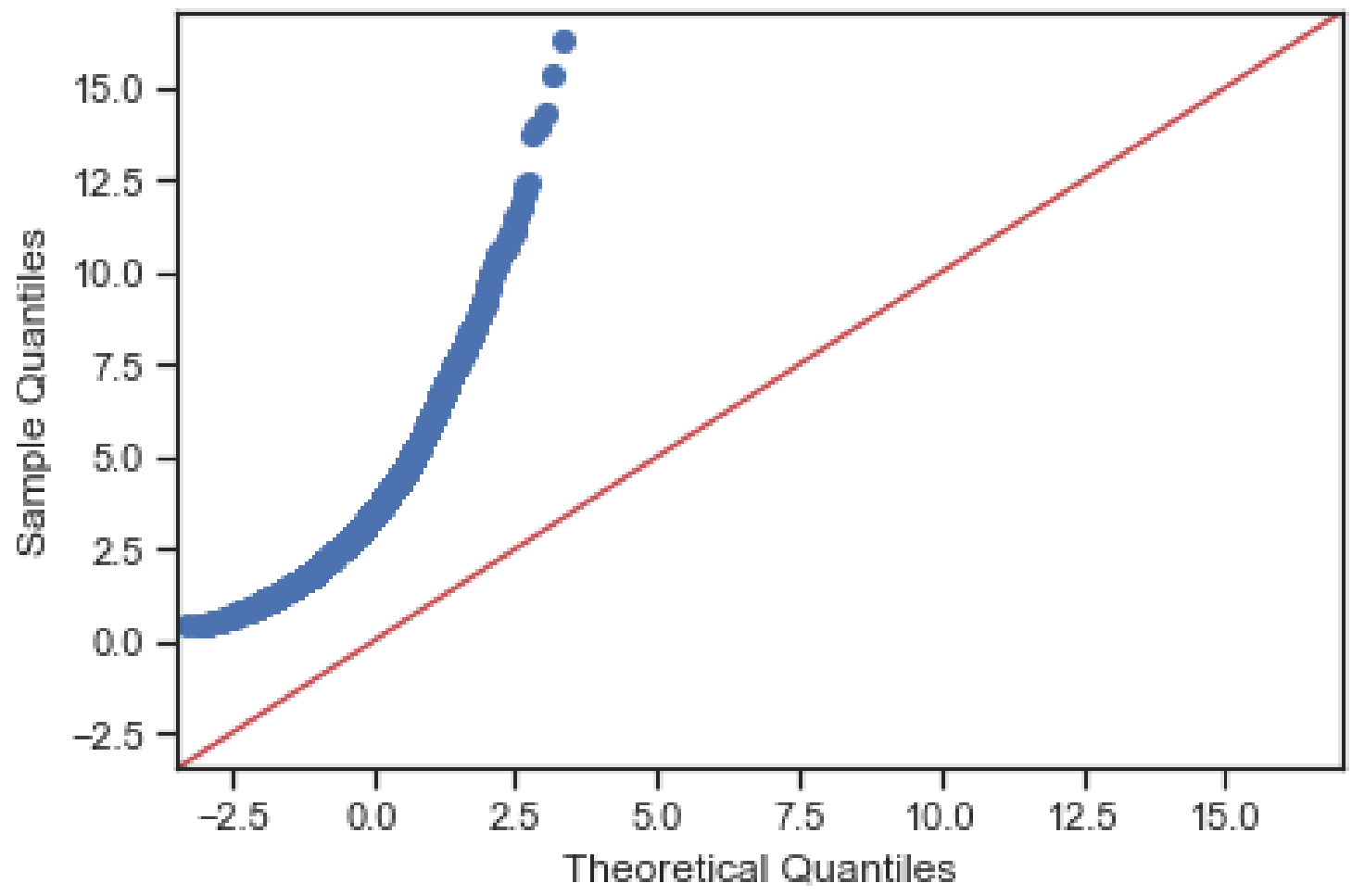

Figure 5. Chi-Squared Q-Q Plot

### 2.2.2.4 Train-Test Split

The prediction algorithms are trained using a certain dataset. However, the performance of a model depends on how well it can predict the response variable when encounters unknown predictors.

Therefore, the dataset is usually divided between two sets, also known as training and test sets. The training dataset is then used to train the prediction algorithm while the test set is allocated to use them as an unknown predictor to analyze the model performance. The ratio of allocating data for training and test is randomly selected, but literature shows 70~80% for training, and 20~30% for the test is common practice [91-93]. In this research, we have separated 80% of the total data to train the models and rest 20% to test the model performance.

## 2.3 Prediction Algorithms

Wind speed prediction is a regression problem where predictors, in this case, are the meteorological parameters, and the response variable is wind speed at 80m height. In general, regression is a classical problem both in statistics and machine learning. Usually, statistical methods are to find the inference while machine learning makes the prediction [94]. Then, again, they intersect in some cases and do the same thing, for example- linear regression. However, statistical learning relies more on distributions, while machine learning is an empirical process and requires data [95]. The statistical approach, thus, considers how data is collected or generated while machine learning may result in accurate prediction without knowing much about the underlying aspects of data. Another line of the boundary is the shape or volume of data. While the statistical approach is very robust about the number of samples (as it considers the distribution of the data), machine learning is more applicable when the dataset is wide [95]. However, sometimes they are used interchangeably, and statistics are the backbone of machine learning [96]. Again some machine learning algorithms use the same bootstrapping methods as statistical models [96]. Besides, researchers are also using deep learning for similar prediction problems [97]. Artificial Neural network (ANN) based models usually yield benefits in prediction tasks than statistical models due to its robustness towards the nature of data, especially when there are missing values or the dataset is nor well preprocessed, raw, and large data [98]. Thus, many machine learning regression algorithms use statistical techniques in innovative ways, while deep learning neural network approaches are also efficient for a similar task. The reason, however, behind the growing popularity of machine learning or deep learning (artificial intelligence, in general) is the availability of computational resources [99]. Therefore, the larger dataset is not a critical issue to work with, which was challenging in the past. In this research, we have considered both approaches, machine learning, and deep learning for our wind speed prediction problem.

### 2.3.1 Machine Learning Approach

Machine learning regression algorithms are widely used for prediction applications while they vary in terms of behind the scene mathematics [100]. The following are descriptions of the algorithms used in this research.

#### 2.3.1.1 Multiple Linear Regression

Multiple linear regression relates two or more dependent variables to an independent variable by fitting a linear equation to the data [101-103]. For $n$ independent variables ($x_1, x_2, \dots, x_n \in \mathbb{R}$) and $m$ observations, the multiple linear regression model (3) is defined as-

$$\hat{y}_i = \beta_0 + \beta_1 x_{i1} + \beta_2 x_{i2} + \cdots + \beta_n x_{in} + \varepsilon_i \quad (i = 1, 2, \dots, m) \tag{3}$$

Where,

$y_i \in \mathbb{R}$: dependent variable, $\hat{y}_i$ is the estimate of $y_i$

$\varepsilon$: residual (deviation of $\hat{y}_i$ from its mean value)

$\beta$: regressor coefficients estimated from least-square estimates ($\beta_0$ is known as intercept, and $\beta_n$ is the slope of the regression line for simple linear regression)

$m$: number of row in the dataset

The sum of squares of residuals ($SS_{residuals}$) of multiple linear regression is shown in (4), and we would like to minimize this value in the model. The equation for multiple linear regression can also be written as (5), where *argmin* stands for the argument of minimum. In other words, it finds the β's that minimize $SS_{residuals}$. β's are easy to find from (6), which is also known as the matrix formulation of linear regression, where *X* and *Y* represent independent and dependent variables, respectively.

$$SS_{residuals} = \sum_{i=1}^{m} (\hat{y}_i - y_i)^2 \tag{4}$$

$$\hat{\beta}^{MLR} = \underset{\beta \epsilon \mathbb{R}}{argmin} \sum_{i=1}^{n} [y_i - \hat{y}_i]^2 = \underset{\beta \epsilon \mathbb{R}}{argmin} \sum_{i=1}^{n} [y_i - (\beta_o + \beta_1 x_{i1} + \beta_2 x_{i2} + \cdots + \beta_n x_{in})]^2 \tag{5}$$

$$\hat{\beta}^{MLR} = (X^T X)^{-1} X^T Y \tag{6}$$

### 2.3.1.2 Ridge Regression

Ridge regression is a variant of linear regression that uses L2 regularization [104]. The fundamental equation for ridge regression is the same as linear regression with a constraint on it (7), where *C* defines the boundaries of ridge regression. The regularization shrinks the parameters and reduces the model complexity by a coefficient of shrinkage. This coefficient is known as the penalty and denoted by $\lambda$ (hyperparameter). Now we look at the equation for ridge regression and it is clear that the first part of (8) is the same as linear regression, and the true difference between linear and ridge is the second term containing the constraint, *B*, shown in (9) and penalty. Due to the inclusion of the penalty, the residual error is minimized, therefore, ridge regression shall produce better accuracy.

$$\beta_0{}^2 + \beta_1{}^2 + \cdots + \beta_n{}^2 \leq C^2 \tag{7}$$

$$\hat{\beta}^{ridge} = \underset{\beta\epsilon\mathbb{R}}{argmin} \sum_{i=1}^{n} [y_i - \hat{y}_i]^2 = \underset{\beta\epsilon\mathbb{R}}{argmin}\, min(\left|\left|y_i - XB\right|\right|_2^2 + \lambda\left|\left|B\right|\right|_2^2) \tag{8}$$

$$\|B\|_2 = \sqrt{\beta_0{}^2 + \beta_1{}^2 + \cdots + \beta_n{}^2} \tag{9}$$

### 2.3.1.3 Least Absolute Shrinkage and Selection Operator (Lasso) Regression

Lasso regression uses L1 regularization. It has benefits over ridge regression when there are more features [91]. The equation of lasso regression (10) is quite straight forward from the ridge regression, the only difference is the second term of the equation.

$$\hat{\beta}^{lasso} = \underset{\beta\epsilon\mathbb{R}}{argmin} \sum_{i=1}^{n} [y_i - \hat{y}_i]^2 = \underset{\beta\epsilon\mathbb{R}}{argmin}\, min(\left|\left|y_i - XB\right|\right|_2^2 + \lambda\|\beta\|_1) \tag{10}$$

Adding regularization is a very important technique in machine learning to prevent overfitting [105]. The difference between the L1 and L2 is just that L2 is the sum of the square of the weights, while L1 is just the sum of the weights

### 2.3.1.4 Bayesian Ridge Regression

Bayesian view of ridge regression is obtained by noting that the minimizer of (8) can be considered as the posterior mean of a model where $\beta_i \sim N\left(0, \frac{\sigma^2}{\lambda}\right)$, for $i = 1,2, \ldots, n$ [106].

### 2.3.1.5 Huber Regression

In the ridge and lasso, the penalty was a hyperparameter. Instead of considering an estimated constant, Huber function (11) is proposed. This method does not rely on $SS_{residuals}$ to minimize the error, rather sensitive to outliers ($|y_i - \hat{y}_i| \leq \delta$). Thus, the name, Huber regression [107], came from the loss function is uses. δ is a hyperparameter, and the choice is critical. Residuals larger than δ are minimized with L1 (which is less sensitive to large outliers), while residuals smaller than δ are minimized appropriately with L2.

$$L_\delta(y_i, \hat{y}_i) = \begin{cases} \frac{1}{2}(y_i - \hat{y}_i)^2 & for\ |y_i - \hat{y}_i| \leq \delta \\ \delta|y_i - \hat{y}_i| - \frac{1}{2}\delta^2 & otherwise \end{cases} \tag{11}$$

### 2.3.1.6 Bagging Regression

Bootstrap aggregating (bagging) prediction models is a general method for fitting multiple versions of a prediction model and then combining them into an aggregated prediction [108, 109]. It is a straight forward algorithm in which *b* bootstrap copies of the original training data are created, the regression is applied to each bootstrap sample, and in the regression context, new predictions are made by averaging the predictions together from the individual base learners. Equation (12) thus demonstrates the formulation by letting $x_i$ as the prior and $\hat{y}_{bag}$ as the bagged prediction. $\hat{y}_{i1}, \hat{y}_{i2}, \ldots, \hat{y}_{ib}$ are the the predictions from individual base learners for $x_i$.

$$\hat{y}_{bag} = \hat{y}_{i1} + \hat{y}_{i2} + \cdots + \hat{y}_{ib} \tag{12}$$

### 2.3.1.7 Random Forest Regression

Random forest regression is also a bootstrap aggregation (bagging) that involves training each decision tree on a different data sample where sampling is done with replacement [109, 110]. Thus, it combines multiple decision trees in determining the model output. A visual illustration of such a model is in Figure 6.

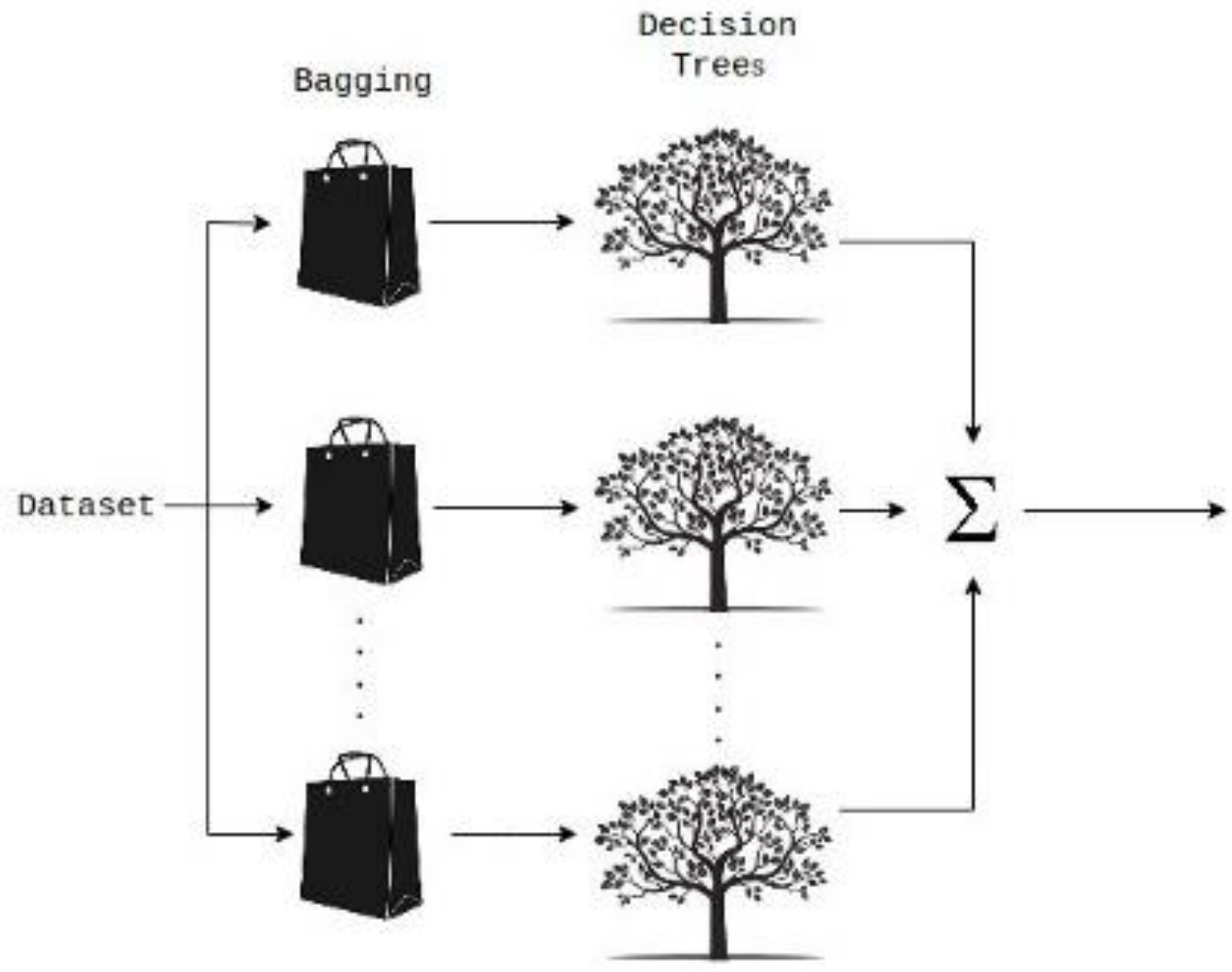

Figure 6. Visual aid operation of random forest regression

### 2.3.1.8 Adaptive Boosting (AdaBoost) Regression

AdaBoost is stagewise estimation procedures to fit an additive logistic regression model, which minimizes the exponential loss function [111]. The operational steps of this algorithm starts with initializing the observation weights $\beta_i$ on the original sample (predictors). The classifier then adjusts the weights for residual minimization and re-iterate the process for a defined number of times. Finally, the model becomes an aggregation of classifier function, $F_i(x)$, and weight minimizer, $\alpha_i$, as shown in Figure 7 [112].

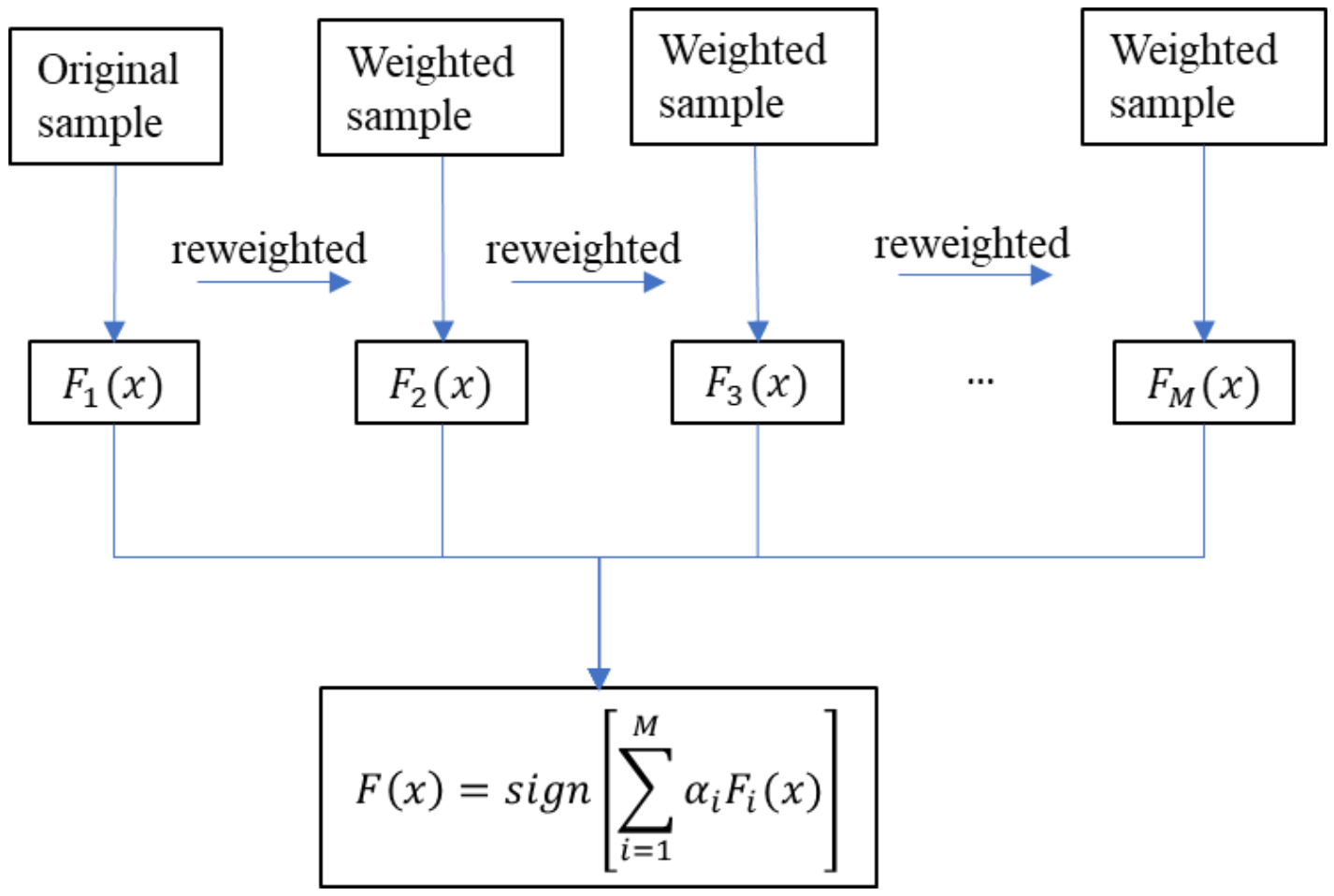

Figure 7. Learning Steps in AdaBoost Regression

### 2.3.1.9 Support Vector Regression (SVR)

Support vector regression is a robust algorithm that can handle both linear and non-linear input regression problems [113]. We consider a linear case, where predictors, $x$ and response $\hat{y_i}$. The formulation for a linear case is shown in (13) where $f_i(x)$ is the transfer function (also known as the kernel), and b is bias. A few of the common transfer functions are linear, non-linear, polynomial, and radial basis functions.

$$\hat{y_i} = \sum_{i=1}^{n} \beta_i f_i(x) + b \tag{13}$$

### 2.3.2 Deep Learning Approach

Deep learning algorithms are now applied to solve problems of a diverse nature, including prediction [114]. Therefore, we are also considering deep learning algorithms for this research. Firstly, we would like to review a few basics of deep learning.

The building blocks of deep learning or artificial neural networks are called perceptron, which mimics a similar functionality (in computation) as neuron (a biological cell of the nervous system that uniquely communicates with each other) [115]. Neurons are consist of dendrite, axon, and cell body, as shown in Figure 8*Figure 8* [116]. Dendrites receive a nervous signal and transmit it to the cell body, which processes the stimuli and decides whether to transmit it to the other neurons (via axon) in the form of chemical transmission.

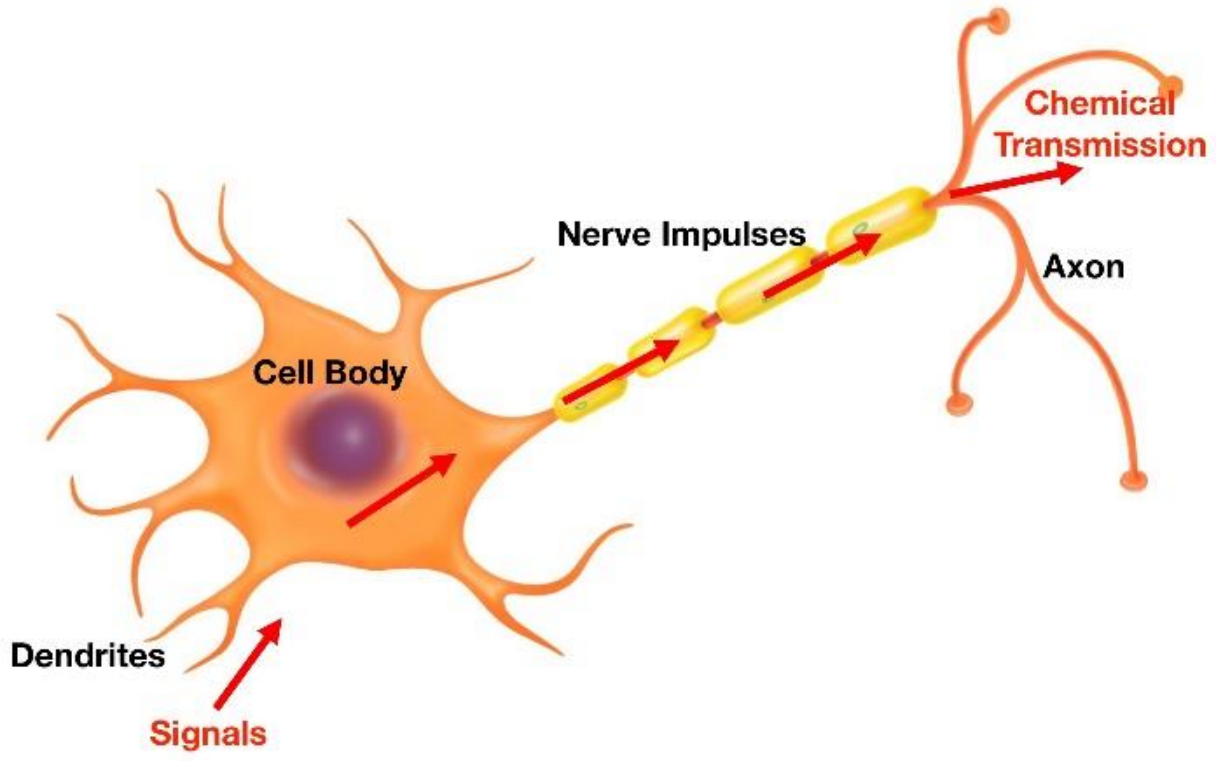

Figure 8. Visual Illustration of Biological Neuron

Now, perceptrons or artificial neurons (as shown in Figure 9 [116]), receive input signals $(x_1, x_2, \ldots\ x_m)$, multiply each input by a weight $(w_1, w_2, \ldots\ w_m)$, add them together with a pre-determined *bias* and pass through the activation function, $f(x)$. The signal the goes to output as 0 or 1 based on the activation function threshold value. A perceptron with inputs, weights, summation and bias, activation function, and output are all together form a single layer perceptron [117]. However, in common neural network diagrams, only input and output layers are shown. In a practical neural network, hidden layers are added between the input and output layers. The number of hidden layers is a hyperparameter and usually determines by trial and error and looking at the model performance. If the neural network has a single hidden layer, the model is called a shallow neural network, while a deep neural network (DNN) consists of several hidden layers [115]. In this research, we have considered DNN, convolutional neural network (CNN), and recurrent neural network (RNN)- in the form of long short-term memory (LSTM), all of which will be discussed in the following sections.

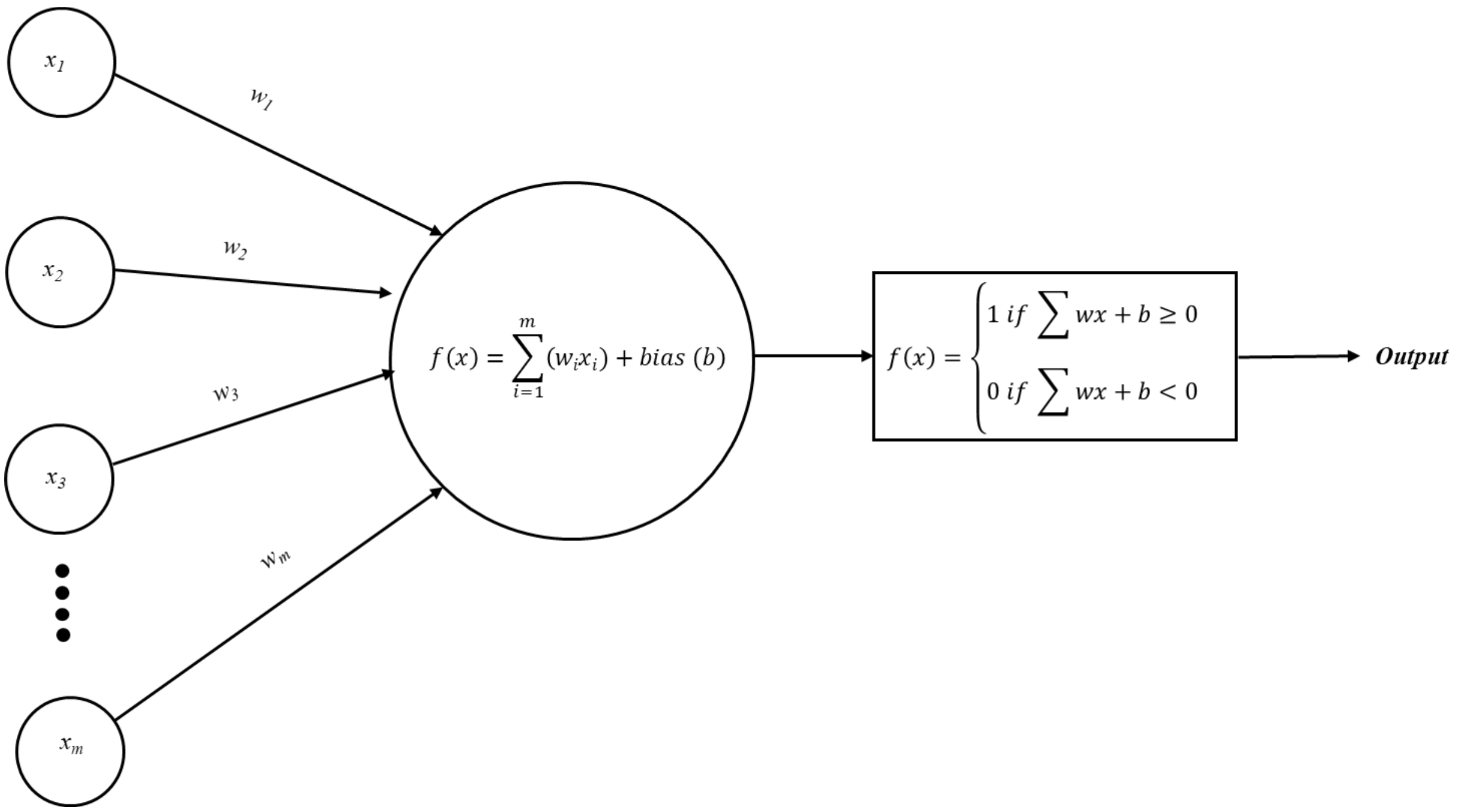

Figure 9. A Simple Artificial Neuron/ Perceptron

### 2.3.2.1 Deep Neural Network (DNN)

DNN is composed of three neural network layers, namely- an input layer, hidden layer(s), and an output layer. The number of hidden layers is a hyperparameter; a suitable value is found through some trial and error [115]. Figure 10 illustrates such a model structure with two hidden layers consisting of three neurons each, five input neurons, and one output neuron. The number of neurons depends on the number of inputs and outputs.

In Figure 10,

Inputs: [$x_1$, $x_2$, $x_3$, $x_4$, $x_5$]

Hidden layer weights: *h*

Output: $\hat{y}$

A simplified DNN kernel is formulated in (14) that considers linear modeling. *x*, *W*, and *c* symbolize input, weights, and bias, respectively, while *w* and *b* are linear model parameters. The hidden layer parameter *h* is shown in (15), where g is the activation function. For DNN modeling, ReLu (16) is used as the hidden layer activation function.

$$f(x; W, c, w, b) = w^T max\{0, W^T + c\} + b \quad (14)$$

$$h = g(W^T x + c) \quad (15)$$

$$f(x) = max(0, x) \quad (16)$$

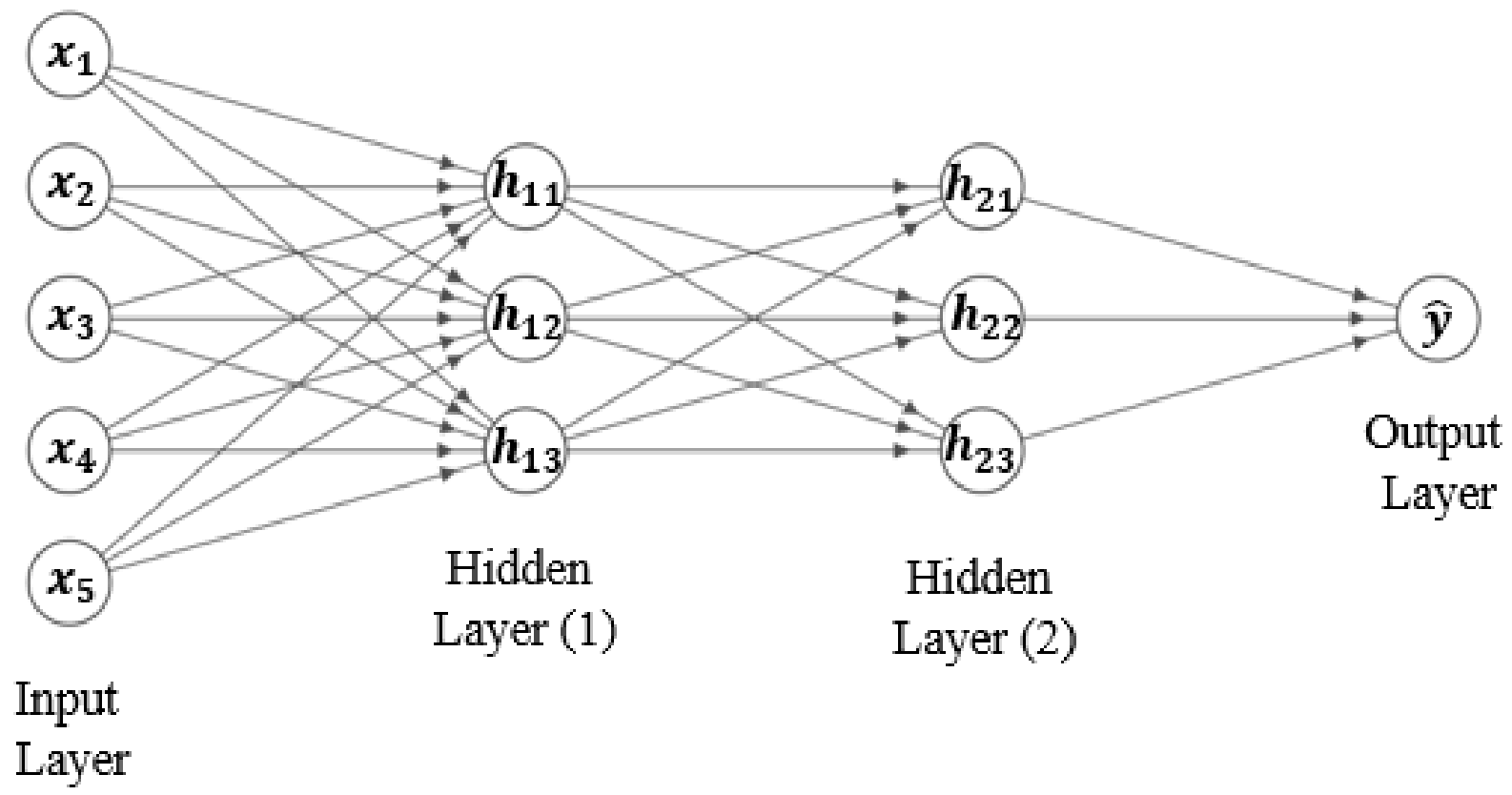

Figure 10. Simplified Architecture of a DNN

### 2.3.2.2 Convolutional Neural Network (CNN)

In this research, CNN, also known as ConvNet, is one way to solve the issue with DNN using convolution rather than matrix multiplication [118]. In other words, CNN is the regularized version of DNN to ensure model robustness towards overfitting. CNN is very popular for image processing; however, in the prediction problem, it is also utilized [119]. In this research, we are using 1D CNN for wind speed prediction. The characteristics and approaches are the same for all CNNs, regardless of dimensionality [120]. The architecture of CNN (Figure 11 shows for 1D CNN) consists of a convolution layer, pooling layer, and a fully connected neural network layer, thus, incorporates local receptive fields to ponder the spatial information, shared weights, and pooling to consider the summary statistics in the output.

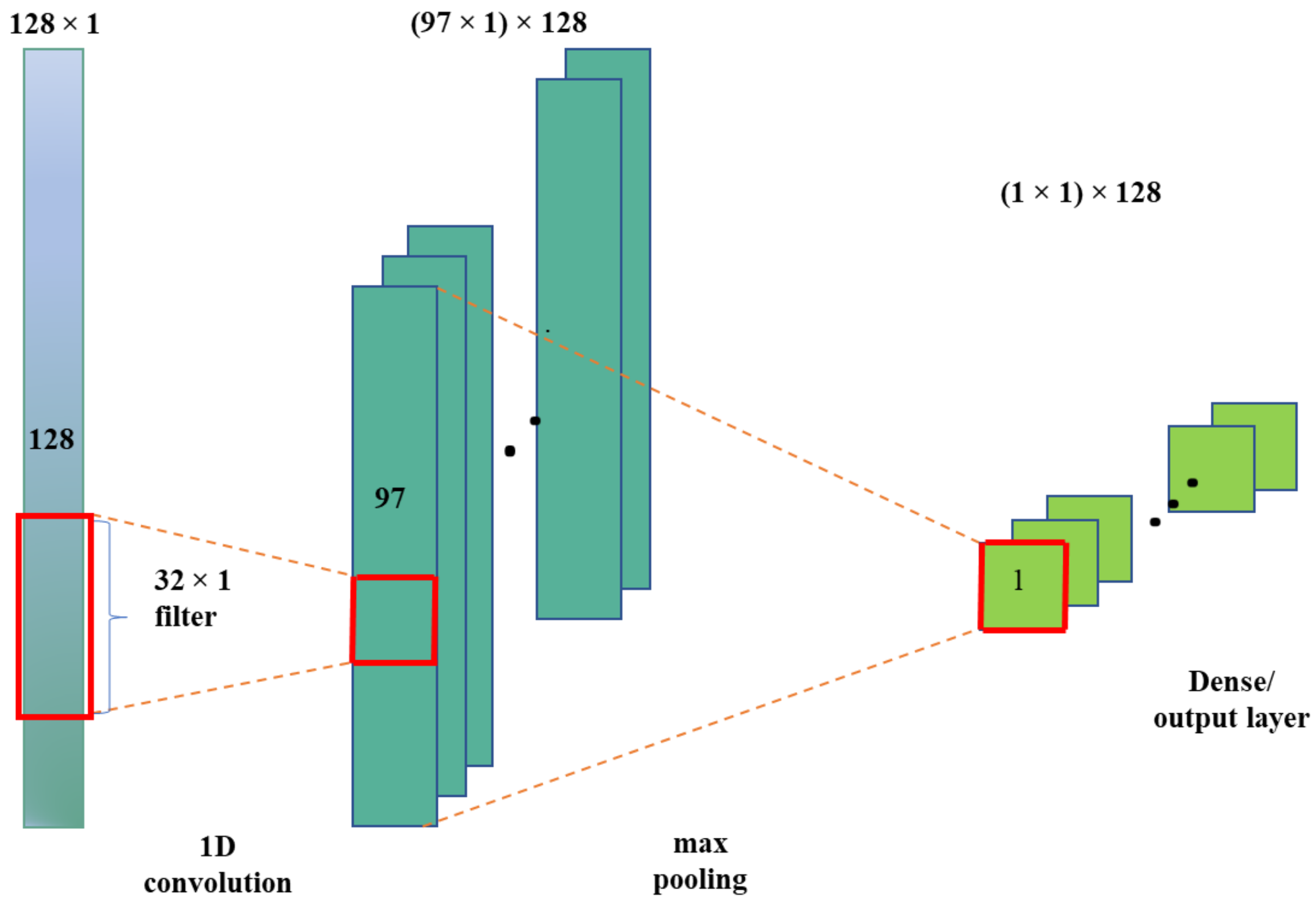

Figure 11. The architecture of 1D Convolution Neural Network

#### 2.3.2.3 Recurrent Neural Network (RNN) - LSTM

In this research, Long Short-Term Memory Networks (LSTM), a form of gated RNN, is proposed to implement. LSTM introduces self-loops to produce paths where the gradient can flow for a long duration; thus, it is capable of learning long-term dependencies [115]. LSTMs are explicitly designed to avoid the long-term dependency problem, as illustrated in Figure 12. The equations describing the operations are listed below.

$$f(t) = \sigma_g\left(W_f x_t + U_f h_{t-1} + b_f\right) \tag{17}$$

$$i_t = \sigma_g(W_i x_t + U_i h_{t-1} + b_i) \tag{18}$$

$$o_t = \sigma_g(W_o x_t + U_o h_{t-1} + b_o) \tag{19}$$

$$c_t = f_t \circ c_{t-1} + i_t \circ \sigma_c(W_c x_t + U_c h_{t-1} + b_c) \tag{20}$$

$$h_t = o_t \circ \sigma_h(c_t) \tag{21}$$

Where,

$x_t \epsilon \mathbb{R}^d$: Input vector to the LSTM unit

$f_t \epsilon \mathbb{R}^h$: Forget states activation vector

$i_t \epsilon \mathbb{R}^h$: Input/update gate's activation vector

$o_t \epsilon \mathbb{R}^h$: Output gate's activation vector

$h_t \epsilon \mathbb{R}^h$: Hidden state vector

$c_t \epsilon \mathbb{R}^h$: Cell state vector

$W \epsilon \mathbb{R}^{h \times d}, U \epsilon \mathbb{R}^{h \times h}, b \epsilon \mathbb{R}^h$: Weight matrices and bias vector parameters which need to be learned during the training

$\sigma_g$: Sigmoid function

$\sigma_c, \sigma_g$: hyperbolic tangent function

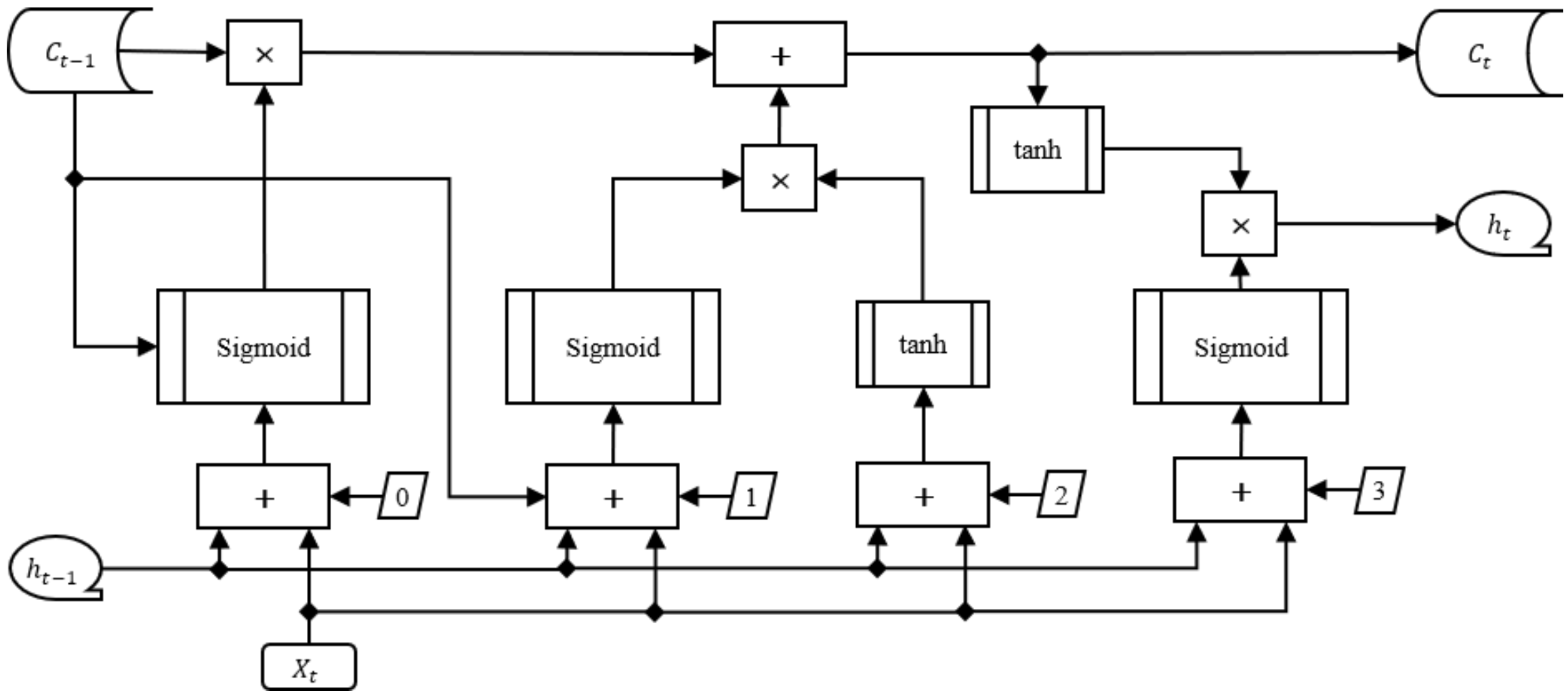

Figure 12. Block Diagram of LSTM Operations

## 2.4 Performance Evaluation

Some commonly used accuracy parameters are employed to evaluate how well a model is performing to predict the intended parameter [121]. Mean absolute error (*MAE*), mean square error (*MSE*), median absolute error (*MedAE*), and R-square *(R2)* scores are considered to investigate the model performances on the test set.

*MAE* is the average of the absolute values of the error (the difference between actual response ($y_i$) and predicted response ($\hat{y_i}$). As described by (22), n is the number of total input sets. The lower this value is, the better the model performance, while the desired is 0.

$$MAE = \frac{\sum_{i=1}^{n} |y_i - \hat{y_i}|}{n} \quad (22)$$

*MSE* is the mean of the square of error terms. Similarly, to *MAE*, it is desired to have 0 or close value for this term. The formula for this measure is in (23).

$$MSE = \frac{\sum_{i=1}^{n} (y_i - \hat{y_i})^2}{n} \quad (23)$$

*MedAE* is the median of all the error terms, defined in (24), thus effective to deal with outliers' effect in the model performance.

$$MedAE = median\,(|y_1 - \widehat{y_1}|, |y_2 - \widehat{y_2}|, |y_3 - \widehat{y_3}|, \dots \dots, |y_n - \widehat{y_n}|) \quad (24)$$

*R2 score* determines how well the model would perform in predicting the response variable as shown in (25) where $\bar{y_i}$ denotes the mean value of all predictions. This value is also known as the coefficients of determination. The best possible value is 1 for this case, and the closer to 1, the better model prediction is.

$$R2\ score = 1 - \frac{\sum_{i=1}^{n} (y_i - \hat{y_i})^2}{\sum_{i=1}^{n} (y_i - \bar{y_i})^2} \quad (25)$$

Further feature fit is tested using the residual plot by graphing the residual (the difference between prediction and actual value) vs. fitted instance.

## 2.5 Simulations and Results

In this section, we will discuss the simulation and performances of the prediction algorithms, as described in Section 2.3, for wind speed prediction in 80m height for the NREL dataset (Section 2.2.1). We listed the algorithms as Model-1 to 12 in Table 2 and fitted them on the training data (as described in Section 2.2.2.4) for learning. Once the training finished, we evaluated model performances according to the accuracy measures described in Section 2.4 on test data. We have also graphed the Q-Q plot (actual value vs. predicted value of wind speed at 80m) for all models.

For ridge regression, alpha was considered 0.01 after a few trial and errors. Similarly, for Lasso also, the alpha parameter was set to 0.01. For SVR, the default kernel initializer was applied. Table 2 depicts the accuracy measures for each algorithm. Overall, the considered algorithms were able to predict the average wind speed properly with an R2 value greater than 0.9 in most cases, as shown. Among the machine learning algorithms, MAE, MSE, and MedAE are minimum for bagging and random forest regression. Both algorithms also show greater accuracy (>96%). Figure 13 illustrates Q-Q plots and respective residual plots (below each Q-Q plot for the same model) for Model 1-9. There is a clear linear pattern in Q-Q plots (for all machine learning algorithms). That verifies the accuracy measures from Table 2, while again bagging and random forest regressions show fewer outliers. The residual plots, in contrast, do not show a linear pattern for any of the models, that supports their accuracy status from Table 2 and validates the feature selection in Section 2.2 [122]. On the other hand, Models 3-5 show the lowest accuracy among the machine learning regression algorithms with an R2 Score of $\approx 0.92$.

Deep learning models- DNN, CNN, and LSTM, are denoted as Model 10-12 in Table 2. Both DNN and CNN use ReLu activation function. DNN uses thirteen hidden layers, while the neural network of CNN consists of 50 neural network layers. Max pooling size for CNN is 2. On the other hand, LSTM uses a linear activation and consists of 50 hidden layers. In terms of accuracy and error parameters, CNN showed the worst performance, while both DNN and LSTM prediction accuracy were high (>96%). However, LSTM (Model-12) showed the best performance in terms of all metrics; thus, it showed the lowest error terms, while the exact accuracy was 97.8%.

Figure 14 illustrates the deep learning model performances. Prediction visualization, model loss, and MSE are plotted for the models and shown top to down for each model. All three models were run for 500 epochs; however, they reached to higher accuracies at around 100 epochs. Also, by observing the graphs, it is evident that CNN shows disperse prediction while LSTM is denser. Also, the graphs showing losses and MSE (per epoch) do not show any phenomenon of overfitting or underfitting.

Overall, we can see from Table 2, plots, and discussion until now, LSTM performed best for our investigation. Therefore, LSTM is the efficient learning algorithm between 12 test models to predict the wind speed at 80m height while temperature at 2m height, estimated sea-level pressure average wind speed at 2m height, average wind direction at 2m height, average wind shear, turbulence intensity at 2m height, and friction velocity of a certain location are known.

Table 2 Comparative Model Performances

| **Model** | **Algorithm** | **Mean Absolute Error (MAE)** | **Mean Squared Error (MSE)** | **Median Absolute Error (MedAE)** | **R2 Score** |
|---|---|---|---|---|---|
| Model-1 | Multiple linear regression | 0.421 | 0.357 | 0.277 | 0.923 |
| Model-2 | Ridge regression (alpha=15) | 0.579 | 0.598 | 0.434 | 0.872 |
| Model-3 | Least absolute shrinkage and selection operator (Lasso) regression (alpha=0.1) | 0.823 | 1.156 | 0.704 | 0.752 |
| Model-4 | Bayesian ridge regression | 0.428 | 0.361 | 0.285 | 0.922 |
| Model-5 | Hubber regression | 0.422 | 0.38 | 0.259 | 0.919 |
| Model-6 | Bagging regression | 0.274 | 0.171 | 0.185 | 0.963 |
| Model-7 | Random forest regression | 0.275 | 0.179 | 0.192 | 0.962 |
| Model-8 | Adaptive boosting (AdaBoost) regression | 0.385 | 0.272 | 0.297 | 0.942 |
| Model-9 | Support vector regression (SVR) | 0.411 | 0.347 | 0.261 | 0.926 |
| Model-10 | Multilayer perceptron (MLP)/ DNN (Hidden layer=13, activation relu) | 0.31 | 0.178 | 0.234 | 0.962 |
| Model-11 | CNN (Filters=64, kernel_size=2, activation='relu', Maxpooling (size=2), Flatten, Dense=50, activation=relu, epoch=500) | 0.634 | 0.831 | 0.45 | 0.82 |
| Model-12 | RNN – LSTM (50, kernel=normal, activation=linear) | 0.226 | 0.107 | 0.145 | 0.978 |

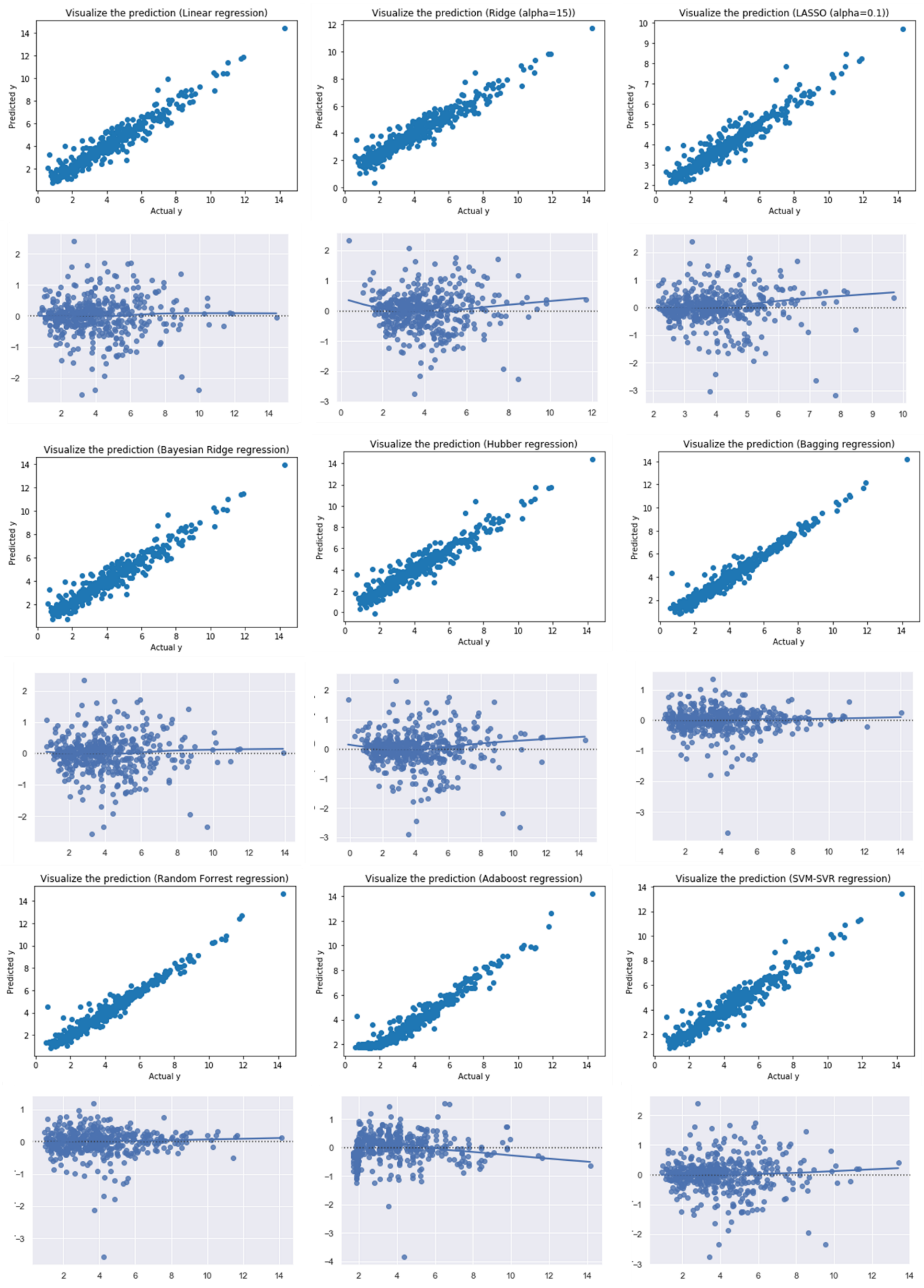

Figure 13. Model 1-9 Prediction Visualization

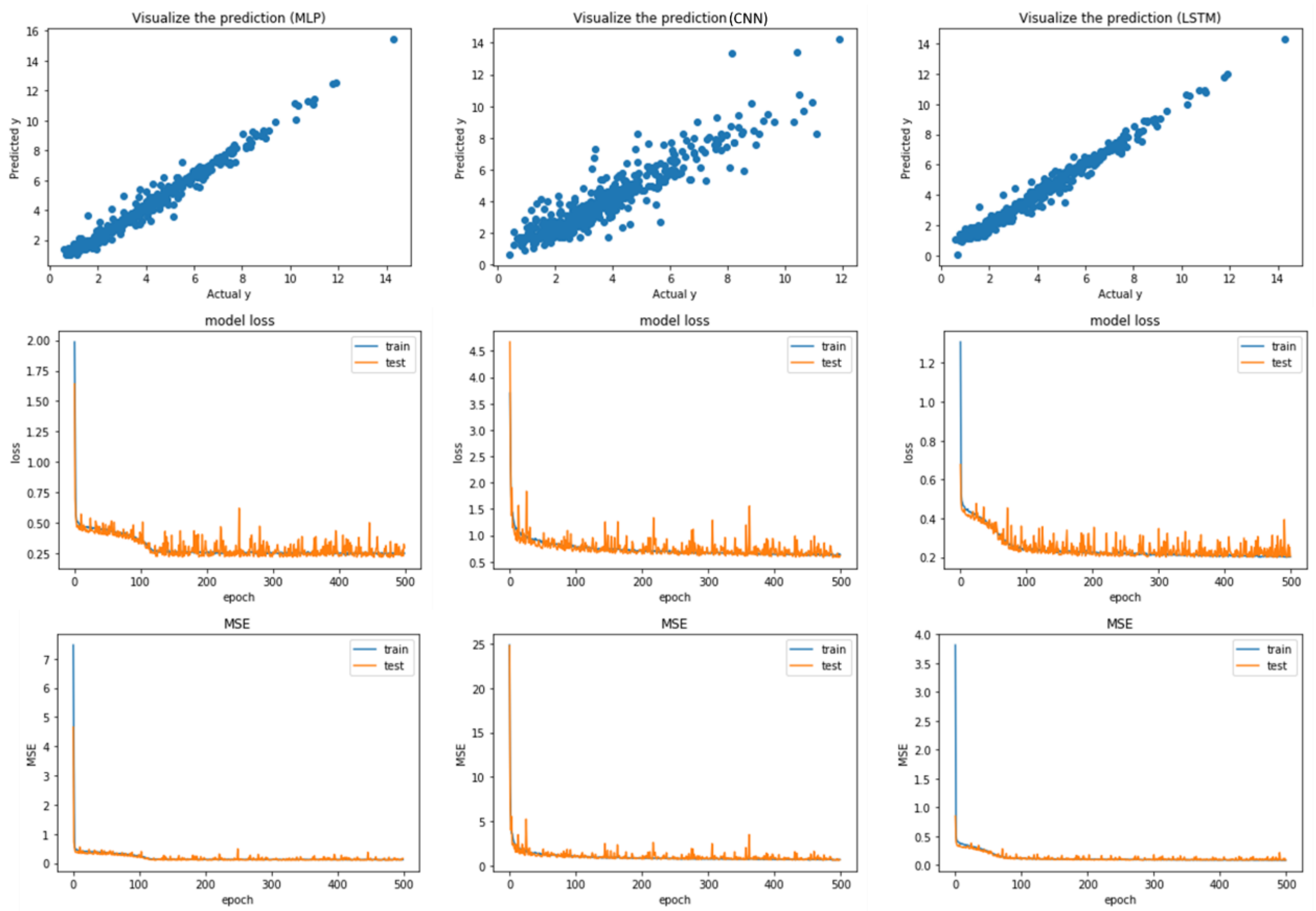

Figure 14. Deep Learning Prediction Visualization

# Chapter Three: Wind Speed Dependency Modeling of Two Locations

Quantifying the dependency between two random variables is a mathematical technique to analyze their joint behavior. However, modeling the dependency sometimes misunderstood with statistical correlation. Dependency has a wider concept compared to correlation. For instance, two random variables could be dependent but not correlated, while if the two random variables are correlated, then they are dependent. Correlation gives some insight into dependency, but ideally, the dependence structure is formulated by leveraging another mathematical approach called copula. The concept of copula has been around for a long time and utilized mostly in econometric modeling and risk assessment. In this section, we would like to consider this strong mathematical tool to study wind speed dependence. In the following sections, we will describe some fundamentals of dependency and copula, wind speed site selection, and dependence modeling.

## 3.1 Dependency and Copulas

### 3.1.1 Dependence

Dependence between two random variables, $X$ and $Y$, exists if they do not satisfy the independence property [123].

We will review independence between two random variables before diving into the dependence. Two random variables or events are independent if realization or occurrence of one does not affect the probability of occurrence, simply probability distribution of others. In other words, the occurrence of one does not affect the odds. Odds of an event is the likelihood that the event will occur. For two random events, $X$ and $Y$, they are independent if, $P(X|Y) = P(X)$ and vice versa. Another way to explain independence is with marginal distributions. Let, the marginal distribution functions of $X$ and $Y$ are $F_1(X)$ and $F_2(Y)$, then they are independent only if (26) is satisfied.

$$F(X, Y) = F_1(X)F_2(Y) \tag{26}$$

Dependence or a measure of dependence summarizes the dependency structure of two random variables. Let's assume, $\delta$ is a measure of dependence between $X\epsilon\mathbb{R}$ and $Y\epsilon\mathbb{R}$. It has some properties (considering $X$ and $Y$ are not independent). According to [124], the following are the dependence properties:

- Conditional symmetry: $\delta(X,Y) = \delta(Y,X)$
- Conditional normalization: $-1 \leq \delta(X,Y) \leq 1$
- *(X, Y)* comonotonic: $\delta(X,Y) = 1$
  *(X, Y)* countermonotonic: $\delta(X,Y) = -1$
- For strictly monotonic transformation $T: \mathbb{R} \to \mathbb{R}\ of\ X$

$$\delta(T(X),Y) = \begin{cases} \delta(Y,X) & T\ increasing \\ -\delta(Y,X) & T\ decreasing \end{cases}$$

Conditional symmetry means the dependence structure does not change between two real-valued variables according to their order. The second property in the above list translates the nature of the dependency measure between two given random variables. The dependency could be positively or negatively while the absolute value of $\delta$ lies in a uniform distribution, $|\delta| \sim U(0,1)$. The perfect dependency bounds are +1 and -1. In the case of the existence of perfect positive dependence, the bound random variables are called comonotonic, while the perfect negative scenario is described as countermonotonic. However, in both cases, uniformly distributed dependence maintains. While perfect monotonicity is defined, strictly monotonicity is quite straight forward. When the dependence measure excludes the comonotonic and countermonotonic conditions, $\delta$ is whether increasing (+) or decreasing (-) only (depending on T, which is a function of $X$) and never reaches to the perfect monotonic conditions; then it is strictly monotonic. In the following section, we will discuss copula as a dependence measure between random variables.

### 3.1.2 Copulas

The copula is a joint distribution of marginals that are parametrically specified. The difference between the copula model and regular joint distribution formation is the separation of marginal distribution and dependence structure of the variables from their joint distribution function [125]. Therefore, copulas are flexible and efficient for modeling dependence [126]. A very simple copula modeling of $X$ and $Y$ with $F_1(X)$ and $F_2(Y)$ marginals, denoting $F(X,Y)$ as the joint distribution and $C(s,t)$ a bivariate copula, where $s$ and $t$ are two copula parameters (copula could be both single or multiple parameters)-

$$F(X,Y) = C_{s,t}(F_1(X), F_2(Y)) \tag{27}$$

$C_{s,t}$ is a multivariate distribution function from the unit d-cube $[0,1]^d$ (where d is the dimension) to the unit interval [0,1] and satisfy the following properties-

- If the realization of the d-1 variables are known with marginal probability = 1, the d outcomes of the joint probability = 1 with an uncertain outcome ($u_i$).

$$C(1 \dots ,1, u_i, 1, \dots , 1) = u_i \ \forall \ i \leq d \ and \ u_i \in [0,1] \tag{28}$$

- Grounded property: If the realization of one variable has the marginal probability zero, then the joint probability of all outcomes is zero.

$$C(u_1, \dots , u_d) = 0 \ if \ u_i = 0 \ \forall \ i \leq d \tag{29}$$

- The joint probability will not be negative as the volume (C) of any d-dimensional interval is non-negative.
- Fréchet bound: the upper and lower bounds of a copula are known as the Fréchet-Hoeffding upper and lower bound, respectively.

$$max\left\{\sum_{i=1}^{d} u_i + 1 - d, 0\right\} \leq C(u) \leq min\{u_1, \dots , u_d\} \tag{30}$$

### 3.1.3 Sklar's Theorem

Sklar's theorem explains the copula's role in multivariate distribution functions and the univariate marginal distributions. If *G(x,y)* is a joint distribution with margins $F_1(x)$ and $F_2(y)$, there exists a copula *C* for all *x* and *y*,

$$G(x, y) = C(F_1(x), F_2(y)) \tag{31}$$

Where,

$F_1$ and $F_2$ are continuous distributions, and *C* is unique, determined by the ranks of $F_1(x)$ and $F_2(y)$: $RanF_1(x) \times RanF_2(x)$. Some important corollaries are-

$$G(X \leq x, Y \leq y) = C\big(F_1(x), F_2(y)\big) \tag{32}$$

$$G(X \leq x, Y > y) = F_1(x) - C\big(F_1(x), F_2(y)\big) \tag{33}$$

$$G(X > x, Y \leq y) = F_2(y) - C\big(F_1(x), F_2(y)\big) \tag{34}$$

$$G(X \leq x \mid Y \leq y = \frac{C\big(F_1(x), F_2(y)\big)}{F_2(y)} \tag{35}$$

$$G(X \leq x \mid Y > y) = \frac{F_1(x) - C(F_1(x), F_2(y))}{1 - F_2(y)} \tag{36}$$

$$G(X \leq x \,|Y = y| = C_{1|2}\big(F_1(x), F_2(y)\big) = \frac{\delta C(v.z)}{\delta z} |v = F_1(x), z = F_2(y) \quad (37)$$

There are three types of copulas, fundamental, implicit, and explicit copulas. While fundamental copulas denote perfect positive dependence, independence, and perfect negative dependence. The implicit copula is extracted from a multivariate distribution. The explicit copula is also called Archimedean copulas. From this point on, we will restrict our interest to only Archimedean bivariate copulas as we shall explore this family for our proposed problem statement.

### 3.1.4 Copula Families

In this section, we will discuss some of the popular bivariate Archimedean copula families. We shall also look at a few of the mixed copula families that combine two or more copulas to solve different problems. As following copulas describe the joint distribution of two variables, we define $(u_1, u_2)$ as our two variables of interest.

#### 3.1.4.1 Frank Copula

Frank Copula, $C_\theta^{Fr}(u_1, u_2)$, considers a maximum range of dependence. It allows the approximation of the upper and lower Fréchet-Hoeffding bounds [127]. Therefore, it is useful for both positive and negative dependence. The copula parameter, $\theta$ is defined as $-\infty < \theta < +\infty$.

$$C_\theta^{Fr}(u_1, u_2) = -\theta^{-1} \, log\left\{1 + \frac{(e^{-\theta u_1}-1)(e^{-\theta u_2}-1)}{(e^{-\theta}-1)}\right\} \quad (38)$$

#### 3.1.4.2 Joe Copula

Joe copula, $C_\theta^J(u_1, u_2)$, is effective when there is upper tail dependence; thus, the two variables have a high positive correlation [128]. The Joe copula parameter, $\theta$ is defined as $1 \leq \theta < +\infty$.

$$C_\theta^J(u_1, u_2) = 1 - [(1-u_1)^\theta + (1-u_2)^\theta - (1-u_1)^\theta\big(1-u_2)^\theta\big]^{1/\theta} \quad (39)$$

#### 3.1.4.3 Gaussian Copula

The Gaussian copula, $C_P^{Ga}(u_1, u_2)$, is highly distributed around the center and shows almost no dependence in the tails. Therefore, sometimes, it is not the best choice when it comes to copula dependency modeling [129]. It is expressed ad an integral over the density of $s_1$, where $s_1$ is a standardized version of $s_2$. $s_2$ is a random variable described by a normal distribution of $d$ diemsions (number of features), $s_2 \sim N_d(\mu, \sigma)$, where $\mu$ and $\sigma$ denote mean and standard deviation, respectively. $p$ is the strength of dependence, $1 \leq p < +\infty$.

$$C_p^{Ga}(u_1, u_2) = \int_{-\infty}^{\phi-1(u_1)} \int_{-\infty}^{\phi-1(u_2)} \frac{1}{2\pi(1-p^2)^{1/2}} exp\left(\frac{-(s_1^2 - 2p_{s_1 s_2} + s_2^2)}{2(1-p^2)}\right) ds_1 ds_2 \qquad (40)$$

### 3.1.4.4 Gumbel Copula

Gumbel copula, $C_\theta^{Gu}(u_1, u_2)$, is useful when there is asymmetric dependence in data [130]. It is a design choice for strong upper tail dependence and weak lower tail dependence. Parameter $\theta$ is defined as $1 \leq \theta < +\infty$.

$$C_\theta^{Gu}(u_1, u_2) = exp(-((-log\, u_1)^{\,\theta} + (-log\, u_2)^{\theta})^{\frac{1}{\theta}}) \qquad (41)$$

### 3.1.4.5 Clayton Copula

Clayton copula, $C_\theta^{Cl}(u_1, u_2)$ [131], on the other hand, has an opposite intuition of Gumbel copula. So, it is appropriate for weak upper tail dependence and strong lower tail dependence. Formula (46) describes this copula family with a parameter $\theta$, defined as $0 \leq \theta < \infty$.

$$C_\theta^{Cl}(u_1, u_2) = (u_1{}^{-\theta} + u_2{}^{-\theta} - 1)^{1/\theta} \qquad (42)$$

### 3.1.4.6 Clayton-Gumbel (BB1) Copula

Clayton-Gumbel (BB1) copula, $C_{\theta,\delta}^{BB1}(u_1, u_2)$ [132], is a combination of both the extreme cases of Clayton and Gumbel copulas. It is a two-parameter copula family. $\theta$ and $\delta$ are the parameters defined by $\delta \geq 1$ and $\theta > 0$ respectively.

$$C_{\theta,\delta}^{BB1}(u_1, u_2) = (1 + \left[(u_1^{-\theta} - 1)^{\delta} + (u_2^{-\theta} - 1)^{\delta}\right]^{1/\delta})^{-1/\theta} \qquad (43)$$

### 3.1.4.7 Joe-Gumbel (BB6) Copula

Joe-Gumbel (BB6) copula, $C_{\theta,\delta}^{BB6}(u_1, u_2)$ [133], is a mixture copula of Joe and Gumbel families. One of its use is to detect asymmetric tail dependencies. The parameters, $\theta$ and $\delta$ are defined by $\delta \geq 1$ and $\theta \geq 1$.

$$C_{\theta,\delta}^{BB6}(u_1, u_2) = 1 - (1 - exp - \left[(-log(1 - u_1)^{\theta}))\right)^{\delta} + (-log(1 - (1 - u_2)^{\theta}))^{\delta}\big]^{1/\delta})^{1/\theta} \qquad (44)$$

### 3.1.4.8 Joe-Clayton (BB7) Copula

Joe-Clayton (BB7) copula, $C_{\theta,\delta}^{BB7}(u_1, u_2)$ [134], allows both positive and negative quantile curves. Therefore, it works great when modeling tail dependence and quantile regression. It has two parameters, $\theta$ and $\delta$, which are defined by $\theta \geq 1$ and $\delta > 0$ respectively.

$$C_{\theta,\delta}^{BB7}(u_1, u_2) = 1 - \left[1 - ((1 - u_1^{\theta})^{-\delta} + (1 - u_2)^{-\delta} - 1)^{-1/\delta}\right]^{1/\theta} \tag{45}$$

### 3.1.4.9 Joe-Frank (BB8) Copula

Joe-Frank (BB8) Copula, $C_{\theta,\delta}^{BB8}(u_1,\ u_2)$ [135] is a mixture model of Joe and Frank copulas. Therefore, it combines their properties of upper tail dependence (Joe) and strong positive or strong negative dependence (Frank). This is also a two-parameter copula family $(\theta, \delta)$, which are constrained by $\theta \geq 1$ and $0 < \delta < 1$.

$$C_{\theta,\delta}^{BB8}(u_1,\ u_2) = \frac{1}{\theta}\left(1 - [1 - \frac{1}{1-(1-\delta)^{\theta}}\left(1 - (1 - \delta u_1)^{\theta}\right)((1 - \delta u_2)^{\theta})]^{\frac{1}{\delta}}\right) \tag{46}$$

## 3.2 Copula Dependency Selection

Wind speed data, as described in Table 3, are grouped to form a set of varying distance wind speed pairs. Table 3 shows the wind speed pairs, the distance between them, and their correlation coefficient using the Spearman method. Later, the copula family to describe the dependency is selected based on two established approaches and shown them in the table. Copulas are fitted against the data using maximum likelihood estimation, and then the criterion is computed for all the available families (list available in Appendix A). Two of the efficient criteria are Akaike Information Criterion (AIC) and Bayesian Information Criterion (BIC) as shown for observations $u_{i,j,}\ i = 1, 2, \ldots, N, k = 1, 2$ in (31) and (32) respectively. C is the copula family, and $\theta$ is its parameter. BIC adds a stronger penalty in the selection; thus, it is preferable in the later modeling. However, both approaches are utilized for copula selection between wind speed pairs already have described and listed in Table 3 for hourly average wind speed data. A list of a similar study was also made for six-hourly, daily, and weekly data pairs, which is available in Appendix B.

$$AIC = -2\sum_{i=1}^{N} ln\left[c\left(u_{i,1}, u_{i,2}|\theta\right)\right] + 2k \tag{47}$$

$$BIC = -2\sum_{i=1}^{N} ln\left[c\left(u_{i,1}, u_{i,2}|\theta\right)\right] + ln(N)k \tag{48}$$

## 3.4 Simulations and Results

### 3.4.1 Wind Speed Sites

We collected hourly wind speed data of 17 different locations from the NREL Wind Prospector Interface [136] for the whole 2012 year. Table 3 lists information about wind speed site locations. Alphabetical labels, letter ‘A’ to ‘Q’ denote the wind speed sites for easy referencing. Table 3 lists further site information- corresponding NREL site ID, latitude/longitude, and city, for reproducibility of the work.

Table 3 Wind Speed Sites Information

| **Site** | **Site ID** | **Latitude** | **Longitude** | **City, State** |
|---|---|---|---|---|
| A | 45537 | 39.219364 | -104.559937 | Black Forest, CO |
| B | 49712 | 40.251148 | -94.675476 | Conception Junction, MO |
| C | 54196 | 40.476978 | -88.022736 | Paxton, IL |
| D | 120519 | 45.543896 | -120.771515 | Biggs, OR |
| E | 90896 | 43.785511 | -94.852295 | Alpha, MN |
| F | 91071 | 43.803856 | -94.826111 | Trimont, MN |
| G | 89755 | 43.694996 | -95.13736 | Lakefield, MN |
| H | 89744 | 43.697094 | -95.472107 | Brewster, MN |
| I | 90688 | 43.772705 | -95.832397 | Wilmont, MN |
| J | 92971 | 43.996262 | -96.038818 | Woodstock, MN |
| K | 93196 | 44.014835 | -96.116516 | Woodstock, MN |
| L | 92961 | 43.995804 | -96.323547 | Pipestone, MN |
| M | 77005 | 42.59547 | -91.385681 | Greeley, IA |
| N | 92377 | 43.903042 | -93.117889 | Blooming Prairie, MN |
| O | 106449 | 42.312954 | -71.039246 | Dorchester, MA |
| P | 106983 | 42.395168 | -71.056274 | Everett, MA |
| Q | 64029 | 39.383095 | -74.450989 | Atlantic City, NJ |

### 3.4.2 Copula Selections

In this section, we would like to find appropriate copula families for different wind speed pairs. We considered 39 wind speed pairs of different distances ranging from 3 to 2700 km. We have analyzed the correlation between each pair and then selected the copula model using both AIC and BIC criteria for comparison. Table 4 lists the wind speed pairs with corresponding distances, correlations, and copula selections from left to right columns. We used Spearman's rank correlation coefficient (described in Section 2.2.2.1), and the values suggest a decreasing trend with the increase in distance. For lower distances until 50 km, the correlation coefficient is greater than 0.9. However, it drops to 0.5 or lower when the distance goes higher than 300 km. For a longer distance (>2000 km), the correlation goes to extremely low ($\leq 0.01$).

For copula selection, both AIC and BIC produced similar results in terms of copula family and parameter estimation. However, we tend to follow the BIC selection and AIC was used only for comparison purposes. The copula families we found appropriate for our wind speed pairs are Frank, t, Gaussian, Gumbel, BB1, BB6, and BB8. For long distances, 2700 km, we could not find a dependence. After analyzing the copula selections with distances in Table 4, we found an interesting pattern of copula selections. For distance, less than 130 km, simple bivariate copulas (mostly Frank and t); then until 440 km, mixed copulas (BB6, BB8); until 1500 km, again some simple copulas (Frank, Gaussian, Gumbel); and for long-distance (>1500 km) mixed copulas (BB1) were found.

Table 4. Copula Selections for Wind Speed Pairs

| **Wind Speeds** | **Distance** | **Corr** | **Bivariate Copula "AIC"** | **Bivariate Copula "BIC"** |
|---|---|---|---|---|
| EF | 3 | 0.9942 | Frank (par = 66.66, tau = 0.94) | Frank (par = 66.66, tau = 0.94) |
| JK | 7 | 0.9808 | t (par = 0.98, par2 = 3.6, tau = 0.88) | t (par = 0.98, par2 = 3.6, tau = 0.88) |
| OP | 9 | 0.9375 | t (par = 0.94, par2 = 4.13, tau = 0.78) | t (par = 0.94, par2 = 4.13, tau = 0.78) |
| KL | 17 | 0.9645 | Frank (par = 23.9, tau = 0.84) | Frank (par = 23.9, tau = 0.84) |
| JL | 23 | 0.9459 | t (par = 0.94, par2 = 6.14, tau = 0.79) | t (par = 0.94, par2 = 6.14, tau = 0.79) |
| EG | 25 | 0.9443 | Frank (par = 18.41, tau = 0.8) | Frank (par = 18.41, tau = 0.8) |
| GH | 27 | 0.9474 | Frank (par = 19.75, tau = 0.81) | Frank (par = 19.75, tau = 0.81) |
| FG | 28 | 0.9424 | Frank (par = 18.24, tau = 0.8) | Frank (par = 18.24, tau = 0.8) |
| IJ | 30 | 0.9459 | t (par = 0.95, par2 = 3.15, tau = 0.8) | t (par = 0.95, par2 = 3.15, tau = 0.8) |
| HI | 31 | 0.9352 | Frank (par = 17.64, tau = 0.79) | Frank (par = 17.64, tau = 0.79) |
| IK | 35 | 0.9323 | Frank (par = 17.2, tau = 0.79) | Frank (par = 17.2, tau = 0.79) |
| IL | 47 | 0.9074 | t (par = 0.91, par2 = 5.7, tau = 0.73) | t (par = 0.91, par2 = 5.7, tau = 0.73) |
| EH | 51 | 0.9136 | Frank (par = 14.55, tau = 0.76) | Frank (par = 14.55, tau = 0.76) |
| FH | 53 | 0.9126 | Frank (par = 14.54, tau = 0.76) | Frank (par = 14.54, tau = 0.76) |
| HJ | 56 | 0.8980 | Frank (par = 13.26, tau = 0.74) | Frank (par = 13.26, tau = 0.74) |
| GI | 57 | 0.9083 | t (par = 0.91, par2 = 4.13, tau = 0.73) | t (par = 0.91, par2 = 4.13, tau = 0.73) |
| HK | 63 | 0.8840 | Frank (par = 12.11, tau = 0.71) | Frank (par = 12.11, tau = 0.71) |
| HL | 76 | 0.8643 | Frank (par = 10.73, tau = 0.68) | Frank (par = 10.73, tau = 0.68) |
| EI | 79 | 0.8648 | Frank (par = 10.75, tau = 0.68) | Frank (par = 10.75, tau = 0.68) |
| GJ | 80 | 0.8740 | t (par = 0.88, par2 = 4.18, tau = 0.69) | t (par = 0.88, par2 = 4.18, tau = 0.69) |
| FI | 81 | 0.8650 | Frank (par = 10.8, tau = 0.69) | Frank (par = 10.8, tau = 0.69) |
| GK | 86 | 0.8645 | Frank (par = 11.09, tau = 0.69) | Frank (par = 11.09, tau = 0.69) |
| EJ | 98 | 0.8338 | Frank (par = 9.33, tau = 0.65) | Frank (par = 9.33, tau = 0.65) |
| FJ | 100 | 0.8350 | Frank (par = 9.4, tau = 0.65) | Frank (par = 9.4, tau = 0.65) |
| GL | 101 | 0.8420 | Frank (par = 9.83, tau = 0.66) | Frank (par = 9.83, tau = 0.66) |
| EK | 104 | 0.8190 | Frank (par = 8.68, tau = 0.63) | Frank (par = 8.68, tau = 0.63) |
| FK | 106 | 0.8198 | Frank (par = 8.73, tau = 0.63) | Frank (par = 8.73, tau = 0.63) |
| EL | 120 | 0.7920 | Frank (par = 7.82, tau = 0.6) | Frank (par = 7.82, tau = 0.6) |
| FL | 122 | 0.7917 | Frank (par = 7.83, tau = 0.6) | Frank (par = 7.83, tau = 0.6) |
| EN | 139 | 0.7927 | BB8 (par = 6, par2 = 0.79, tau = 0.59) | BB8 (par = 6, par2 = 0.79, tau = 0.59) |
| EM | 310 | 0.5465 | BB8 (par = 3.61, par2 = 0.75, tau = 0.38) | BB8 (par = 3.61, par2 = 0.75, tau = 0.38) |
| OQ | 434 | 0.4847 | BB6 (par = 1.17, par2 = 1.35, tau = 0.32) | BB6 (par = 1.17, par2 = 1.35, tau = 0.32) |
| PQ | 440 | 0.4831 | BB6 (par = 2.27, par2 = 0.9, tau = 0.31) | BB8 (par = 2.27, par2 = 0.9, tau = 0.31) |
| BC | 565 | 0.3848 | Survival BB1 (par = 0.48, par2 = 1.06, tau = 0.24) | Gumbel (par = 1.33, tau = 0.25) |
| AB | 852 | 0.1719 | Gaussian (par = 0.17, tau = 0.11) | Gaussian (par = 0.17, tau = 0.11) |
| AC | 1417 | 0.1089 | Frank (par = 0.66, tau = 0.07) | Frank (par = 0.66, tau = 0.07) |
| AD | 1502 | 0.1340 | Gaussian (par = 0.14, tau = 0.09) | Gaussian (par = 0.14, tau = 0.09) |
| BD | 2195 | 0.0768 | BB1 (par = 0.07, par2 = 1.02, tau = 0.05) | BB1 (par = 0.07, par2 = 1.02, tau = 0.05) |
| CD | 2700 | -0.0144 | Independence (par = 0, tau = 0) | Independence (par = 0, tau = 0) |

### 3.4.3 Copula Prediction (Case Study)

In this section, we would like to evaluate the selected copula families in Section 3.4.2. Our goal is to evaluate how efficient the selected copula family is in terms of predicting the samples of the joint distribution compared to the real data. For a case study, we chose 'PQ' wind speed pair. In the next sections, we will find marginal distributions for wins speeds 'P' and 'Q', then fit selected copula (BB8, from Table 4) and analyze the accuracy.

#### 3.4.3.1 Dataset

The wind speed data of 'P' and 'Q' locations are described in Table 5. The unit of wind speeds is in meters per second (m/s). The dataset contains hourly average wind speed data for both locations, totaling 8570 data points (we filtered out wind speeds above 15 m/s as it is beyond the scope of wind power application [137]). There is a small difference in mean, standard deviation, and median. However, both speeds have a similar standard error. Figure 15 shows a visual distribution of both speeds for an initial inspection if we still need data cleaning.

Table 5 Wind Speed Data Description (PQ)

| **Speed** | **No. of Datapoints** | **Mean (m/s)** | **Std. Dev.** | **Median (m/s)** | **Min (m/s)** | **Max (m/s)** | **Skewness** | **kurtosis** | **Std Error** |
|---|---|---|---|---|---|---|---|---|---|
| P | 8570 | 6.14 | 2.88 | 5.97 | 0.11 | 14.98 | 0.31 | -0.38 | 0.03 |
| Q | 8570 | 6.92 | 3.04 | 6.77 | 0.04 | 14.95 | 0.16 | -0.59 | 0.03 |

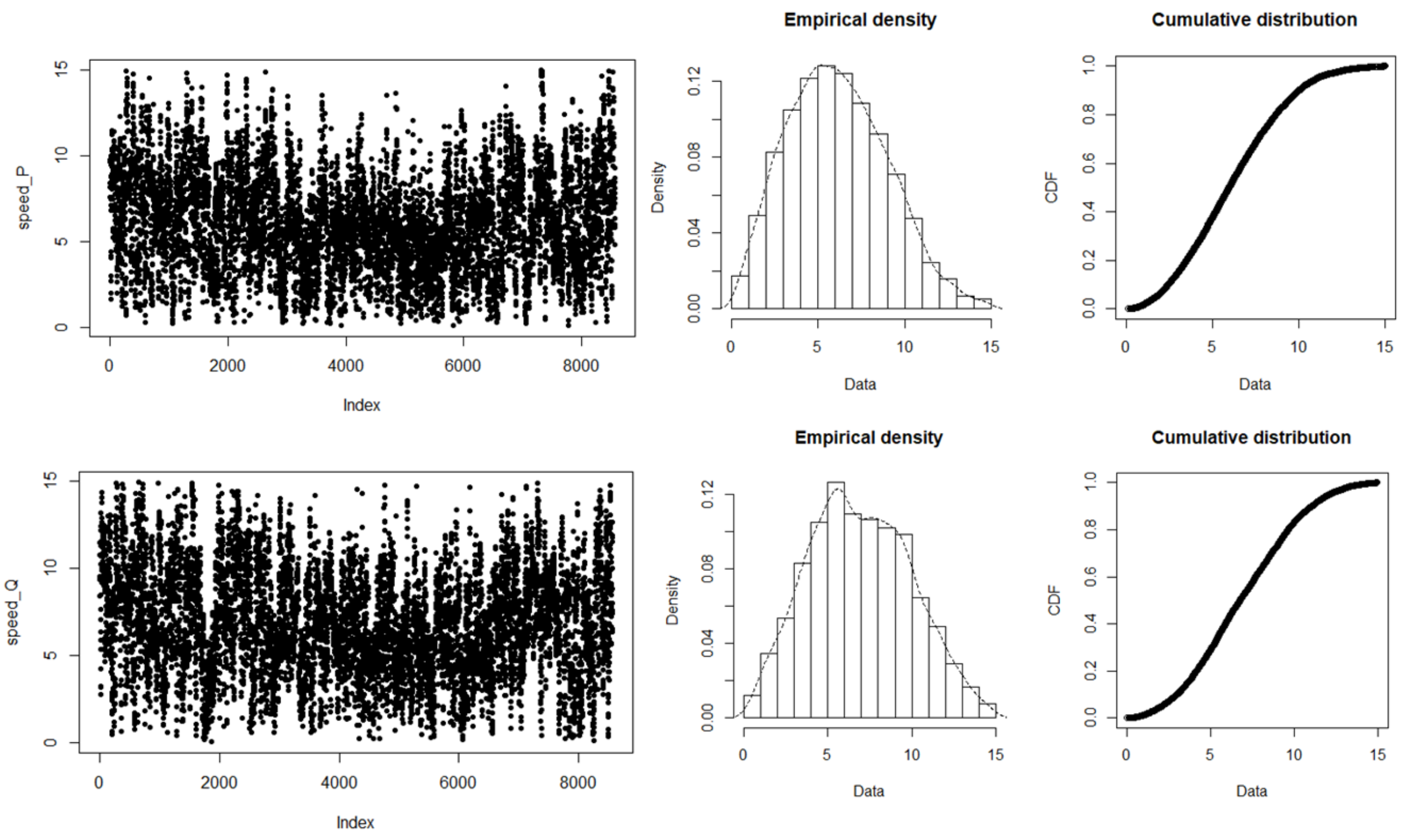

Figure 15. Wind Speed Data Visualization

### 3.4.3.2 Marginal Distributions

The marginal distribution of variables is very important for copula application. Therefore, in this section, we inspect the data to find an appropriate distribution family and analyze the distribution parameters empirically. Cullen and Frey graph [138] Thus, Figure 16 provides us a visual picture to begin the fitting process with some candidate distributions. Based on observation on the graph, we selected Weibull, Gamma, and Log-normal distributions as candidates. Weibull distribution for shape parameter a and scale parameter b is given in (49). (50) and (51) describes Gamma (shape parameter a and scale parameter $s$) and Log-normal distributions (where $\mu$ and $\sigma$ are the mean and standard deviation of the logarithm) respectively. Later, those distributions were fitted empirically on our dataset, and the parameter estimates are listed in Table 6.

$$f(x) = \frac{a}{b}\left(\frac{x}{b}\right)^{a-1} e^{-\left(\frac{x}{b}\right)^{a}} \tag{49}$$

$$f(x) = \frac{1}{\Gamma(a)s^{a}} x^{a-1} e^{-\left(\frac{x}{s}\right)} \tag{50}$$

$$f(x) = \frac{1}{x\sigma\sqrt{2\pi}} e^{-\left(\frac{(\log x-\mu)^2}{2\sigma^2}\right)} \tag{51}$$

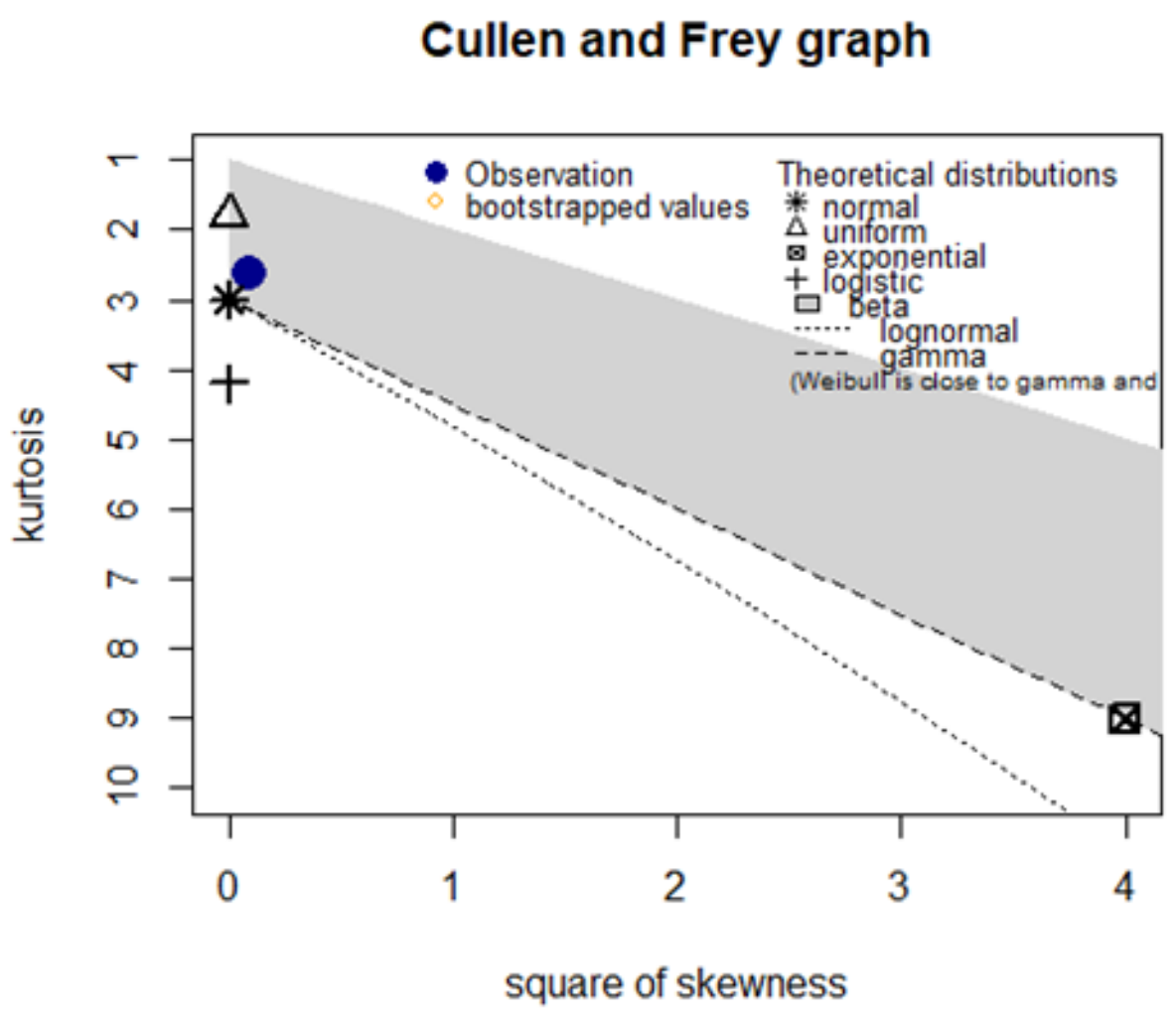

(a) Speed_P

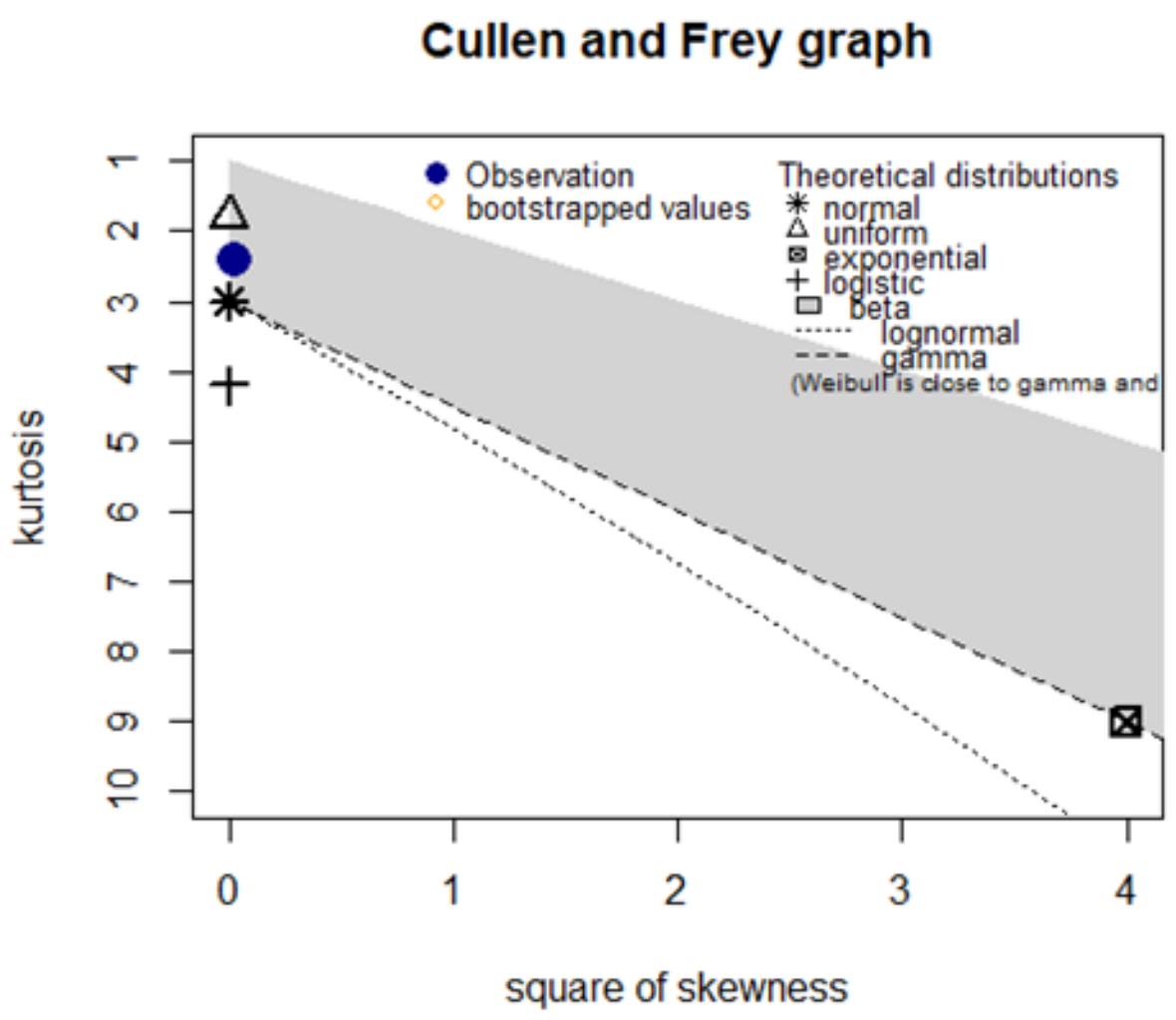

(b) Speed_Q

Figure 16. Cullen and Frey Graphs

Table 6. Estimated Parameters of Candidate Distributions

| **Distribution** | **Speed_P** | **Speed_Q** |
|---|---|---|
| Weibull | Estimate Std. Error<br>Shape: 2.254787 0.01934266<br>Scale: 6.922302 0.03485391<br>Loglikelihood: -21011.08<br>AIC: 42026.17<br>BIC: 42040.28 | Estimate Std. Error<br>Shape: 2.427970 0.02101355<br>Scale: 7.790036 0.03639017<br>Loglikelihood: -21586.55<br>AIC: 43177.1<br>BIC: 43191.21 |
| Gamma | Estimate Std. Error<br>Shape: 3.6631846 0.053607228<br>Rate: 0.5968006 0.009360584<br>Loglikelihood: -21312.16<br>AIC: 42628.32<br>BIC: 42642.44 | Estimate Std. Error<br>Shape: 4.0542265 0.059560922<br>Rate: 0.5861729 0.009167688<br>Loglikelihood: -21987.9<br>AIC: 43979.8<br>BIC: 43993.91 |
| Log-normal | Estimate Std. Error<br>Mean: 1.6717717 0.006437771<br>Std: 0.5959721 0.004552134<br>Loglikelihood: -22051.89<br>AIC: 44107.77<br>BIC: 44121.88 | Estimate Std. Error<br>Mean: 1.8055804 0.006151945<br>Std: 0.5695119 0.004350022<br>Loglikelihood: -22809.43<br>AIC: 45622.85<br>BIC: 45636.97 |

Candidate distributions then tested to see how they fit on the data. Figure 17 and 18 depicts the plots showing Weibull fits better than others. Q-Q and P-P plots also support this conclusion. Figure 19 and 20 further displays the plots only for Weibull distribution for better a visual.

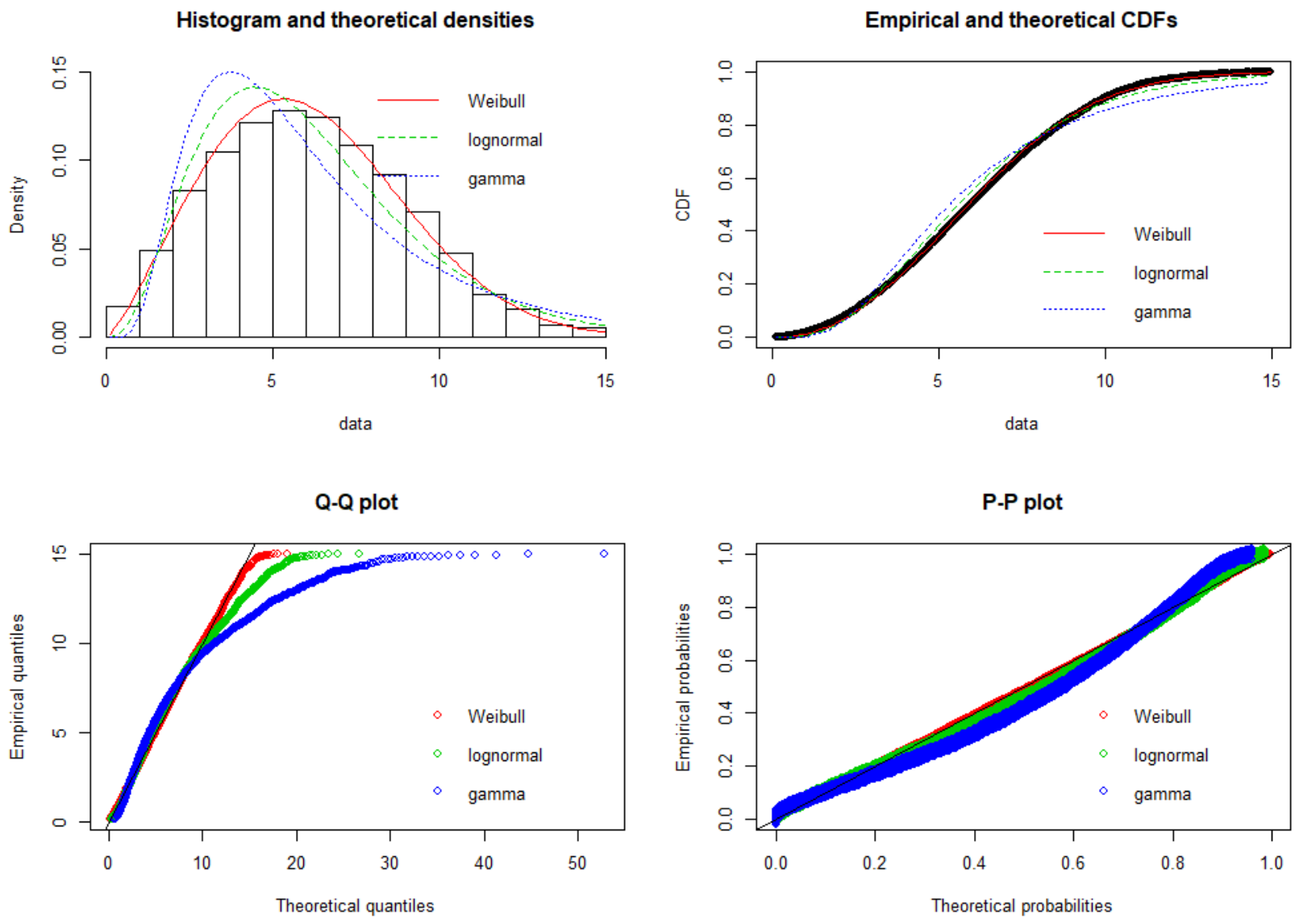

Figure 17. Distribution Fitting Inspections (Speed_P)

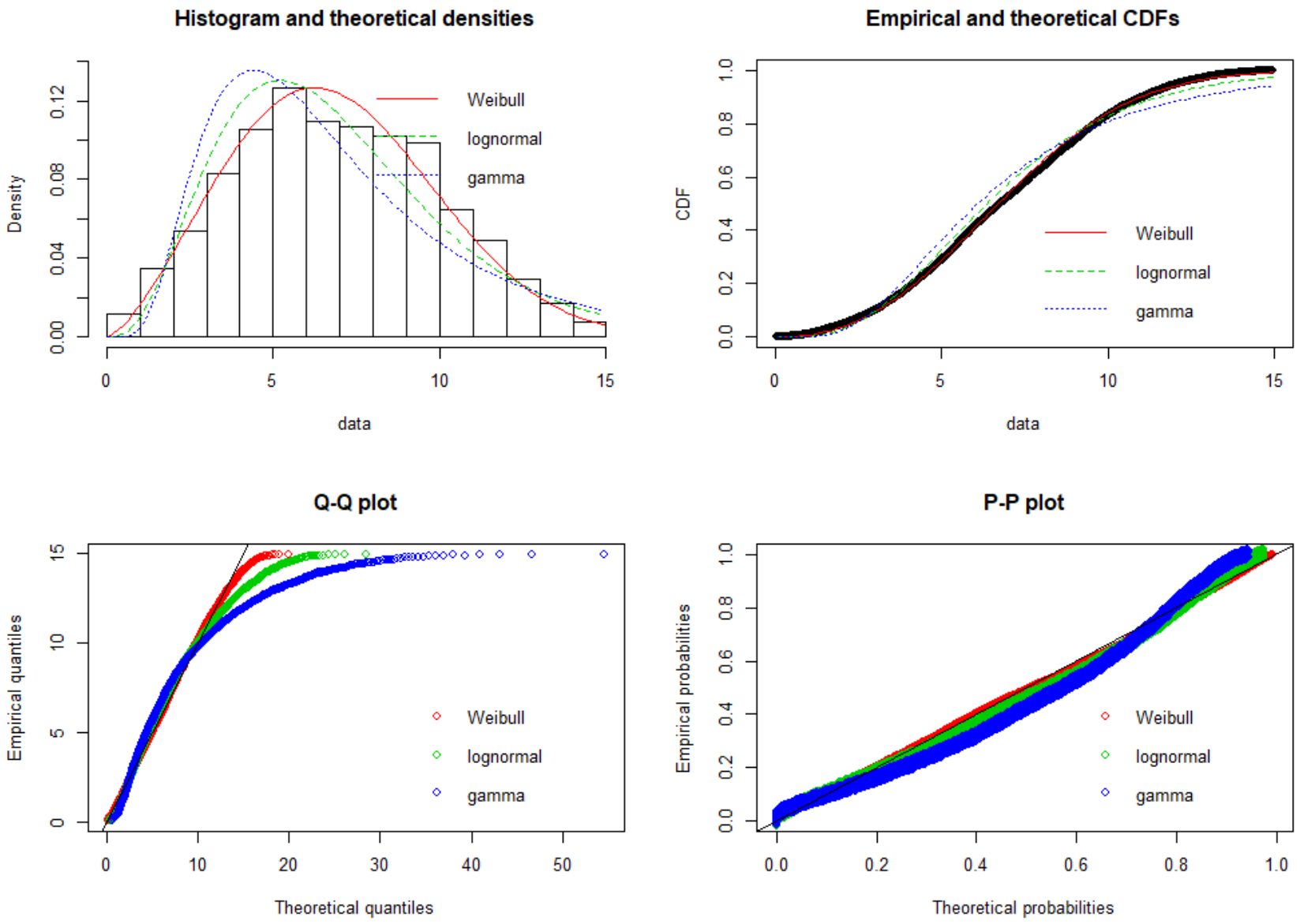

Figure 18 Distribution Fitting Inspections (Speed_Q)

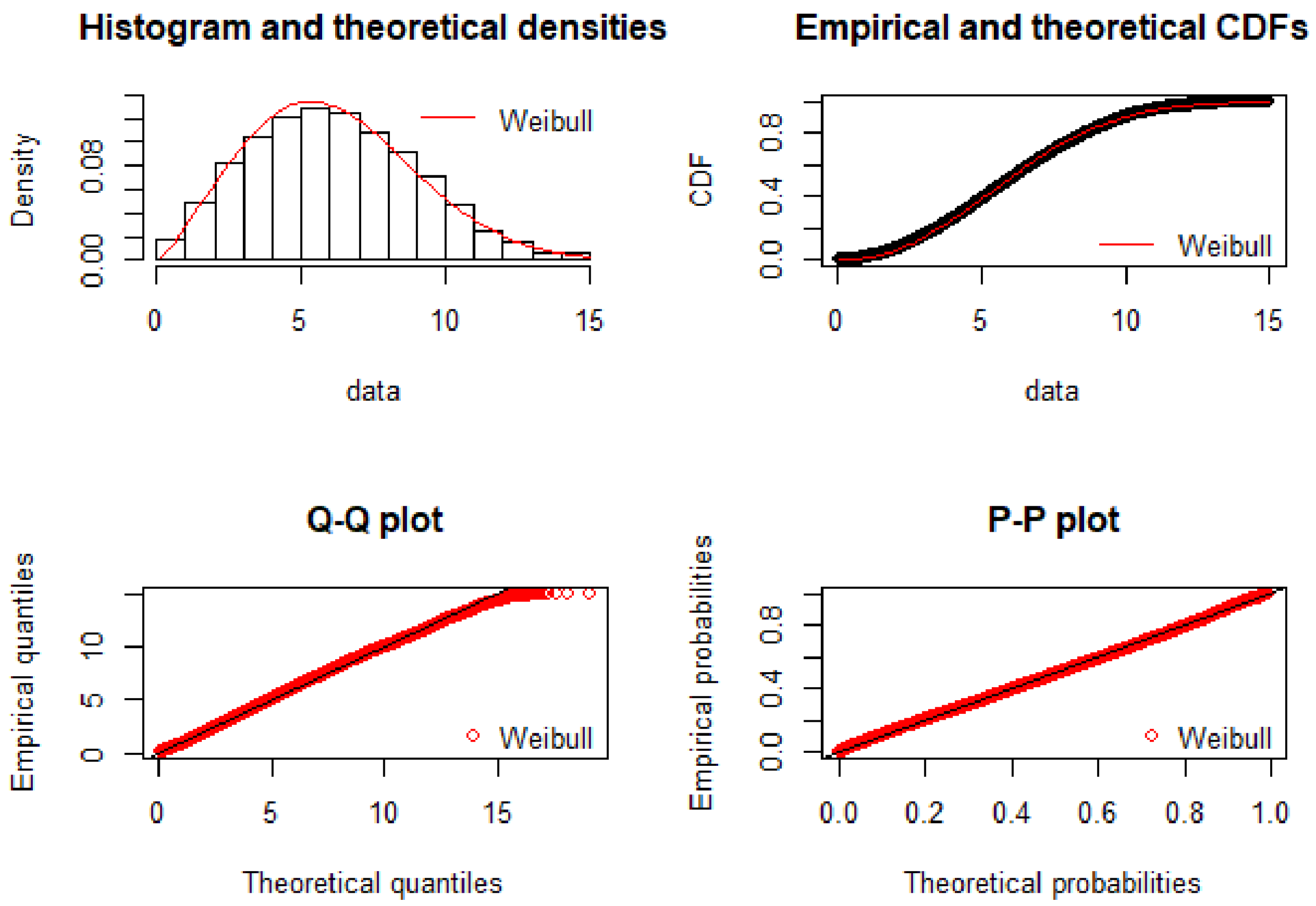

Figure 19.Weibull Distribution Fitting for Speed_P

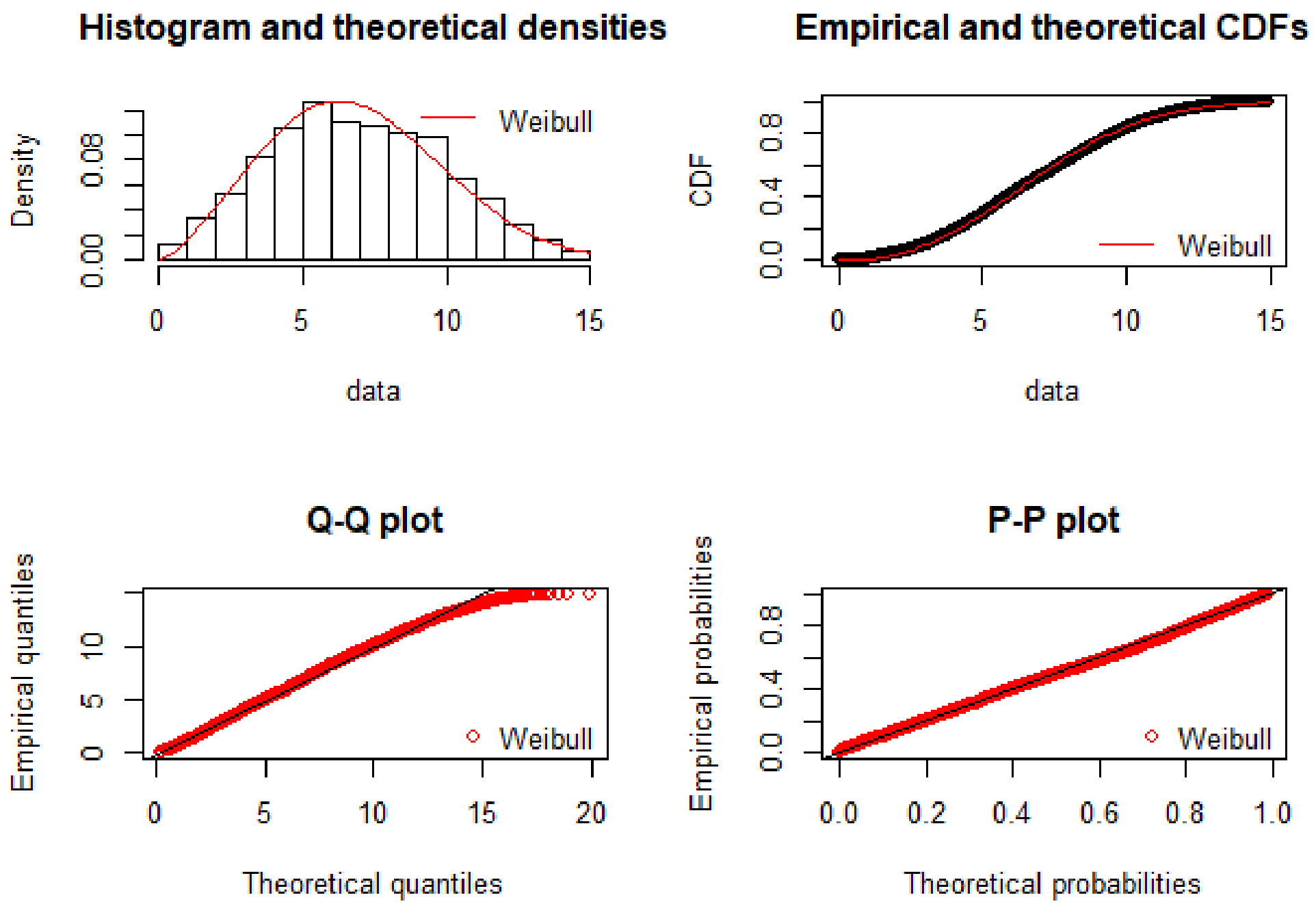

Figure 20 Weibull Distribution Fitting for Speed_Q

Finally, we wanted to see how simulated and compare Weibull samples with wind speed data (‘P’ and ‘Q’), as shown in Figure 21.

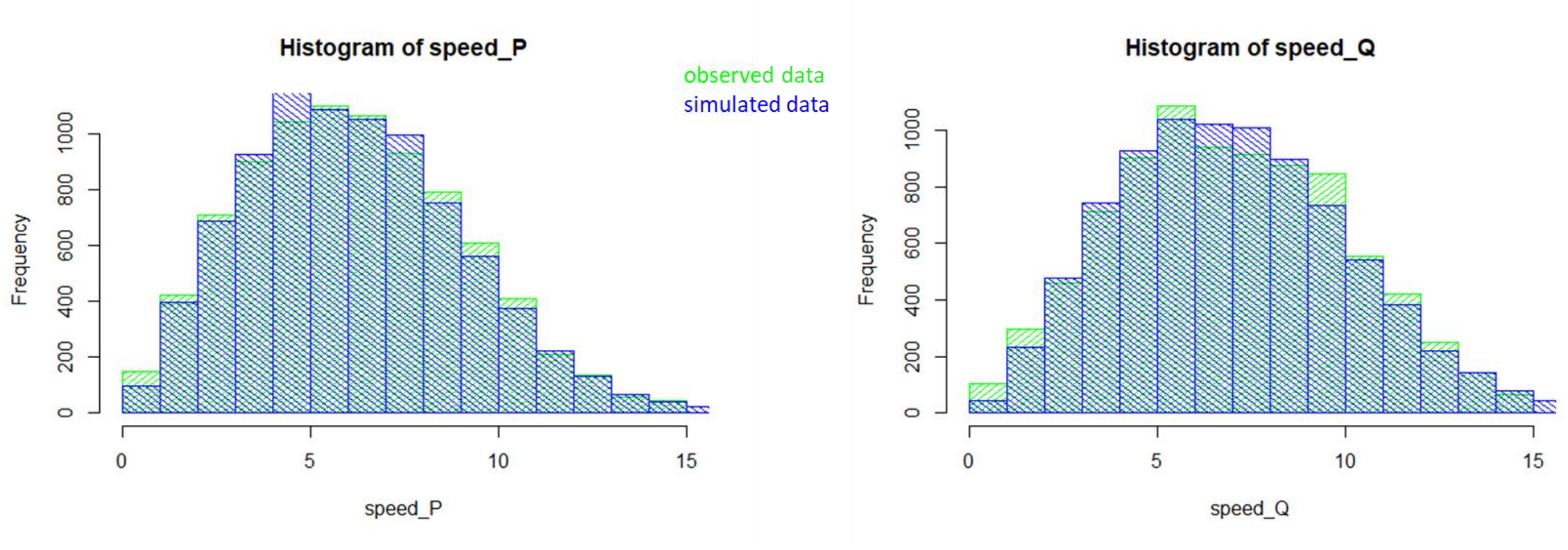

Figure 21. Weibull Distribution and Real Data Comparison

### 3.4.3.3 Copula Modeling of Joint Distribution

The next step of the proposed research is to examine the copula family for ‘PQ’ wind speed pair that was selected before. Table 4 provides the copula selection is BB8 copula for this case study, which is also known as Joe-Frank copula with estimates of $\theta$ and $\delta$ parameters are 2.27 and 0.9, respectively.

Firstly, we would like to see how the two wind speeds look like one with the other. Therefore, Figure 22 illustrates there is no linear relationship with two speeds.

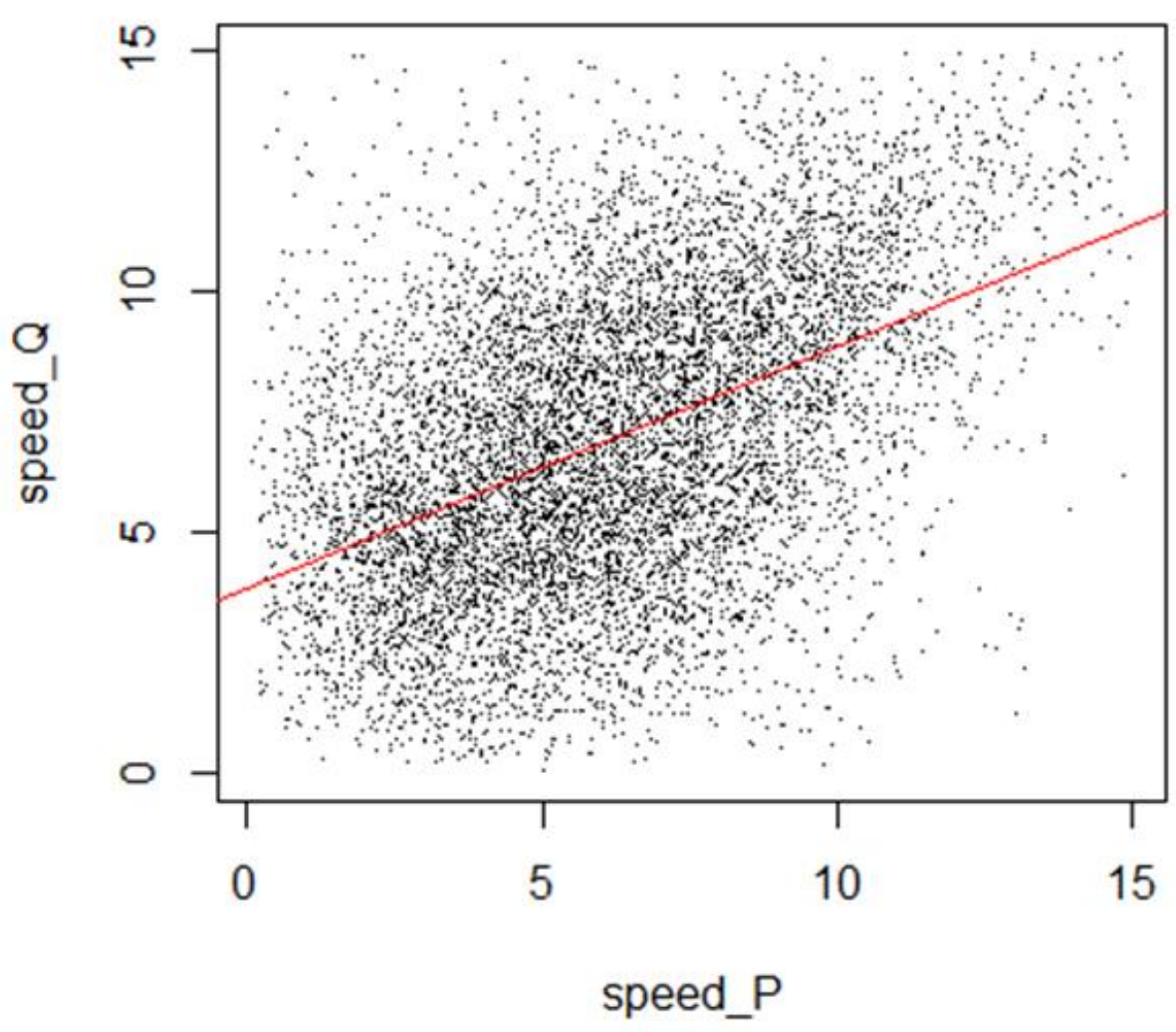

Figure 22. Data Visualization Before Copula Application

The copula is modeled using the estimated parameters from Table 4 (θ=2.27 and δ=0.9). The marginals were two Weibull distributions with different shape and scale parameters for ‘P’ and ‘Q’. The estimated shape and scale parameters are (2.26, 6.92) and (2.43, 7.79) for ‘P’ and ‘Q’ respectively. Figure 23 then shows us copula simulated data points. They are still in vector space; thus, values are in the [0,1] range. Our final goal from copula fitting is to investigate the accuracy of simulated samples from copula generated joint distribution and find the densities of the joint distribution.

Figures 24 and 25 depict the PDF and CDF of the fitted model to randomly drawn samples from Joe-Frank copula with estimates of θ and δ parameters are 2.27 and 0.9, respectively.

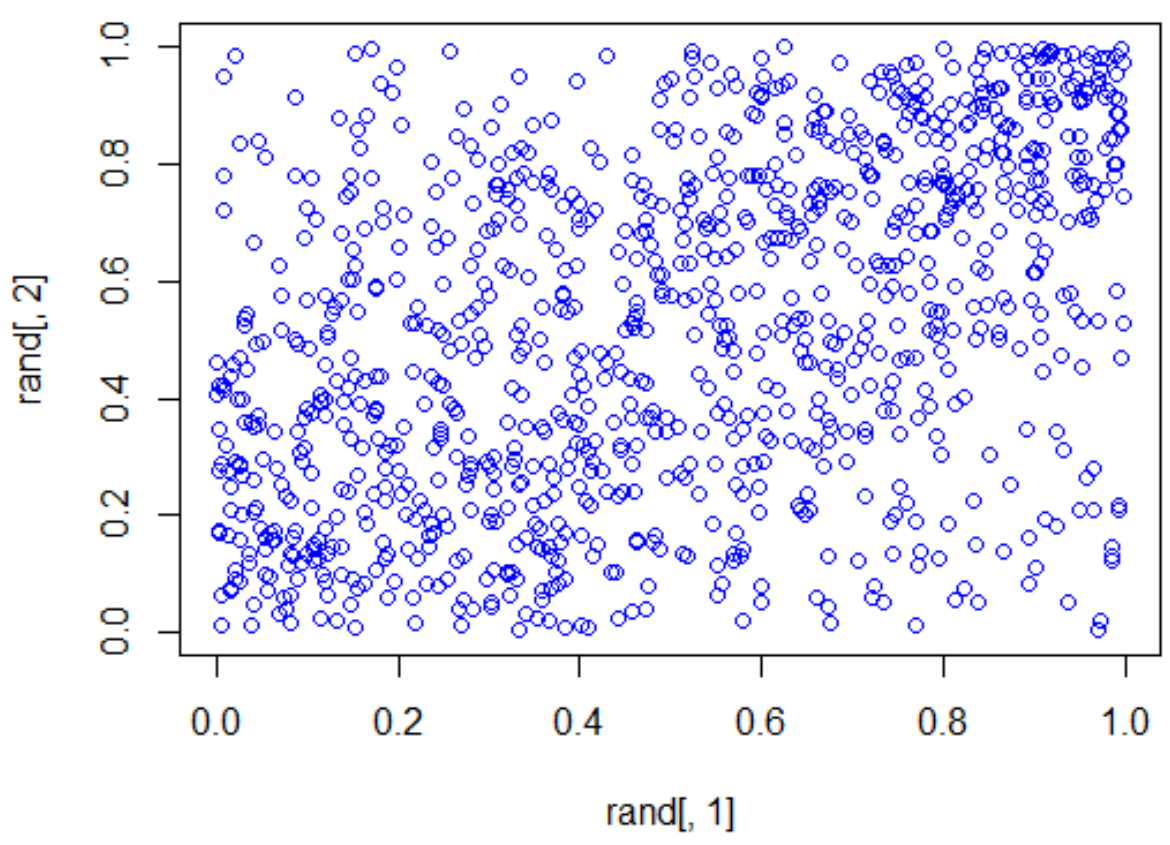

Figure 23. Random Datapoints Generation from BB8 Copula

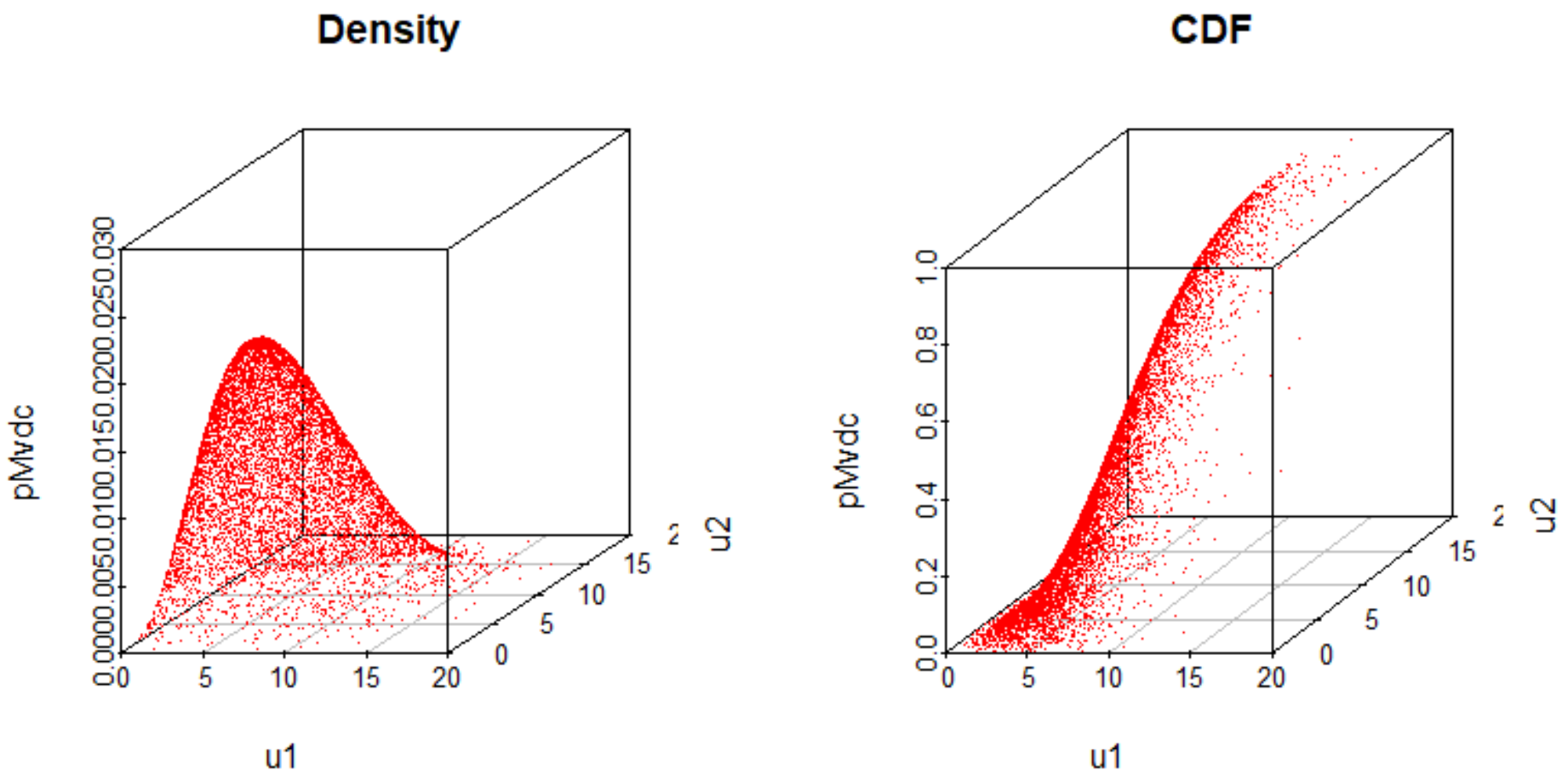

Figure 24. PDF and CDF Plots of Joe-Frank copula with estimated parameters (θ=2.27 and δ=0.9)

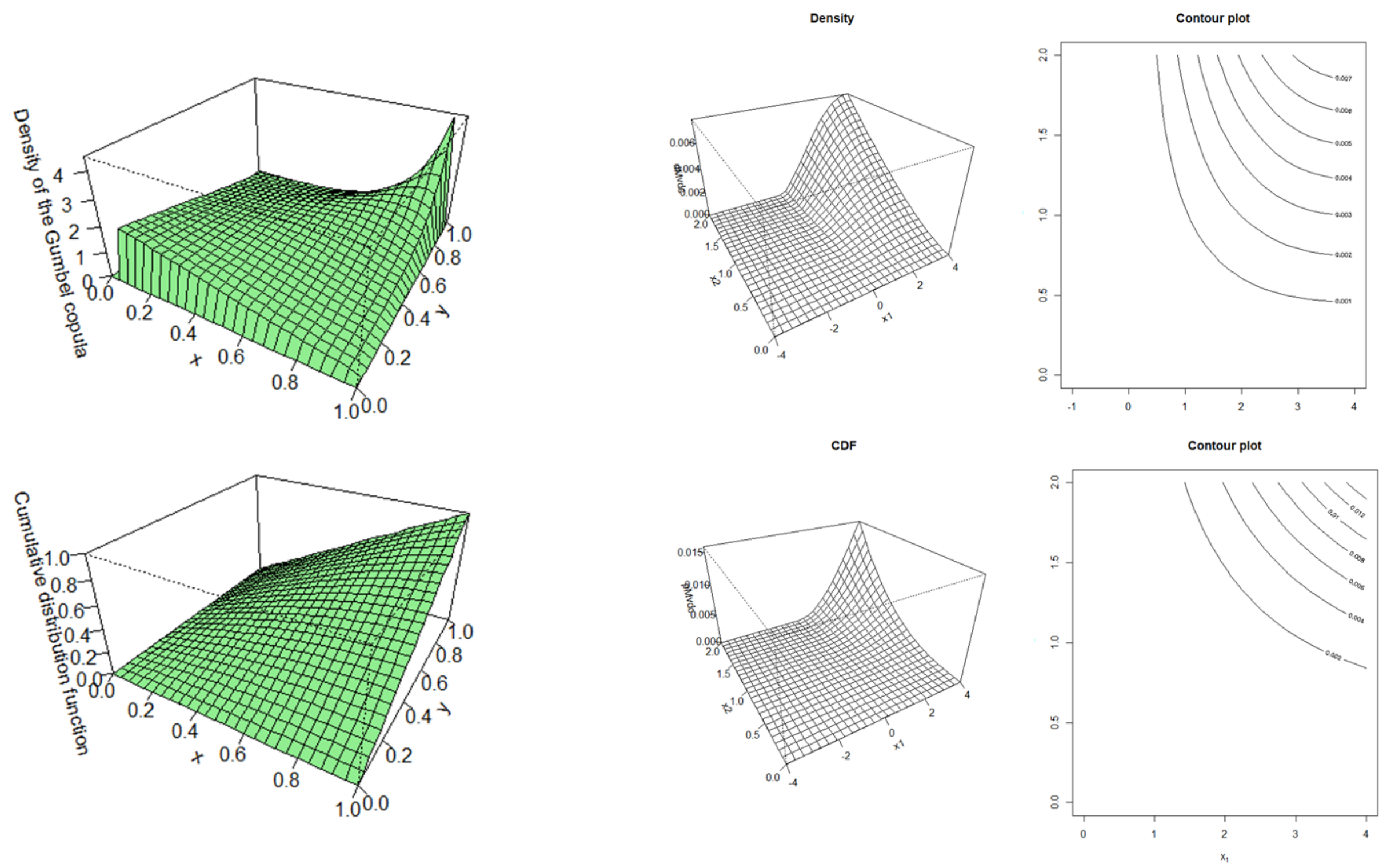

Figure 25. Density and Contour plots of Joe-Frank copula with estimated parameters (θ=2.27 and δ=0.9)

### 3.4.3.4 Performance Evaluation

In this section, we compare the simulated samples with real data and analyze the accuracy of the model. The plot is showing both data values in Figure 26. The joint distribution we have developed through copula application is producing similar data as real wind speeds. However, we are still interested to see how the correlation measure is and the standard error of correlations.

The correlation of simulated wind speed pair is 0.458241, while for the real data, it is 0.4582387. Thus, correlation measures are very close. Furthermore, the standard error is 0.0094, which is within the acceptable range (<0.05). This measure indicates, the confidence interval is within the true population correlation. Finally, Figure 27 compares the simulated data with real data in a Q-Q plot.  the true values and predicted values stay in a linear pattern that verifies the accuracy of the copula model.

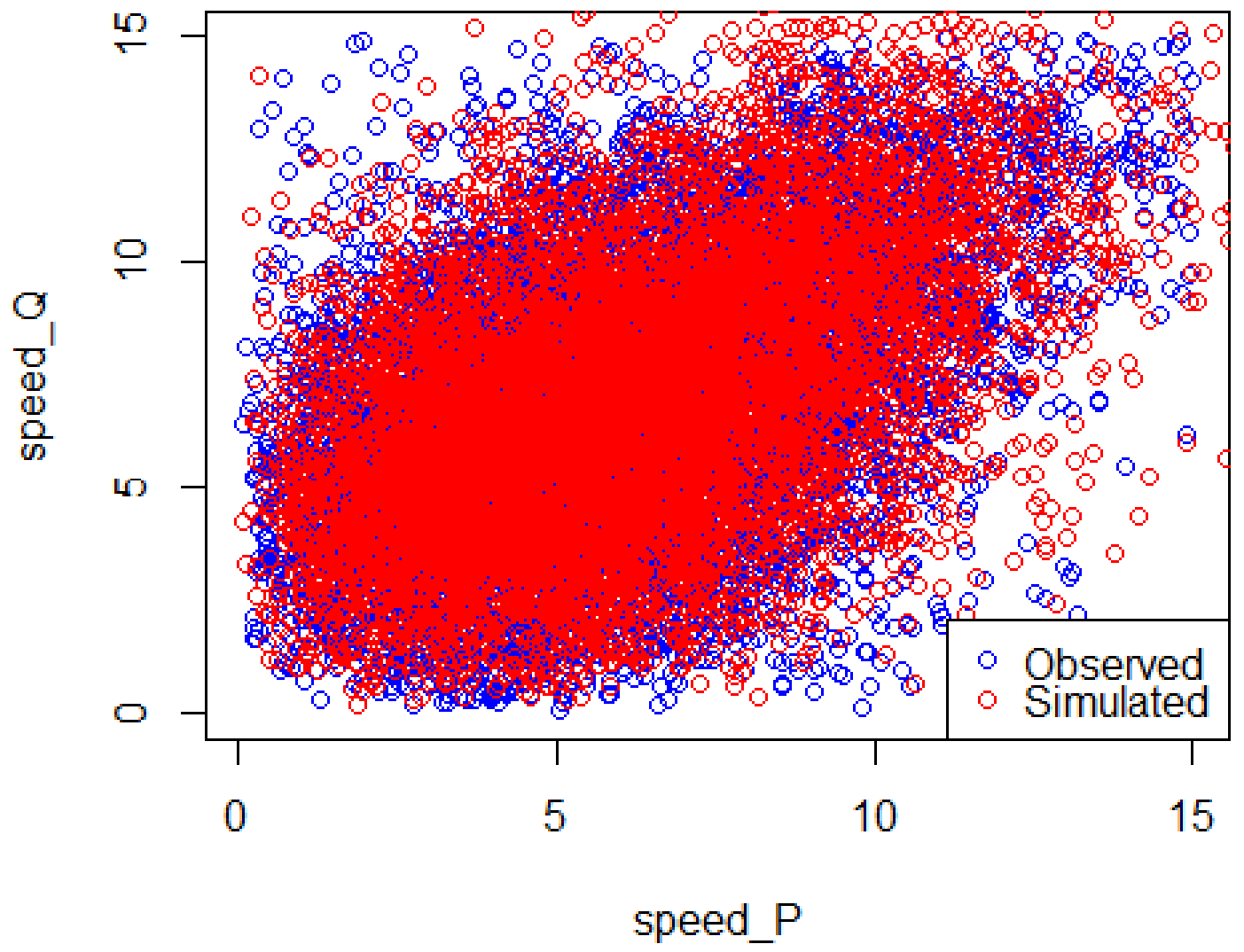

Figure 26. Comparison of Real and Simulated Data

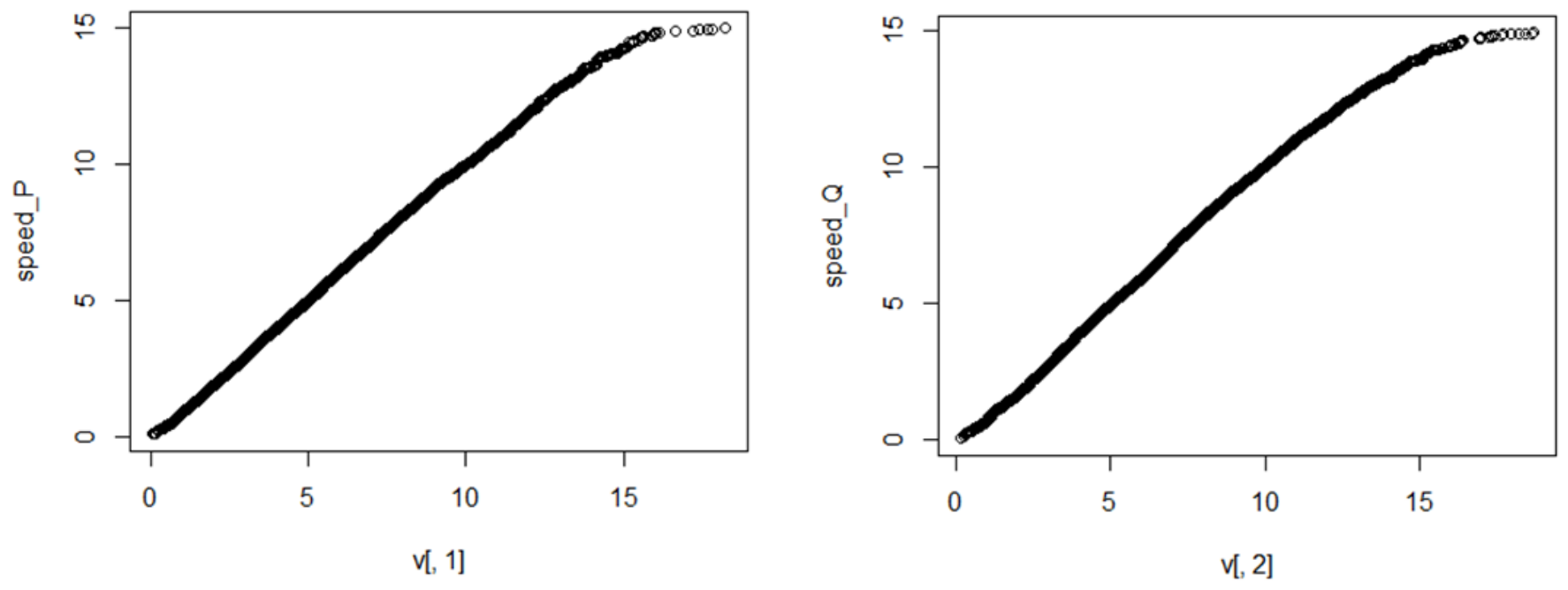

Figure 27. Copula Model Accuracy (Q-Q Plot)

## 3.5 Insights of Copula Model Uncertainty

Uncertainty analysis of any model is crucial for the reliability in a dynamic scenario. Like the statistical inference, uncertainty about parameters plays a vital role in characterizing generalized efficacy for copula dependency models. In this section, we will discuss some insights about the uncertainty aspects of wind speed copula models. Intuitively, there exists uncertainty when it comes to copula model selection, the uncertainty about the distance between two wind speed locations. Again, copula selection depends on the data, so the model uncertainty also encounters the variation from the data itself. Uncertainty about data comes from the sample size, and different data groups for a specific wind speed pair (different yearly data than one single year); thus, uncertainty about the marginal distributions. Although it is well established in the literature that Weibull distribution best describes the wind speed, however, the distribution parameters usually change. Therefore, how the uncertainty of wind speed distribution (or distribution parameters), that is, the marginal distributions in our case, affects the copula model. Besides, each copula family has its describing parameters, which are usually estimated empirically. Uncertainty about these parameters shall also affect the model's robustness. Therefore, a few of the areas to look at when it comes to the copula model uncertainty are-

1. Uncertainty of the distance between wind speed location on copula model selection
2. Uncertainty of marginal (wind speed) distributions on the copula model
3. Uncertainty about copula parameters
4. The overall uncertainty of the copula model: integrating 1, 2, and 3 (uncertainty from dynamically selected locations, with different wind speed profile, and the model itself)

Bayesian inference provides a rigorous methodology to incorporate the uncertainty about model parameters. The present computational capabilities inspire similar but complex uncertainty studies [139, 140]. In literature, the uncertainty of copula models incorporates a Bayesian method to find the posterior of copula joint distribution [141, 142].

As the name suggests, it leverages the Bayes rule (52) to updates belief/ hypothesis as evidence/ information becomes available [54, 143]. Three very important terminologies for are- “prior”, “likelihood”, and “posterior”. Prior is the distribution of an uncertain quantity, in other words, the probability of a hypothesis before considering any evidence or prior information. So, the prior probability is an unconditional probability distribution, let $P(A)$. The likelihood is the probability of observing new evidence, given the initial hypothesis, $P(B|A)$. Posterior is the updated probability of the quantity or hypothesis knowing some evidence. We denote posterior as $P(A|B)$. The total probability of observing the evidence $P(B)$ is the marginal likelihood and does not affect the posterior, so omitted in (54).

$$P(A|B) = \frac{P(B|A)P(A)}{P(B)} \tag{52}$$

Another way to express (52) is-

$$Posterior = \frac{Likelihood * Prior}{Marginal\ likelihood} \tag{53}$$

$$Posterior \propto Likelihood * Prior \qquad (544)$$

Now, if the prior and posterior in the model belongs to the same family of distribution, it is called conjugate. The conjugate prior eases the algebraic manipulation, and the mathematical approach is known as a closed-form expression. However, for non-conjugate prior, which is the case for the two copula families, the mathematical complexity is a burden. Markov Chain Monte Carlo (MCMC) simulation is then performed to analyze the posterior distribution [54]. Many software packages support Gibbs Sampling that can do MCMC simulation, a few examples are- JAGS, WinBUGS.

For the BB8 copula model, we need to apply Bayesian inference on $\theta$ and $\delta$ to find the posterior considering the uncertainty about the copula parameters. The prior distributions may be a gamma distribution for $\theta$ (>1) and uniform distribution for $\delta$ (0,1). However, existing libraries do not provide enough support for the MCMC simulation of BB8 copula. Therefore, further research is needed for this application.

# Chapter Four: Conclusion and Future Work

Climate change is one of the most concerning issues of this century. Emission from electric power generation is a crucial factor that drives the concern to the next level. However, renewable energy is getting wide attention as a solution to mitigate the effect. Renewable energy sources are widespread and available globally; however, one of the major challenges is to understand their characteristics in a more informative way.

In the first part of this research, we predicted wind speed at a height that is challenging to reach by using easy to access weather parameters. This research will be useful for wind farm planning and feasibility study. We investigated multiple artificial intelligence algorithms and concluded that LSTM outperformed other models with 97.8% prediction accuracy.

In the later part of the work, we have analyzed to find the best copula model and its parameter estimates for a set of wind speed pairs. The copula models found useful for wind speed dependency modeling are- Frank, Gumbel, Joe, Gaussian, Clayton, Clayton-Gumbel (BB1), Joe-Gumbel (BB6), Joe-Clayton (BB7), and Joe-Frank (BB8). For our case study, we selected BB8 copula by using the Bayesian Information Criterion (BIC) and found the proposed joint distribution can model the wind speed joint distribution accurately, with a standard error of 0.0094.

Later, we discussed the uncertainty aspects of wind speed copula dependency modeling intuitively with some directions for future researchers.

One of the major extensions of this research would be the copula uncertainty modeling for wind speed dependency. The copula modeling could be extended to wind-solar, wind-load, or multivariate wind-solar-load dependency structure realization.

In addition, the outcomes of this research will be useful for economic and risk analysis of wind power planning. The work could also be used to model a robust wind power controlling scheme, that would later be useful for wind power grid integration.

# Appendix

## Appendix A: Copula Family identifier in VineCopula

0 = independence copula
1 = Gaussian copula
2 = Student t copula (t-copula)
3 = Clayton copula
4 = Gumbel copula
5 = Frank copula
6 = Joe copula
7 = BB1 copula
8 = BB6 copula
9 = BB7 copula
10 = BB8 copula
13 = rotated Clayton copula (180 degrees; survival Clayton") \cr `14` = rotated Gumbel copula (180 degrees; survival Gumbel")
16 = rotated Joe copula (180 degrees; survival Joe") \cr `17` = rotated BB1 copula (180 degrees; survival BB1")
18 = rotated BB6 copula (180 degrees; survival BB6")\cr `19` = rotated BB7 copula (180 degrees; survival BB7")
20 = rotated BB8 copula (180 degrees; "survival BB8")
23 = rotated Clayton copula (90 degrees)
'24' = rotated Gumbel copula (90 degrees)
'26' = rotated Joe copula (90 degrees)
'27' = rotated BB1 copula (90 degrees)
'28' = rotated BB6 copula (90 degrees)
'29' = rotated BB7 copula (90 degrees)
'30' = rotated BB8 copula (90 degrees)
'33' = rotated Clayton copula (270 degrees)
'34' = rotated Gumbel copula (270 degrees)
'36' = rotated Joe copula (270 degrees)
'37' = rotated BB1 copula (270 degrees)
'38' = rotated BB6 copula (270 degrees)
'39' = rotated BB7 copula (270 degrees)
'40' = rotated BB8 copula (270 degrees)
'104' = Tawn type 1 copula
'114' = rotated Tawn type 1 copula (180 degrees)
'124' = rotated Tawn type 1 copula (90 degrees)
'134' = rotated Tawn type 1 copula (270 degrees)
'204' = Tawn type 2 copula
'214' = rotated Tawn type 2 copula (180 degrees)
'224' = rotated Tawn type 2 copula (90 degrees)
'234' = rotated Tawn type 2 copula (270 degrees)

# Appendix B: Copula Selections for different time average wind speeds

| WS | 6 Hourly | | Daily | | Weekly | |
|---|---|---|---|---|---|---|
| | **Corr** | **Bivariate Copula "BIC"** | **Corr** | **Bivariate Copula "BIC"** | **Corr** | **Bivariate Copula "BIC"** |
| EF | 0.9979284 | Frank (par = 100, tau = 0.96) | 0.9987454 | Survival Gumbel (par = 61.25, tau = 0.98) | 0.9974198 | Survival Gumbel (par = 22.79, tau = 0.96) |
| JK | 0.9910234 | t (par = 0.99, par2 = 8, tau = 0.91) | 0.9930283 | Survival Gumbel (par = 12.7, tau = 0.92) | 0.9858894 | Frank (par = 36.87, tau = 0.9) |
| OP | 0.9678966 | t (par = 0.97, par2 = 8, tau = 0.85) | 0.9790544 | Gumbel (par = 7.62, tau = 0.87) | 0.9887115 | Gaussian (par = 0.99, tau = 0.9) |
| KL | 0.9847116 | Frank (par = 35.07, tau = 0.89) | 0.9899218 | Gaussian (par = 0.99, tau = 0.9) | 0.9810514 | Gaussian (par = 0.98, tau = 0.88) |
| JL | 0.970002 | Gaussian (par = 0.97, tau = 0.83) | 0.9773925 | Gaussian (par = 0.98, tau = 0.86) | 0.9589582 | Gaussian (par = 0.96, tau = 0.82) |
| EG | 0.9707373 | Frank (par = 24.31, tau = 0.85) | 0.9828112 | Frank (par = 31.73, tau = 0.88) | 0.9822609 | Survival Gumbel (par = 8.17, tau = 0.88) |
| GH | 0.9744947 | t (par = 0.97, par2 = 6.48, tau = 0.85) | 0.9862989 | Gaussian (par = 0.99, tau = 0.89) | 0.9838736 | Survival Gumbel (par = 9.83, tau = 0.9) |
| FG | 0.9697138 | Frank (par = 24.08, tau = 0.85) | 0.9832993 | Frank (par = 32.3, tau = 0.88) | 0.9828254 | Survival Gumbel (par = 8.12, tau = 0.88) |
| IJ | 0.9772875 | t (par = 0.98, par2 = 5.32, tau = 0.86) | 0.9890961 | Frank (par = 41.4, tau = 0.91) | 0.9894372 | Gaussian (par = 0.99, tau = 0.91) |
| HI | 0.9707448 | Frank (par = 25.42, tau = 0.85) | 0.9818121 | Gaussian (par = 0.98, tau = 0.88) | 0.9883083 | Survival Gumbel (par = 11.83, tau = 0.92) |
| I K | 0.9667266 | Frank (par = 23.8, tau = 0.84) | 0.9814868 | Frank (par = 31.14, tau = 0.88) | 0.9775843 | Frank (par = 28.45, tau = 0.87) |
| I L | 0.9493195 | t (par = 0.95, par2 = 16.34, tau = 0.79) | 0.966746 | Gaussian (par = 0.96, tau = 0.83) | 0.9560555 | Gaussian (par = 0.96, tau = 0.81) |
| E H | 0.9525109 | Frank (par = 19.44, tau = 0.81) | 0.9755204 | Frank (par = 27.52, tau = 0.86) | 0.9766973 | Survival Gumbel (par = 7.54, tau = 0.87) |
| F H | 0.9514789 | Frank (par = 19.3, tau = 0.81) | 0.9756085 | Frank (par = 27.2, tau = 0.86) | 0.9747621 | Survival Gumbel (par = 6.81, tau = 0.85) |
| H J | 0.9437773 | Frank (par = 17.76, tau = 0.8) | 0.9678831 | Gaussian (par = 0.97, tau = 0.83) | 0.9721819 | Survival Gumbel (par = 7.48, tau = 0.87) |
| G I | 0.9508921 | t (par = 0.95, par2 = 6.96, tau = 0.8) | 0.9731382 | t (par = 0.97, par2 = 8, tau = 0.85) | 0.9843574 | Survival Gumbel (par = 9.05, tau = 0.89) |
| H K | 0.9309822 | Frank (par = 15.63, tau = 0.77) | 0.9583183 | Gaussian (par = 0.96, tau = 0.81) | 0.9636349 | Gaussian (par = 0.96, tau = 0.83) |
| H L | 0.9166687 | Frank (par = 13.74, tau = 0.74) | 0.9451536 | Gaussian (par = 0.94, tau = 0.79) | 0.9479923 | Gaussian (par = 0.95, tau = 0.8) |
| E I | 0.9148547 | Frank (par = 13.7, tau = 0.74) | 0.9501874 | Frank (par = 18.27, tau = 0.8) | 0.9608128 | Survival Gumbel (par = 5.99, tau = 0.83) |
| G J | 0.9229158 | Frank (par = 15.17, tau = 0.76) | 0.9605679 | Gaussian (par = 0.96, tau = 0.81) | 0.9791163 | Survival Gumbel (par = 8.22, tau = 0.88) |
| F I | 0.914169 | Frank (par = 13.61, tau = 0.74) | 0.9508437 | Gaussian (par = 0.94, tau = 0.79) | 0.9608128 | Survival Gumbel (par = 5.66, tau = 0.82) |
| G K | 0.91464 | Frank (par = 14.14, tau = 0.75) | 0.9560879 | Gaussian (par = 0.95, tau = 0.8) | 0.961861 | Survival Gumbel (par = 5.62, tau = 0.82) |
| E J | 0.8854571 | Frank (par = 11.45, tau = 0.7) | 0.9357043 | Frank (par = 15.8, tau = 0.77) | 0.952427 | Survival Gumbel (par = 5.62, tau = 0.82) |
| F J | 0.8857889 | Frank (par = 11.46, tau = 0.7) | 0.9378113 | Frank (par = 16.07, tau = 0.78) | 0.9519432 | Survival Gumbel (par = 5.45, tau = 0.82) |
| GL | 0.8970869 | BB1 (par = 0.19, par2 = 2.99, tau = 0.69) | 0.9413653 | Gaussian (par = 0.94, tau = 0.77) | 0.9392034 | Gaussian (par = 0.95, tau = 0.79) |
| EK | 0.8712044 | Frank (par = 10.45, tau = 0.68) | 0.9260555 | Gaussian (par = 0.92, tau = 0.74) | 0.9281567 | Survival Gumbel (par = 4.22, tau = 0.76) |
| FK | 0.8709256 | Frank (par = 10.42, tau = 0.68) | 0.9279412 | Gaussian (par = 0.92, tau = 0.74) | 0.929205 | Survival Gumbel (par = 4.21, tau = 0.76) |
| EL | 0.8485385 | Frank (par = 9.33, tau = 0.65) | 0.9083632 | Gaussian (par = 0.9, tau = 0.71) | 0.9052572 | Survival Gumbel (par = 3.92, tau = 0.74) |
| FL | 0.8474945 | Gaussian (par = 0.83, tau = 0.63) | 0.9091313 | Gaussian (par = 0.9, tau = 0.71) | 0.9039671 | Survival Gumbel (par = 3.84, tau = 0.74) |
| EN | 0.8509871 | BB1 (par = 0.14, par2 = 2.62, tau = 0.64) | 0.9075329 | t (par = 0.91, par2 = 11.44, tau = 0.73) | 0.9299307 | Gaussian (par = 0.94, tau = 0.77) |
| EM | 0.5946354 | Gumbel (par = 1.65, tau = 0.39) | 0.6784139 | Gumbel (par = 1.91, tau = 0.48) | 0.7809224 | Gumbel (par = 2.47, tau = 0.6) |
| OQ | 0.5281529 | Gumbel (par = 1.58, tau = 0.37) | 0.6064848 | Gumbel (par = 1.8, tau = 0.45) | 0.8020481 | Gumbel (par = 2.51, tau = 0.6) |
| PQ | 0.5222581 | Gumbel (par = 1.56, tau = 0.36) | 0.6037476 | Gumbel (par = 1.77, tau = 0.44) | 0.8077729 | Gumbel (par = 2.65, tau = 0.62) |
| BC | 0.4117035 | Gumbel (par = 1.36, tau = 0.27) | 0.4652312 | Gumbel (par = 1.46, tau = 0.31) | 0.5990969 | Gaussian (par = 0.63, tau = 0.43) |
| AB | 0.1918313 | Gaussian (par = 0.2, tau = 0.13) | 0.1477147 | Gaussian (par = 0.17, tau = 0.11) | 0.4540397 | Rotated Tawn type 1 180 degrees (par = 3.69, par2 = 0.19, tau = 0.17) |
| AC | 0.1208388 | Frank (par = 0.73, tau = 0.08) | 0.07209759 | Independence (par = 0, tau = 0) | 0.2750363 | Frank (par = 1.87, tau = 0.2) |
| AD | 0.1542902 | Gaussian (par = 0.16, tau = 0.1) | 0.1639981 | Gumbel (par = 1.12, tau = 0.11) | 0.4181584 | Gaussian (par = 0.49, tau = 0.33) |
| BD | 0.07317099 | Survival Gumbel (par = 1.05, tau = 0.05) | -0.0063046 8 | Independence (par = 0, tau = 0) | -0.01467505 | Independence (par = 0, tau = 0) |
| CD | -0.02745832 | Independence (par = 0, tau = 0) | -0.1221796 | Independence (par = 0, tau = 0) | -0.2405257 | Gaussian (par = -0.31, tau = -0.2) |

# Appendix C: Wind Speed Prediction Code (Python 3.7)

```
# Data Preprocessing

# Importing the libraries

import numpy as np
import matplotlib.pyplot as plt
import matplotlib
import pandas as pd
import sklearn
import scipy as spy
import seaborn as sns
import statsmodels.api as sm
import math
from patsy import dmatrices
from statsmodels.stats.outliers_influence import variance_inflation_factor
import warnings
warnings.filterwarnings("ignore")
from sklearn.svm import SVC
from sklearn.metrics import confusion_matrix
from sklearn.metrics import mean_squared_error, r2_score, mean_absolute_error,
median_absolute_error, explained_variance_score

# Importing the dataset

data = pd.read_csv('wind_data_v2_22_3.csv')
data = data.iloc[:, 1:12]
data_describe = (data.describe())

'''#fiture histogram
data.hist(grid = True, figsize=[30,20])'''
#correlation
corr = data.corr()

corr.to_csv('corr_mat.csv')
corr_mat = corr.iloc[:, -1]
# plot the heatmap
sns.heatmap(corr,
        xticklabels=corr.columns,
        yticklabels=corr.columns)

values = data.values
```

```
# specify columns to plot
groups = [0, 1, 2, 3, 4, 5, 6, 7]
i = 1
# plot each column
plt.figure(figsize = [30,20])
for group in groups:
	plt.subplot(len(groups), 1, i)
	plt.plot(values[:, group])
	plt.title(data.columns[group], y=0.5, loc='right')
	i += 1
plt.show()

#pairplot
sns.set(style="ticks", color_codes=True)
sns.pairplot(data)


#Define X and y variables

X = data.iloc[:, 0:7]
y = data.iloc[:,-1]

#Nornmalization
headers = list(X.columns.values)

# Feature Scaling
from sklearn import preprocessing
data_scaler = preprocessing.MinMaxScaler(feature_range=(0, 1))
X = data_scaler.fit_transform(X)

X = pd.DataFrame(X,columns=headers)

#splittind the dataset into training set and test set
from sklearn.model_selection import train_test_split
X_train, X_test, y_train, y_test = train_test_split(X, y, test_size = 0.2, random_state=0)

#Linear Regression

# Fitting Multiple Linear Regression to the Training set
from sklearn.linear_model import LinearRegression
regressor = LinearRegression()
regressor.fit(X_train, y_train)
y_pred = regressor.predict(X_test)
LinearReg_Coef = regressor.coef_
print(LinearReg_Coef)
# linear regression visualize
```

```
plt.scatter(y_test,y_pred)
plt.title('Visualize the prediction (Linear regression)')
plt.xlabel('Actual y')
plt.ylabel('Predicted y')
plt.show()

#regression score, performance/accuracy
print ("##Linear regression Accuracy##")
print ("Mean absolute error =", round(mean_absolute_error(y_test, y_pred), 3))
print ("Mean squared error =", round(mean_squared_error(y_test, y_pred), 3))
print ("Median absolute error =", round(median_absolute_error(y_test, y_pred), 3))
print ("Explained variance score =", round(explained_variance_score(y_test, y_pred), 3))
print ("R2 score =", round(r2_score(y_test, y_pred), 3))

res =y_test-y_pred

#residual plot
sns.set(color_codes=True)
sns.residplot(y_pred, res, lowess=True, color="b")
# For each X, calculate VIF and save in dataframe
vif = pd.DataFrame()
vif["VIF Factor"] = [variance_inflation_factor(X, i) for i in range(X[1])]
vif.round(1)

# ridge regression
from sklearn.linear_model import Ridge
rr = Ridge(alpha=0.01)
rr.fit(X_train, y_train)
y_pred_rr = rr.predict(X_test)
RidgeReg_Coef = rr.coef_
print(RidgeReg_Coef)

#Visualize the prediction
plt.scatter(y_test,y_pred_rr)
plt.title('Visualize the prediction (Ridge (alpha=0.01))')
plt.xlabel('Actual y')
plt.ylabel('Predicted y')
plt.show()

#performance/accuracy
print ("##Ridge (alpha=0.01) Accuracy##")
print ("Mean absolute error =", round(mean_absolute_error(y_test, y_pred_rr), 3))
print ("Mean squared error =", round(mean_squared_error(y_test, y_pred_rr), 3))
print ("Median absolute error =", round(median_absolute_error(y_test, y_pred_rr), 3))
print ("Explained variance score =", round(explained_variance_score(y_test, y_pred_rr), 3))
print ("R2 score =", round(r2_score(y_test, y_pred_rr), 3))
```

```
#Ridge (aplha=100)
rr100 = Ridge(alpha=1000)
rr100.fit(X_train, y_train)
y_pred_100 = rr100.predict(X_test)
RidgeReg_Coef100 = rr100.coef_
print(RidgeReg_Coef100)
#Visualize the prediction
plt.scatter(y_test,y_pred_100)
plt.title('Visualize the prediction (Ridge (aplha=100))')
plt.xlabel('Actual y')
plt.ylabel('Predicted y')
plt.show()

print ("##Ridge (alpha=100) Accuracy##")
print ("Mean absolute error =", round(mean_absolute_error(y_test, y_pred_100), 3))
print ("Mean squared error =", round(mean_squared_error(y_test, y_pred_100), 3))
print ("Median absolute error =", round(median_absolute_error(y_test, y_pred_100), 3))
print ("Explained variance score =", round(explained_variance_score(y_test, y_pred_100), 3))
print ("R2 score =", round(r2_score(y_test, y_pred_100), 3))

#LASSO
from sklearn.linear_model import Lasso
lasso = Lasso()
lasso.fit(X_train,y_train)
y_pred_lasso = lasso.predict(X_test)
coeff_used = np.sum(lasso.coef_!=0)
LASSO_Coef = lasso.coef_
print(LASSO_Coef)
#Visualize the prediction
plt.scatter(y_test,y_pred_lasso)
plt.title('Visualize the prediction (LASSO)')
plt.xlabel('Actual y')
plt.ylabel('Predicted y')
plt.show()

print ("##LASSO Accuracy##")
print ("Mean absolute error =", round(mean_absolute_error(y_test, y_pred_lasso), 3))
print ("Mean squared error =", round(mean_squared_error(y_test, y_pred_lasso), 3))
print ("Median absolute error =", round(median_absolute_error(y_test, y_pred_lasso), 3))
print ("Explained variance score =", round(explained_variance_score(y_test, y_pred_lasso), 3))
print ("R2 score =", round(r2_score(y_test, y_pred_lasso), 3))

#Lasso with aplha 0.01
lasso001 = Lasso(alpha=0.01, max_iter=10e5)
lasso001.fit(X_train,y_train)
```

```
y_pred_lasso001 = lasso001.predict(X_test)
coeff_used001 = np.sum(lasso001.coef_!=0)
LASSO_Coef001 = lasso001.coef_
print(LASSO_Coef001)
#Visualize the prediction
plt.scatter(y_test,y_pred_lasso001)
plt.title('Visualize the prediction (LASSO (alpha=0.01))')
plt.xlabel('Actual y')
plt.ylabel('Predicted y')
plt.show()

print ("##LASSO (alpha=0.01) Accuracy##")
print ("Mean absolute error =", round(mean_absolute_error(y_test, y_pred_lasso001), 3))
print ("Mean squared error =", round(mean_squared_error(y_test, y_pred_lasso001), 3))
print ("Median absolute error =", round(median_absolute_error(y_test, y_pred_lasso001), 3))
print ("Explained variance score =", round(explained_variance_score(y_test, y_pred_lasso001),
3))
print ("R2 score =", round(r2_score(y_test, y_pred_lasso001), 3))

#LASSO alpha = 0.0001
lasso00001 = Lasso(alpha=0.0001, max_iter=10e5)
lasso00001.fit(X_train,y_train)
y_pred_lasso00001 = lasso00001.predict(X_test)
coeff_used00001 = np.sum(lasso00001.coef_!=0)
LASSO_Coef00001 = lasso00001.coef_
print(LASSO_Coef00001)
#Visualize the prediction
plt.scatter(y_test,y_pred_lasso00001)
plt.title('Visualize the prediction (LASSO (alpha=0.0001))')
plt.xlabel('Actual y')
plt.ylabel('Predicted y')
plt.show()

print ("##LASSO (alpha=0.0001) Accuracy##")
print ("Mean absolute error =", round(mean_absolute_error(y_test, y_pred_lasso00001), 3))
print ("Mean squared error =", round(mean_squared_error(y_test, y_pred_lasso00001), 3))
print ("Median absolute error =", round(median_absolute_error(y_test, y_pred_lasso00001), 3))
print ("Explained variance score =", round(explained_variance_score(y_test,
y_pred_lasso00001), 3))
print ("R2 score =", round(r2_score(y_test, y_pred_lasso00001), 3))
…………………………………………………………………………………………………………
# Data Preprocessing

# Importing the libraries
from __future__ import print_function
```

```
import numpy as np
from numpy import nan as NA
import matplotlib
import matplotlib.pyplot as plt
import pandas as pd
import sklearn
from sklearn.model_selection import cross_val_score
from sklearn.model_selection import KFold
from sklearn.preprocessing import StandardScaler
from sklearn.pipeline import Pipeline
from sklearn.svm import SVC
from sklearn.metrics import confusion_matrix
from sklearn.metrics import mean_squared_error, r2_score, mean_absolute_error,
median_absolute_error, explained_variance_score
import scipy as spy
import seaborn as sns
import statsmodels.api as sm
import math
from math import sqrt
from keras.models import Sequential
from keras.layers.core import Activation, Dense, Dropout, RepeatVector, SpatialDropout1D

from keras.optimizers import SGD, RMSprop, Adam
from keras.utils import np_utils
from keras.wrappers.scikit_learn import KerasRegressor
from keras.callbacks import ModelCheckpoint
from keras.utils import np_utils
import warnings
warnings.filterwarnings("ignore")

from numpy import concatenate
from pandas import concat

np.random.seed(42)    # setting seed before importing from Keras
from keras.layers.embeddings import Embedding
from keras.layers.recurrent import LSTM
from keras.layers.wrappers import TimeDistributed

from keras.preprocessing import sequence

import collections
import os

# Importing the dataset
data = pd.read_csv('wind_data_v2_22_1.csv')
data = data.iloc[:, 1:12]
```

```
data_describe = (data.describe())

#Define X and y variables

X = data.iloc[:, 0:10]
y = data.iloc[:,-1]

headers = list(X.columns.values)

# Feature Scaling
from sklearn import preprocessing
data_scaler = preprocessing.MinMaxScaler(feature_range=(0, 1))
X = data_scaler.fit_transform(X)

X = pd.DataFrame(X,columns=headers)
X = X.astype('float32')

X.ndim
type(X)

X = X.values

X.ndim
type(X)
# reshape input to be 3D [samples, timesteps, features]
X = np.reshape(X, (X.shape[0], 1, 10))

X.ndim
type(X)

#splittind the dataset into training set and test set
from sklearn.model_selection import train_test_split
X_train, X_test, y_train, y_test = train_test_split(X, y, test_size = 0.2, random_state=0)
print(X_train.shape, y_train.shape, X_test.shape, y_test.shape)

# design network
model = Sequential()
model.add(LSTM(50, input_shape=(X_train.shape[1], X_train.shape[2])))

# The Output Layer :
model.add(Dense(6, kernel_initializer='normal',activation='relu'))
model.add(Dense(6, kernel_initializer='normal',activation='relu'))

model.add(Dense(1, kernel_initializer='normal',activation='linear'))
model.summary()
```

```
# Compile the network :
model.compile(loss='mean_absolute_error',
            optimizer='adam',
            metrics=['mean_squared_error', 'mean_absolute_error'])

# fit network
history = model.fit(X_train, y_train,
                 epochs=500, batch_size=1,
                 validation_split = 0.25,
                 verbose=1)

y_pred=model.predict(X_test, batch_size=1, verbose=1)

plt.scatter(y_test,y_pred)
plt.title('Visualize the prediction (LSTM)')
plt.xlabel('Actual y')
plt.ylabel('Predicted y')
plt.show()

print ("##LSTM Accuracy##")
print ("Mean absolute error =", round(mean_absolute_error(y_test, y_pred), 3))
print ("Mean squared error =", round(mean_squared_error(y_test, y_pred), 3))
print ("Median absolute error =", round(median_absolute_error(y_test, y_pred), 3))
print ("Explained variance score =", round(explained_variance_score(y_test, y_pred), 3))
print ("R2 score =", round(r2_score(y_test, y_pred), 3))

# list all data in history
print(history.history.keys())
# plot history
plt.plot(history.history['loss'], label='train')
plt.plot(history.history['val_loss'], label='test')
plt.title('model loss')
plt.ylabel('loss')
plt.xlabel('epoch')
plt.legend()
plt.show()

# plot history
plt.plot(history.history['mean_squared_error'], label='train')
plt.plot(history.history['val_mean_squared_error'], label='test')
plt.title('MSE')
plt.ylabel('MSE')
plt.xlabel('epoch')
plt.legend()
plt.show()
```

```
# plot history
plt.plot(history.history['mean_absolute_error'], label='train')
plt.plot(history.history['val_mean_absolute_error'], label='test')
plt.title('MAE')
plt.ylabel('MAE')
plt.xlabel('epoch')
plt.legend()
plt.show()
```

……………………………………….skipped codes…………………………………………

# Appendix D: Distribution Fitting Code (R)

```
library(fitdistrplus)
library(EnvStats)
library (lognorm)
library (readr)
library(repmis)
library(psych)
library (MCMCpack)
library(CholWishart)
set.seed(100)

#import data
speed_PQ <-
source_data("https://github.com/amimulehsan/wind_data/blob/master/PQ_speed.csv?raw=True")
speed_P <- (speed_PQ)$V2
speed_Q <- (speed_PQ)$V3

#distribution
my_data <- speed_Q    #change variable name
plot(my_data, pch=20)

#data description
summary(my_data)

summary(speed_P)
summary(speed_Q)

#plot the empirical density and the histogram
plotdist(my_data, histo = TRUE, demp = TRUE)

#descriptive statistics to help in making a decision
descdist(my_data, discrete=FALSE, boot=500)

#fitting a distribution (add more distribution as intended)
fit_w  <- fitdist(my_data, "weibull")
fit_g  <- fitdist(my_data, "gamma")
fit_ln <- fitdist(my_data, "lnorm")
#fit_ln <- fitdist(my_data, "Wishart")

#dWishart(x = my_data, df = 10, log=TRUE)

#summary of distribution fit
summary(fit_w)
summary(fit_g)
```

```
summary(fit_ln)

#plot the distribution fit results
par(mfrow=c(2,2))
plot.legend <- c("Weibull", "lognormal", "gamma")
denscomp(list(fit_w, fit_g, fit_ln), legendtext = plot.legend)
cdfcomp (list(fit_w, fit_g, fit_ln), legendtext = plot.legend)
qqcomp  (list(fit_w, fit_g, fit_ln), legendtext = plot.legend)
ppcomp  (list(fit_w, fit_g, fit_ln), legendtext = plot.legend)


par(mfrow=c(2,2))
plot.legend <- c("Weibull")
denscomp(list(fit_w), legendtext = plot.legend)
cdfcomp (list(fit_w), legendtext = plot.legend)
qqcomp  (list(fit_w), legendtext = plot.legend)
ppcomp  (list(fit_w), legendtext = plot.legend)

#distribution parameters estimation and compare with summary fit
eweibull(speed_Q)
estimateParmsLognormFromSample(speed_Q)

#marginal distribution fit with real data
hist(speed_Q, breaks = 20, col = "green", density = 20)
hist(rweibull(NROW(speed_Q), shape = 2.428298, scale = 7.790141), breaks = 20,col = "blue",
add = T, density = 20, angle = -45)
```

# Appendix E: Copula Modeling Code (R)

```
library(MASS)
library(rgl) # interactive 3-D plots
library(copula)
library(VineCopula)
library("scatterplot3d")
library(compositions)
library(RCurl)
library (readr)
library(repmis)
set.seed(500)

#import data
speed_PQ <-
source_data("https://github.com/amimulehsan/wind_data/blob/master/PQ_speed.csv?raw=True")
speed_P <- (speed_PQ)$V2
speed_Q <- (speed_PQ)$V3

#data description
describe(speed_P)
describe(speed_Q)

#correlation of data variables
cor(speed_P,speed_Q,method='spearman')

#plot data with regression line
plot(speed_P, speed_Q,pch='.')
abline(lm(speed_Q~speed_P),col='red',lwd=1)

#copula Selection
u <- pobs(as.matrix(cbind(speed_P,speed_Q)))[,1]
v <- pobs(as.matrix(cbind(speed_P,speed_Q)))[,2]
selectedCopula <- BiCopSelect(u,v,familyset=NA, selectioncrit = "BIC")
selectedCopula
summary(selectedCopula)

#fit the suggested model
g.cop <- BB8Copula()
m <- pobs(as.matrix(cbind(speed_P,speed_Q)))
fit <- fitCopula(g.cop,m)
coef(fit)

#build the copula for positive dependence
set.seed(100)
rand <- rCopula(1000, BB8Copula(param = c(2.267418, 0.897979 )))
```

```
plot(rand[,1], rand[,2],  col="blue", main="BB8 copula")

#goodness of fit
gf <- gofCopula(BB8Copula(), m, N = 50)
gf

# Build the bivariate distribution
P_shape = 2.254787
P_scale = 6.922302
Q_shape = 2.427970
Q_scale = 7.790036
my_dist <- mvdc(BB8Copula(param = c(2.267418, 0.897979)),
             margins = c("weibull","weibull"),
             paramMargins = list(list(shape = P_shape, scale = P_scale),
                            list(shape = Q_shape, scale = Q_scale)))

# Generate random sample observations from the multivariate distribution
v <- rMvdc(NROW(speed_Q), my_dist)

# Compute the density
pdf_mvd <- dMvdc(v, my_dist)
# Compute the CDF
cdf_mvd <- pMvdc(v, my_dist)

# 3D plain scatterplot of the generated bivariate distribution
par(mfrow = c(1, 2))
scatterplot3d(v[,1],v[,2], pdf_mvd,
          color="red", main="Density",
          xlab = "u1", ylab="u2", zlab="pMvdc",pch=".")
scatterplot3d(v[,1],v[,2], cdf_mvd,
          color="red", main="CDF",
          xlab = "u1", ylab="u2", zlab="pMvdc",pch=".")

#Plot PDF
persp(my_dist, dMvdc, xlim = c(-4, 4), ylim=c(0, 2), main = "Density")
contour(my_dist, dMvdc, xlim = c(-1, 4), ylim=c(0, 2), main = "Contour plot")

#Plot CDF
persp(my_dist, pMvdc, xlim = c(-4, 4), ylim=c(0, 2), main = "CDF")
contour(my_dist, pMvdc, xlim = c(-0, 4), ylim=c(0, 2), main = "Contour plot")

#density of the copula (PDF) green plot
persp(BB8Copula(param = c(2.267418, 0.897979)), dCopula, theta = 30, phi = 30,
    expand = 0.5, col = "lightgreen", ltheta = 100,xlab = "x",ticktype = "detailed",
    ylab = "y", zlab = "Density of the Gumbel copula")
```

```
#density of the copula (CDF)green plot
persp(BB8Copula(param = c(2.267418, 0.897979)),pCopula, theta = 30, phi = 30,
      expand = 0.5, col = "lightgreen", ltheta = 100,xlab = "x",ticktype = "detailed",
      ylab = "y", zlab = "Cumulative distribution function")

# Plot the data for a visual comparison
plot(speed_P, speed_Q, main = 'Comparison between real and simulated data', col = "blue")
points(v[,1], v[,2], col = 'red')
legend('bottomright', c('Observed', 'Simulated'), col = c('blue', 'red'), pch=21)

cor(speed_P,speed_Q,method='spearman')
cor(v[,1], v[,2], method = 'spearman')

#q-q plot
qqplot(v[,1], speed_P)
qqplot(v[,2], speed_Q)

#correlation test to find standard error
cor.test.plus <- function(x) {
  list(x,
       Standard.Error = unname(sqrt((1 - x$estimate^2)/x$parameter)))
}

cor.test.plus(cor.test(v[,1], v[,2], alpha=.05))

cor.test(v[,1], v[,2], method = 'spearman')
```